\documentclass[11pt,letterpaper, onecolumn]{article}
\usepackage[font=small, labelfont={bf}, labelsep=space, justification=justified, skip=0pt]{caption}
\usepackage{subcaption}
\usepackage{fullpage}
\usepackage{times}
\usepackage{array}
\usepackage{ragged2e}
\usepackage{multirow}
\usepackage{tabularx} 
\usepackage{soul}
\usepackage[numbers,sort&compress]{natbib}
\usepackage{adjustbox}
\usepackage{stackengine}    
\usepackage{diagbox}        
\usepackage{booktabs}       
\usepackage{fancyhdr, amssymb, mathtools}
\usepackage[ruled,vlined]{algorithm2e}
\usepackage[dvipsnames]{xcolor}
\usepackage[margin = 1 in]{geometry}
\usepackage{graphicx}
\usepackage{outlines}
\usepackage{amsmath}

\DeclareMathOperator*{\argmin}{argmin}
\usepackage{dirtytalk}

\usepackage{color}

\usepackage{multirow}
\usepackage{arydshln}

\usepackage[english]{babel} 
\usepackage{blindtext}
\usepackage{amsfonts}
\usepackage{amssymb}
\usepackage{bbm}
\usepackage{bm}
\usepackage{upgreek}
\usepackage{mathtools}

\usepackage{xltabular}
\usepackage{wrapfig}
\usepackage{enumitem}
\usepackage{floatrow}
\usepackage[table]{xcolor}
\usepackage{colortbl}
\definecolor{VoidColor}{rgb}{1,1,1}
\definecolor{MatOne}{rgb}{0.2,0.6509803921568628,0.27058823529411763}
\definecolor{MatTwo}{rgb}{0.23137254901960785,0.2980392156862745,0.7529411764705882}
\definecolor{MatThree}{rgb}{0.7058823529411765,0.01568627450980392,0.14901960784313725}
\newcommand{\matbox}[1]{%
  \fcolorbox{black}{#1}{\textcolor{#1}{\rule{2.2em}{0.9ex}}}%
}
\usepackage{xfrac}
\usepackage[hang,flushmargin]{footmisc} 
\usepackage[colorlinks=true, allcolors=blue]{hyperref}

\usepackage{diagbox} 
\usepackage{multirow} 
\usepackage{cleveref} 
\Crefname{figure}{Fig.}{Figs.}

\newcommand{\cmt}[1]{} 

\newcommand{\eb}{\boldsymbol{e}}
\newcommand{\fb}{\boldsymbol{f}}

\newcommand{\mb}{\boldsymbol{m}}
\newcommand{\nb}{\boldsymbol{n}}
\newcommand{\rb}{\boldsymbol{r}}
\newcommand{\tb}{\boldsymbol{t}}
\newcommand{\ub}{\boldsymbol{u}}
\newcommand{\hb}{\boldsymbol{h}}

\newcommand{\vb}{\boldsymbol{v}}

\newcommand{\xb}{\boldsymbol{x}}

\newcommand{\zb}{\boldsymbol{z}}

\newcommand{\Bb}{\boldsymbol{B}}
\newcommand{\Cb}{\boldsymbol{C}}

\newcommand{\Fb}{\boldsymbol{F}}
\newcommand{\Ib}{\boldsymbol{I}}
\newcommand{\Jb}{\boldsymbol{J}}

\newcommand{\Tb}{\boldsymbol{T}}

\newcommand{\Wb}{\boldsymbol{W}}
\newcommand{\Xb}{\boldsymbol{X}}

\newcommand{\alphab}{\boldsymbol{\alpha}}
\newcommand{\epsilonb}{\boldsymbol{\varepsilon}}
\newcommand{\sigmab}{\boldsymbol{\sigma}}
\newcommand{\thetab}{\boldsymbol{\theta}}
\newcommand{\xib}{\boldsymbol{\xi}}

\newcommand{\rhob}{\boldsymbol{\rho}}
\newcommand{\phib}{\boldsymbol{\phi}}

\newcommand{\zetab}{\boldsymbol{\zeta}}

\newcommand{\niG}{\mathrm{G}} 
\newcommand{\nic}{\mathrm{c}} 
\newcommand{\nig}{\mathrm{g}} 
\newcommand{\niK}{\mathrm{K}} 
\newcommand{\niW}{\mathrm{W}} 
\newcommand{\niM}{\mathrm{M}} 
\newcommand{\nim}{\mathrm{m}} 

\newcommand{\niN}{\mathrm{N}} 
\newcommand{\nimm}{\mathrm{mm}} 
\newcommand{\niPa}{\mathrm{Pa}} 

\newcommand{\nimum}{\upmu\mathrm{m}}

\DeclareMathOperator{\Res}{Res}

\usepackage{titling}
\thanksheadextra{}{}
\thanksheadextra{}{} 
\usepackage{lipsum} 
\usepackage{amsmath}
 
\title{Multi-material Multi-physics Topology Optimization with Physics-informed Gaussian Process Priors}
\date{\vspace{-5ex}}
\usepackage{authblk}

\author[1]{Xiangyu Sun}
\author[1]{Shirin Hosseinmardi}
\author[1]{Amin Yousefpour}
\author[1,2]{Ramin Bostanabad \thanks{ Corresponding Author: raminb@uci.edu \\\href{https://github.com/Bostanabad-Research-Group/}{GitHub Repository}}}
\affil[1]{Department of Mechanical and Aerospace Engineering, University of California, Irvine}
\affil[2]{Department of Civil and Environmental Engineering, University of California, Irvine}

\begin{document}
\pagenumbering{arabic}
\sloppy
\maketitle
\noindent \textbf{Abstract}\\
\noindent  
Machine learning (ML) has been increasingly used for topology optimization (TO). However, most existing ML-based approaches focus on simplified benchmark problems due to their high computational cost, spectral bias, and difficulty in handling complex physics. These limitations become more pronounced in multi-material, multi-physics problems whose objective or constraint functions are not self-adjoint.
To address these challenges, we propose a framework based on physics-informed Gaussian processes (PIGPs). In our approach, the primary, adjoint, and design variables are represented by independent GP priors whose mean functions are parametrized via neural networks whose architectures are particularly beneficial for surrogate modeling of PDE solutions. 
We estimate all parameters of our model simultaneously by minimizing a loss that is based on the objective function, multi-physics potential energy functionals, and design-constraints.
We demonstrate the capability of the proposed framework on benchmark TO problems such as compliance minimization, heat conduction optimization, and compliant mechanism design under single- and multi-material settings. 
Additionally, we leverage thermo-mechanical TO with single- and multi-material options as a representative multi-physics problem. We also introduce differentiation and integration schemes that dramatically accelerate the training process.
Our results demonstrate that the proposed PIGP framework can effectively solve coupled multi-physics and design problems simultaneously--generating super-resolution topologies with sharp interfaces and physically interpretable material distributions. We validate these results using open-source codes and the commercial software package COMSOL.


\section{Introduction} \label{sec intro}

Topology optimization (TO) is a powerful design tool in applications involving structural mechanics, fluid flow, and multi-physics systems \citep{bendsoe_topology_2004,borrvall_topology_2003,zhu_topology_2016,wu_advances_2021}. TO aims to optimally distribute material over a spatial domain by minimizing an objective function subject to governing physics and design constraints. Numerous TO methodologies exist, including the solid isotropic material with penalization (SIMP) approach \citep{bendsoe_generating_1988,andreassen_efficient_2011}, the level-set method \citep{van_level_2013,chen_topology_2017}, and evolutionary structural optimization (ESO) \citep{tanskanen_evolutionary_2002}. 

TO has rapidly evolved to address increasingly challenging design problems. Representative examples include stress-constrained optimization \citep{da_stress_2019}, designs involving dynamic fracture or plasticity \citep{han_general_2025,wu_topology_2022,jia_controlling_2023}, magneto-mechanical materials \citep{zhao_encoding_2023,zhao_topology_2022}, thermo-mechanical structural design \citep{wang_topology_2024,giraldo_multi_2020,wang_multi_2023}, and design of soft robotics \citep{kobayashi_computational_2024,wang_codesign_2025}. 
The focus of this work is on multi-material and multi-physics applications where the primary challenges are related to: (1) the high-dimensional and non-convex design space induced by multiple candidate materials and their relative properties and spatial arrangements \citep{zuo_mm_2017,da_mm_2022,sanders_polymat_2018,tavakoli_alternating_2014,sanders_multi_2018,asadpoure_topology_2025}; (2) the expensive design evaluations that rely on multi-physics partial differential equations (PDEs) \citep{park_conceptual_2019,ali_toward_2022}; and (3) complex and non-self-adjoint objectives in applications such as compliant gripper or actuator designs \citep{pedersen_topology_2001,zhang_topology_2018,liu_topology_2024,jin_compliant_2025}. 
These challenges may coexist as well, for instance in multi-material stress-constrained TO \citep{xu_stress_2021,han_multi_2025}, multi-physics actuator design \citep{sigmund_design_2001,du_topology_2009,onodera_design_2025}, multi-material multi-physics compliant device design \citep{sigmund_design_2001_2}, and multi-material thermo-mechanical TO \citep{wang_multi_2023}.

The majority of existing TO algorithms, such as SIMP and level-set, follow the ``nested'' analysis and design (NAND) setting which begins with an initial design and then computes the state variables (e.g., displacements in compliance minimization or CM) by solving the governing equations using numerical methods such as the finite element method (FEM). The resulting state fields are subsequently used to evaluate sensitivities which guide the design update using optimizers such as the optimality criteria (OC) method \citep{bendsoe_optimization_1995} or the method of moving asymptotes (MMA) \citep{svanberg_method_1987}. This analysis-update cycle is repeated until convergence.

While nested evaluation of state and sensitivity fields has certain advantages (e.g., modularity), it can incur high costs particularly in the presence of nonlinear constitutive response \citep{chen_topology_2017} or in large-scale (especially 3D) TO problems with millions of degrees of freedom \citep{alexandersen_large_2016,zheng_efficient_2024,jia_fenitop_2024,alexandersen_large_2026}. Simultaneous TO or SAND (simultaneous analysis and design) offers an alternative solution where the design and state variables are solved concurrently and it can improve computational efficiency in certain settings via augmented Lagrangian formulations \citep{sankara_truss_1994}. Nevertheless, nested frameworks remain more prevalent in the literature as simultaneous formulations are more difficult to implement and require effective algorithms to integrate design updates within the analysis loop.

Complementary to traditional methods, machine learning (ML) techniques are increasingly employed for TO. The majority of early works in this area leveraged generative models such as variational autoencoders (VAEs) \cite{RN941} and generative adversarial networks (GANs) \cite{RN965} to do instant or iteration-free designs \cite{RN1864,RN2026}. The idea behind these works is to generate large training data via traditional methods and then use that dataset to build a generative model. 
Other approaches focus on computational acceleration. For example, \citet{wang_deep_2021} trained a deep NN to predict high-resolution structures from low-resolution inputs with negligible cost.
\citet{chi_universal_2021} and \citet{senhora_machine_2022} accelerated large-scale TO by replacing expensive FEM-based sensitivity solvers with NN surrogates. Additionally, \citet{gavris_topology_2024} proposed a filtering technique based on graph neural networks (NNs) to suppress gray transition regions and reduce the number of iterations required for convergence. Despite these advancements, a primary limitation of such data-driven methods is their heavy reliance on extensive training datasets generated by traditional TO solvers.

Complementary to data-driven approaches, physics-informed machine learning (PIML) schemes have emerged where ML models are either combined with traditional methods or replace them. Broadly, existing PIML-based methods leverage three distinct strategies:
(1) NN-parameterized design variables embedded in conventional approaches \citep{hoyer_neural_2019, zhang_tonr_2021, chandrasekhar_MMTOuNN_2021, sanu_neural_2025, liu_neural_2025};
(2) PIML-based state solutions with design updates based on conventional optimizers (e.g., OC/MMA) \citep{he_deep_TO_2023, zhao_physics_2024}; and
(3) parameterization of both design and state variables by a model whose architecture and loss function are based on the design objective and physics \citep{jin_compliant_2025, jeong_complete_2023, yousefpour_simultaneous_2025, sun_smo_2025}.
The first strategy inherits most issues associated with traditional methods while also increasing the costs. The second and third strategies offer mesh-free alternatives by solving governing PDEs through the minimization of PDE residuals \citep{yousefpour_simultaneous_2025} or the deep energy method (DEM) \citep{haghighat_pinn_2021, nguyen_DEM_2020} which leverages variational formulations, as opposed to the strong form of the PDE system, to improve training stability and convergence rates.

Similar to conventional approaches, most ML-based TO methods have a nested structure which can be computationally expensive when state variables are parameterized by NNs.  For example, \Citet{jeong_complete_2023} reported an average runtime of four hours for 2D benchmark CM examples, where displacements were solved via the DEM within a nested density update loop using NN parameterizations.
However, we argue that the nested structure is not strictly necessary in ML-based settings as we have recently shown for minimizing power dissipation involving Stokes flow \citep{yousefpour_IDETC_2025,yousefpour_simultaneous_2025,yousefpour_localized_2025} and CM in solids \citep{sun_smo_2025}. 

Regardless of SAND- or NAND-based formulations, ML-based TO approaches including our previous works have not been tested on non-trivial applications involving multi-material, multi-physics, and non-self-adjoint designs. Hence, in this work we develop a unified framework for a broad class of TO problems (including CM, heat conduction optimization, compliant mechanism design, and thermo-mechanical device design) under multi-material and multi-physics settings. Our SAND approach is based on physics-informed Gaussian process (PIGP) priors which combine the strengths of both kernels and deep NNs. The kernels enable \textit{automatic} satisfaction of all Dirichlet-type constraints on design or state variables \citep{mora_gaussian_2024} which, in turn, guarantees the admissibility of our parameterized solution space. The deep NNs provide the representation power that is needed to learn various spatial distributions for material phases and state variables in 2D or 3D across various applications. 

Within our framework, the design, state, and adjoint fields are each endowed with independent GP priors using independent kernels and mean functions. For a multi-dimensional variable (e.g., displacement vector or multi-material phases), we also use independent kernels across the dimensions but a shared multi-output mean function.
This architecture improves scalability, facilitates numerical stability, and accelerates convergence to thermal and mechanical equilibrium. 
We construct all the mean functions using a specialized NN architecture named parametric grid convolution attention network (PGCAN) \citep{shishehbor_parametric_2024}. This architecture mitigates the spectral bias of multi-layer perceptrons (MLPs) and better captures sharp gradients and localized features that are critical in TO.

We devise multiple strategies to ensure efficiency and robustness of our approach. We always fix the hyperparameters of all the kernels to avoid repeated matrix inversions and estimate the parameters of all PGCANs by minimizing a physics-based loss. This loss combines (1) the objective function augmented with adjoint contributions, (2) the variational potential energy functionals for both state and adjoint variables, and (3) all design constraints such as prescribed mass or cost fractions. During training, all parameters are simultaneously updated efficiently via backpropagation (i.e., reverse-mode automatic differentiation or AD) and curriculum training. Our loss terms include spatial integrals of differentiated fields which can be estimated using numerical integration and AD. However, to accelerate this estimation without introducing error or any mesh-dependency, we leverage shape functions and reduced-order Gauss integration. Specifically, in each training iteration we evaluate the state and adjoint variables at a set of dynamic collocation points (CPs) that match a group of finite elements (FEs) whose centers are used to infer the design fields \citep{jeong_complete_2023}. We find this approximation strategy to be more robust than alternatives based on finite difference (FD). 

The remainder of this paper is organized as follows. \Cref{sec pde} reviews the governing equations related to our multi-physics problems. \Cref{sec method} introduces the proposed approach including formulations for general multi-material and multi-physics TO, parameterization with GPs, numerical approximation via shape functions, and continuous sensitivity analysis under multi-physics settings. \Cref{sec result} presents numerical results across multiple examples and methods including our approach, ordered SIMP \citep{da_mm_2022}, TO with polygonal FE (PolyMat) \citep{sanders_polymat_2018}, and COMSOL. Finally, \Cref{sec conclusion} concludes the paper and outlines future research directions.

\section{Governing Equations} \label{sec pde} 
Our studies involve thermomechanical systems and hence we review the corresponding governing equations and their coupling in \Cref{subsec pde mech,subsec pde heat,subsec pde coup}. 

\subsection{Mechanical} \label{subsec pde mech} 
Denote the displacement vector by $\ub(\xb) = [u_1(\xb),\dots,u_{n_d}(\xb)]^{\text{T}}$ where $\xb = [x_1, \ldots, x_{n_d}]^{\text{T}}$ represents the coordinates of a point in the design domain and $n_d \in \{2,3\}$ is the spatial dimensionality. Under the assumption of small deformation, the kinematic relation between $\ub(\xb)$ and the strain tensor $\epsilonb(\xb)$ is expressed as:
\begin{equation}\label{eq kinematic}
    \epsilonb(\xb) = \frac{1}{2} \big( \nabla \ub(\xb) + \nabla \ub^{\text{T}}(\xb) \big),
\end{equation}
where $\nabla$ is the gradient operator. For a linear elastic material with the fourth-order stiffness tensor $\Cb(\xb)$, the constitutive relation is given by:
\begin{equation}\label{eq constitutive}
    \sigmab(\xb) = \Cb(\xb) : \epsilonb(\xb) = \Cb(\xb):\nabla\ub(\xb),
\end{equation}
where the last equality is due to the minor symmetry of $\Cb(\xb)$, i.e., $\frac{1}{2}\Cb(\xb): \big( \nabla \ub(\xb) - \nabla \ub^{\text{T}}(\xb) \big) = \bm{0}$. Ignoring the body forces, the governing PDE system is now written as:
\begin{equation}\label{eq pde}
    \nabla \cdot \sigmab(\xb) = \nabla \cdot  \big( \Cb(\xb):\nabla\ub(\xb) \big) = \mathbf{0} , \quad \forall \xb \in \Omega,
\end{equation}
where $\nabla\cdot(\cdot)$ and $\Omega$ denote the divergence operation and the domain, respectively. We consider the following four types of BCs in this paper:
\begin{subequations} \label{eq bc}
    \begin{align}
    &\ub(\xb) = \tilde{\ub}(\xb), \quad \forall \xb \in \partial \Omega_{h}, \label{eq bc 1} \\
    &\sigmab(\xb) \cdot \nb(\xb) = \fb(\xb), \quad \forall \xb \in \partial \Omega_t, \label{eq bc 2}\\
    &\sigmab(\xb) \cdot \nb(\xb) = 0, \quad \forall \xb \in \partial \Omega_0, \label{eq bc 3}\\
    &\sigmab(\xb) \cdot \nb(\xb) = -K_s\delta(\xb - \xb_s)\big(\ub(\xb)\cdot\eb_s\big)\eb_s, \quad s = 1,\dots,n_s,\quad \forall \xb_s \in \partial \Omega_s, \label{eq bc 4}
    \end{align}
\end{subequations}
where $\partial \Omega_h$, $\partial \Omega_t$, $\partial \Omega_0$, and $\partial \Omega_s$ denote the boundary regions associated with displacement BCs, traction BCs, free surfaces, and surfaces connected to elastic springs, respectively. 
The displacement BCs are specified by the vector-valued function $\tilde{\ub}(\xb) := [\tilde{u}_1(\xb),\dots,\tilde{u}_{n_d}(\xb)]^{\text{T}}$, the outward unit normal vector along the boundary is $\nb(\xb) := [n_1(\xb),\dots,n_{n_d}(\xb)]^{\text{T}}$, and the applied traction is given by the function $\fb(\xb) := [f_1(\xb),\dots,f_{n_d}(\xb)]^{\text{T}}$.
The compliant BCs in \Cref{eq bc 4} are modeled using spring elements where $K_s$ denotes the spring stiffness and $n_s$ is the number of attached springs. The quantities $\eb_s$ and $\delta(\xb)$ represent the direction of each spring and the Dirac delta function, respectively. 

The displacement field corresponding to the boundary value problem in \Cref{eq pde,eq bc} can be obtained by minimizing the total potential energy:
\begin{equation}\label{eq pi solid}
    L_M\big(\ub(\xb)\big) = \frac{1}{2} \int_{\Omega} \sigmab(\xb) : \epsilonb(\xb) \, dV + \sum_{s=1}^{n_s}\frac{1}{2}K_i(\ub(\xb)\cdot\eb_s)^2- \int_{\partial \Omega_t} \fb^{\text{T}}(\xb) \ub(\xb) \, ds,
\end{equation}
subject to satisfying the prescribed BCs. The three terms on the right-hand-side (RHS) from left to right are the strain energy, the energy stored in the springs, and the external work. Our method for minimizing \Cref{eq pi solid} is detailed in \Cref{sec method}.

\subsection{Heat Transfer} \label{subsec pde heat} 
Steady-state heat conduction, with spatially varying conductivity $\kappa(\xb)$ and volumetric convection to the ambient temperature $T_{\infty}$, is governed by the following PDE:
\begin{equation}\label{eq pde heat}
    -\nabla \cdot \big(\kappa(\xb) \nabla T(\xb)\big) 
    + h_v \big(T(\xb) - T_\infty\big) = s(\xb), 
    \quad \forall \xb \in \Omega,
\end{equation}
where $s(\xb)$ is a heat source and $h_v$ denotes the volumetric convection coefficient. The convection term in \Cref{eq pde heat} corresponds to the behavior of a thin 2D plate or a 3D porous structure exchanging heat uniformly with the surrounding environment. 

The BC corresponding to \Cref{eq pde heat} are generally classified into Neumann and Dirichlet types:
\begin{subequations} \label{eq bc heat}
    \begin{align}
    -\kappa(\xb)\nabla T(\xb)\cdot \nb(\xb) &= q(\xb), \quad \forall \xb \in \partial \Omega_{N}, \label{eq bc heat 1} \\
    T(\xb) &= \tilde{T}(\xb), \quad \forall \xb \in \partial \Omega_D, \label{eq bc heat 2}
    \end{align}
\end{subequations}
where $q(\xb)$ prescribes the outward heat flux on the boundary $\partial\Omega_N$ and $\tilde{T}(\xb)$ determines the temperature along $\partial\Omega_D$. Regions without specified flux or temperature are treated as thermally insulated. Similar to \Cref{eq pi solid}, the potential functional $L_T\big(T(\xb)\big)$ for the heat conduction problem can be written as:
\begin{equation}\label{eq pi heat}
\begin{aligned}
    L_T\big(T(\xb)\big)
    &= \frac{1}{2} \int_{\Omega} 
        \kappa(\xb) |\nabla T(\xb)|^2 \, dV
       + \frac{1}{2} \int_{\Omega} 
        h_v(\xb) \big(T(\xb) - T_\infty\big)^2 \, dV\\  
    &  - \int_{\Omega} s(\xb)\,T(\xb) \, dV
       + \int_{\partial \Omega_N} q(\xb)\,T(\xb) \, ds,
\end{aligned}
\end{equation}
which must be minimized to obtain the temperature field subjected to a prescribed BC. We refer to the first and third terms on the RHS of \Cref{eq pi heat} as the thermal energy and the source energy, respectively.

\subsection{Thermo-mechanical Coupling} \label{subsec pde coup} 
Under the small-deformation assumption, the strain tensor $\epsilonb(\xb)$ can be additively decomposed as:
\begin{equation}\label{eq kinematic tm 1}
    \epsilonb(\xb) = \epsilonb_{m}(\xb) + \epsilonb_{th}(\xb),
\end{equation}
where $\epsilonb_{m}(\xb)$ and $\epsilonb_{th}(\xb)$ denote the mechanical and thermal strain tensors, respectively. The thermal strain is expressed as:
\begin{equation}\label{eq kinematic tm 2}
    \epsilonb_{th}(\xb) = \alpha(\xb) (T(\xb) - T_{\infty})\Ib = \alphab(\xb) \Delta T(\xb),
\end{equation}
where $\alpha(\xb)$ is the isotropic thermal expansion coefficient, $\Ib$ is the identity tensor, and $\alphab(\xb) := \alpha(\xb)\Ib$. With this decomposition, the constitutive relation in \Cref{eq constitutive} becomes:
\begin{equation}\label{eq constitutive tm}
    \sigmab(\xb) = \Cb(\xb) : \epsilonb_m(\xb) = \Cb(\xb) : \big(\epsilonb(\xb) - \epsilonb_{th}(\xb)\big).
\end{equation}
The BCs for the mechanical and thermal fields are given in \Cref{eq bc,eq bc heat}, respectively. For thermo-mechanical coupling, the potential energy functional in \Cref{eq pi solid} can be modified as:
\begin{equation}\label{eq pi tm}
\begin{aligned}
    L_{C}\big(\ub(\xb), T(\xb)\big)
    &= \frac{1}{2} \int_{\Omega} \big(\epsilonb(\xb) - \epsilonb_{th}(\xb)\big) : \Cb(\xb) : \big(\epsilonb(\xb) - \epsilonb_{th}(\xb)\big)\,dV \\
    &\quad
    + \sum_{s=1}^{n_s}\frac{1}{2}K_i\big(\ub(\xb)\cdot\eb_s\big)^2 
    - \int_{\partial \Omega_t} \fb^{\text{T}}(\xb)\,\ub(\xb)\,ds.
\end{aligned}
\end{equation}
The displacement $\ub(\xb)$ with the presence of a temperature field is solved by minimizing \Cref{eq pi tm} with $T(\xb)$ being the minimizer of \Cref{eq pi heat}. As detailed in \Cref{sec method}, we simultaneously solve for both $\ub(\xb)$ and $T(\xb)$ within a single optimization loop. 

\section{Proposed Framework} \label{sec method}
In multi-material and multi-physics TO, the objective is to determine the spatial distribution of material that minimizes the objective functional $L_C(\cdot)$ while satisfying a set of constraints arising from the governing physics and design requirements. Here, we consider thermomechanical coupling and so the objective can be expressed explicitly as:
\begin{equation}\label{eq LC}
\begin{aligned}
L_C(\cdot) := L_C\big(\ub(\xb,\rhob(\xb)), T(\xb,\rhob(\xb)),\rhob(\xb)\big),
\end{aligned}
\end{equation}
where $\rhob(\xb) = [\rho_0(\xb),\rho_1(\xb),\dots,\rho_{n_m}(\xb)]^{\text{T}}$ is the design vector and $n_m$ denotes the number of candidate materials. Inspired by the concept of phases in materials science, we interpret each $\rho_i(\xb)$ as the local volume fraction of the $i$th phase at spatial location $\xb$ where $\rho_0(\xb)$ represents void with near-zero density or thermal conductivity. The design field satisfies the partition-of-unity constraint $\sum_{i = 0}^{n_m}\rho_i(\xb) = 1$ ensuring that exactly one unit of volume is distributed among all phases at each point. 

While we expect the final topology to satisfy the multiphase discrete constraint that requires a spatial location to be occupied by a single material phase, mixtures of material phases appear during intermediate optimization steps. As a result, homogenization is required to compute the effective material properties at each spatial point. Inspired by relevant works \citep{sigmund_design_2001,chandrasekhar_MMTOuNN_2021}, we assume that quantities such as the stiffness $\Cb(\cdot)$, thermal conductivity $\kappa(\cdot)$, thermal expansion coefficient $\alpha(\cdot)$, and the heat source $s(\cdot)$ depend on the design field $\rhob(\xb)$ through a penalization scheme. That is:
\begin{subequations} \label{eq const p}
    \begin{align}
    E(\xb) &= \sum^{n_m}_{i=0}E_i\rho^{p}_i(\xb), \label{eq const p 1} \\
    \kappa(\xb) & = \sum^{n_m}_{i=0}\kappa_i\rho^{p}_i(\xb), \label{eq const p 2} \\
    \alpha(\xb) & = \sum^{n_m}_{i=0}\alpha_i\rho^{p}_i(\xb), \label{eq const p 3} \\
    s(\xb) & = \sum^{n_m}_{i=0}s_i\rho^{p}_i(\xb), \label{eq const p 4} 
    \end{align}
\end{subequations}
where $p$ is the penalization exponent that promotes binary design and the constants $E_i$, $\kappa_i$, $\alpha_i$, and $s_i$ correspond to, respectively, the Young's modulus, thermal conductivity, thermal expansion coefficient, and heat source for each material phase. For simplicity, a uniform Poisson's ratio $\nu = 0.31$ is used for all phases. 

Motivated by the formulation in our recent work \citep{sun_smo_2025}, we propose to solve for $\rhob(\xb)$ in a generic thermomechanical TO problem as (the dependence of $\rhob$ on $\xb$ is occasionally dropped for notational simplicity):
\begin{subequations}\label{eq cm1}
\begin{align}
\widehat{\rhob}(\xb) =
\argmin_{\rhob} 
L_C\big(&\ub(\xb,\rhob), T(\xb,\rhob), \rhob\big)
= \argmin_{\rhob}\int_{\Omega} l_c\big(\ub(\xb,\rhob), T(\xb,\rhob), \rhob\big)\, dV, \label{eq cm1 a}
\intertext{subject to:}
\ub(\xb,\rhob) = \argmin_{\ub} L_M\big(&\ub(\xb,\rhob), T(\xb,\rhob), \rhob\big) = \argmin_{\ub}\bigg( \frac{1}{2}
\int_{\Omega} l_m\big(\ub(\xb,\rhob), T(\xb,\rhob), \rhob(\xb)\big)\, dV \notag\\
&+ \sum_{s=1}^{n_s}\frac{1}{2}K_i\big(\ub(\xb)\cdot\eb_s\big)^2 - \int_{\partial \Omega_{t}} \fb(\xb)^{\text{T}} \ub(\xb,\rhob) \, dA\bigg), \label{eq cm1 b}
\\
T(\xb,\rhob) = \argmin_{T} L_T\big(&T(\xb,\rhob), \rhob\big) = \argmin_{T}\bigg(\frac{1}{2} \int_{\Omega}\kappa(\xb) |\nabla T(\xb,\rhob)|^2 \, dV \notag\\
& + \frac{1}{2} \int_{\Omega} h_v \big(T(\xb,\rhob) - T_\infty\big)^2 \, dV 
- \int_{\Omega} s(\xb)\,T(\xb,\rhob) \, dV \notag \\
& + \int_{\partial \Omega_N} q(\xb)\,T(\xb,\rhob) \, ds\bigg),\label{eq cm1 c}
\\
C_M\big(\rhob\big) &= \int_{\Omega} \sum_{i=0}^{n_m} \rho_i(\mathbf{x})\, \bar{\rho_i} \, dV - \psi_m M_0 \leq 0,\label{eq cm1 d} 
\\
C_P\big(\rhob\big) &= \int_{\Omega} \sum_{i=0}^{n_m} \rho_i(\mathbf{x})\, \bar{p_i} \, dV - \psi_p P_0 \leq 0, \label{eq cm1 e} 
\\
C_i\big(\rhob\big) &= 
\int_{\Omega} c_i\big(\rhob\big)\, dV = 0, 
\quad i = 1,\dots,n_c, \label{eq cm1 f}
\\
R_{b}^{(i)}\big(\ub(\xb,\rhob),T(\xb,\rhob)\big) &= 
\int_{\partial\Omega_{\ub}} r_b^{(i)}\big(\ub(\xb,\rhob),T(\xb,\rhob)\big)\, dA = 0, 
\quad i = 1,\dots,n_{b}, \label{eq cm1 g}
\end{align}
\end{subequations}
where $C_M(\cdot)$, $C_P(\cdot)$, and $C_i(\cdot)$ denote the total mass, total price, and design constraints, respectively. $\bar{\rho}_i$ and $\bar{p}_i$ represent the discrete physical density and unit-volume price associated with each material phase.
The optimization is subject to global constraints defined with respect to the domain’s maximum capacity. Let $M_0$ and $P_0$ denote the reference mass and price, respectively, computed under the hypothetical case that the entire design domain is filled with the densest and the most expensive material. The corresponding target fractions, $\psi_m$ and $\psi_p$, then bound the admissible designs by requiring the actual mass $M$ and price $P$ to satisfy \Cref{eq cm1 d} and \Cref{eq cm1 e}, respectively.
We assume that $L_C(\cdot)$ adopts an integral representation over the domain, expressed with the local functional $l_c(\ub, T, \rhob)$, and that $l_m(\ub, T, \rhob)$ denotes the strain-energy density.
The BC constraints for both displacement and temperature fields are written as $R_b^{(i)}\big(\ub(\xb,\rhob), T(\xb,\rhob)\big)$, each defined as the integral of the corresponding local function $r_b^{(i)}(\cdot)$. The integral representation assumes that driving the residual integral to zero forces the point-wise residual to vanish, given that the local functions are adequately smooth and bounded. The equilibrium conditions for the mechanical and thermal fields are enforced through the minimization of \Cref{eq cm1 b} and \Cref{eq cm1 c}, respectively.

Our approach for solving \Cref{eq cm1} is (1) simultaneous since we optimize the objective function while jointly minimizing the PDE residuals, and (2) mesh-free since we refrain from discretizing any of the state or design variables. 
We begin by parameterizing all variables via the differentiable functions $\zb(\xb;\zetab)$:
\begin{equation}\label{eq param}
\zb(\xb;\zetab) := [ u_1(\xb,\rhob),\dots,u_{n_d}(\xb,\rhob),T(\xb,\rhob),\rho_0(\xb),\dots,\rho_{n_m}(\xb)]^{\text{T}},
\end{equation}
where $\zetab$ denotes the parameters of $\zb(\cdot)$. 
For notational convenience, we introduce the shorthands  
$\zb_{\rho}(\xb;\zetab) := [\rho_0(\xb),\dots,\rho_{n_m}(\xb)]^{\text{T}}$,   
$\zb_u(\xb;\zetab) := [u_1(\xb,\rhob),\dots,u_{n_d}(\xb,\rhob)]^{\text{T}}$,  
and  
$z_t(\xb;\zetab) := T(\xb,\rhob)$.  
With these definitions, we cast the constrained formulation in \Cref{eq cm1} as an unconstrained optimization problem using the penalty method \citep{nocedal_numerical_2006}:
\begin{equation}\label{eq penal1}
    \begin{aligned}
        \widehat{\zetab} = \argmin_{\zetab} L\big(\zb(\xb;\zetab)\big) = \argmin_{\zetab} \bigg[
        &L_C\big(\zb(\cdot)\big) 
        + \omega_m L_M\big(\zb(\cdot)\big) 
        + \omega_t L_T\big(z_t(\cdot),\zb_{\rho}(\cdot)\big) \\
        & + \omega_{v} C_M^2\big(\zb_{\rho}(\cdot)\big)
        + \omega_{p} C_P^2\big(\zb_{\rho}(\cdot)\big)\\
        &+ \sum_{i=1}^{n_c} \omega_i C_i^2\big(\zb_{\rho}(\cdot)\big) + \sum_{j=1}^{n_b} \omega_{j+n_c} R_b^{(j)}\big(\zb_{u}(\cdot),z_t(\cdot)\big)^2
        \bigg],
    \end{aligned}
\end{equation}
where $\omega_m$, $\omega_t$, $\omega_v$, $\omega_p$, and $\omega_i$ (for $i = 1,\dots,n_b + n_c$) are scalar penalty weights associated with each loss term. We solve the unconstrained optimization problem in \Cref{eq penal1} via the Adam optimizer which iteratively adjusts $\zetab$ until the convergence criterion is met. 

Effective and efficient minimization of $L\big(\zb(\xb;\zetab)\big)$ in \Cref{eq penal1} relies on the designed functional form for $\zb(\xb;\zetab)$, accurate gradient calculations with respect to $\xb$ and $\rhob$ for updating $\zetab$, and numerically estimating the integrals in \Cref{eq cm1} with sufficient accuracy. In what follows, we first describe the proposed functional form for $\zb(\xb;\zetab)$ in \Cref{subsec method param 1}. Then, in \Cref{subsec method param 2} we provide details on our adjoint analysis which we use for gradient calculations and parameter updates. In \Cref{subsec method param 1,subsec method param 2} we parameterize all the variables with GP priors whose mean functions are neural field representations with a particular architecture that is ideally suited to approximating PDE solutions. We describe these mean functions in \Cref{subsec neural mean} and conclude this section by providing implementation details regarding numerical approximations and stability in \Cref{subsec method sf,subsec method training}. 

\subsection{Parameterization of State and Design Variables} \label{subsec method param 1}
To solve the optimization problem in \Cref{eq penal1}, it is essential to ensure that all loss components are effectively minimized within a reasonable number of iterations. Since these terms have largely different ranges, we design $\zb(\xb;\zetab)$ such that it automatically satisfies BCs-- enabling us to exclude \Cref{eq cm1 f} from the constraints. 
Specifically, we assign GP priors to all state and design variables where each GP is equipped with its own kernel and mean function. For the displacement and design variables, the mean functions are multi-output and defined as:
\begin{subequations} \label{eq mean m}
    \begin{align}
    \mb_u(\xb;\thetab_u) &:= [m_{u_1}(\xb;\thetab_u),\dots,m_{u_{n_d}}(\xb;\thetab_u)]^{\text{T}}, \label{eq mean m 1}\\
    \mb_{\rho}(\xb;\thetab_{\rho}) &:= [m_{\rho_0}(\xb;\thetab_{\rho}),\dots,m_{\rho_{n_m}}(\xb;\thetab_{\rho})]^{\text{T}}, \label{eq mean m 2}
    \end{align}
\end{subequations}
where $m_{u_1}(\xb;\thetab_u),\dots,m_{u_{n_d}}(\xb;\thetab_u)$ and $m_{\rho_0}(\xb;\thetab_{\rho}),\dots,m_{\rho_{n_m}}(\xb;\thetab_{\rho})$ are the mean functions for the displacement components and material-phase design variables, respectively. 
The temperature field is parameterized by its own mean function $m_T(\xb;\thetab_t)$. Distinct parameter sets are used for each variable to enhance the representation power of $\zb(\xb;\zetab)$. The specific parametric forms adopted in this work are detailed in \Cref{subsec neural mean}.

We take the displacement field as an example to illustrate that our GP-based representation automatically satisfies the BCs \textit{regardless} of the values assigned to its kernel's hyperparameters and its mean function's parameters. 
The posterior distribution of a GP prior conditioned on the prescribed displacement data is:
\begin{subequations} \label{eq GP u}
    \begin{align}
    u_i(\xb^*; \thetab_u, \phib_{u})
    &:= \mathbb{E}[u_i^* \mid \tilde{\ub}_i, \Xb_{u_i}]
    = m_{u_i}(\xb^*; \thetab_u) + \Wb_{u_i}^{\text{T}} \rb_{u_i}, \label{eq GP u 1} \\
    \Wb_{u_i}
    &= c_u^{-1}(\Xb_{u_i}, \Xb_{u_i}; \phib_{u}) \,
       c_u(\Xb_{u_i}, \xb^*; \phib_{u}), \label{eq GP u 2} \\
    \rb_{u_i}
    &= \tilde{\ub}_i - m_{u_i}(\Xb_{u_i}; \thetab_u),\quad i = 1,\dots,n_d, \label{eq GP u 3}
    \end{align}
\end{subequations}
where $\xb^* = [x_1^*,\dots, x_{n_d}^*]^{\mathrm{T}}$ is an arbitrary query point in the design domain and $c_u(\xb,\xb^{\prime}; \phib_{u})$ denotes the kernel (or covariance function) with hyperparameters $\phib_{u}$. 
The vector $\tilde{\ub}_i=\tilde{u}_i(\Xb_{u_i})$ contains the prescribed displacements at $\Xb_{u_i}$ and $c_u(\Xb_{u_i},\Xb_{u_i};\phib_{u})$ is the corresponding covariance matrix. 
The residual vector $\rb_{u_i}$ measures the mismatch between the mean prediction $m_{u_i}(\Xb_{u_i};\thetab_u)$ and the prescribed values $\tilde{\ub}_i$. By setting $\xb^* = \Xb_i$ in \Cref{eq GP u 1}, the left-hand side (LHS) recovers $\tilde{\ub}_i$ exactly, which is independent of the choice of mean function or kernel. Hence, the reconstructed displacement field obtained at arbitrary $\xb^*$ automatically satisfies the displacement BCs, provided that the set $\Xb_{u_i}$ is sufficiently large.

Similar to \Cref{eq GP u} we parametrize the temperature $T(\cdot)$ field as:
\begin{subequations} \label{eq GP t}
    \begin{align}
    T(\xb^*; \thetab_t, \phib_{t})
    &:= \mathbb{E}[T^* \mid \tilde{\Tb}, \Xb_{t}]
    = m_{t}(\xb^*; \thetab_t) + \Wb_{t}^{\text{T}} \rb_{t}, \label{eq GP t 1} \\
    \Wb_{t}
    &= c_{t}^{-1}(\Xb_{t}, \Xb_{t}; \phib_{t}) \,
       c_{t}(\Xb_{t}, \xb^*; \phib_{t}), \label{eq GP t 2} \\
    \rb_{t}
    &= \tilde{\Tb} - m_{t}(\Xb_{t}; \thetab_t), \label{eq GP t 3}
    \end{align}
\end{subequations}
where $\tilde{\Tb}=\tilde{T}(\Xb_t)$ is the vector of the prescribed temperatures at $\Xb_t$. Finally, the parameterization for $\rhob(\xb)$ is:
\begin{subequations} \label{eq GP rho}
    \begin{align}
    \rho_i(\xb^*; \thetab_{\rho}, \phib_{\rho}) &:= \mathbb{E}[\rho^*_i | \tilde{\rhob}_i, \Xb_{\rho_i}] = m_{\rho_i}(\xb^*; \thetab_{\rho}) + \Wb_{\rho_i}^{\text{T}} \rb_{\rho_i}, \label{eq GP rho 1} \\
    \Wb_{\rho_i} & = c_{\rho}^{-1}(\Xb_{\rho_i}, \Xb_{\rho_i}; \phib_{\rho}) \, c_{\rho}(\Xb_{\rho_i}, \xb^*; \phib_{\rho}), \label{eq GP rho 2} \\
    \rb_{\rho_i} & = \tilde{\rhob_i} - m_{\rho_i}(\Xb_{\rho_i}; \thetab_{\rho}),\quad i = 0,1,\dots,n_m, \label{eq GP rho 3}
    \end{align}
\end{subequations}
where the vector $\tilde{\rhob}_i = \tilde{\rho}_i(\Xb_{\rho})$ is the prescribed density at $\Xb_{\rho}$. We note that due to the partition of unity constraining $\rho_i = 1$ requires assigning zero values to other material phases at the same point. 

Any GP requires the specification of a kernel and a mean function. We describe the latter in \Cref{subsec neural mean} and for the former use the Gaussian kernel:
\begin{equation} \label{eq kernel}
    c(\xb, \xb^{\prime}; \phib) = s^2\exp \left\{ - (\xb - \xb')^{\text{T}} \, \mathrm{diag}(\phib) \, (\xb - \xb') \right\} + \mathbbm{1}\{\xb == \xb'\} \, \delta,
\end{equation}
where $\mathbbm{1}$ is the binary indicator function that equals 1 when $\xb = \xb'$ and 0 otherwise, $s^2$ denotes the process variance, and $\delta = 10^{-5}$ is the jitter that ensures $c(\Xb, \Xb; \phib)$ is invertible.  

As recommended by \citet{mora_gaussian_2024} we assign independent kernels with \textit{fixed} hyperparameters to each variable. This choice
(1) can naturally accommodate heterogeneous data, especially when BCs or prescribed design features are specified at different locations or along complex geometries;  
(2) reduces memory usage by constructing separate covariance matrices for each output variable, rather than a single large joint matrix; and  
(3) improves computational efficiency and stability, as all covariance matrices with fixed $\phib$ can be pre-computed, inverted, and cached.  
Following \citep{sun_smo_2025}, we use $\phib = 0.5$ for all kernels in our studies and do \textit{not} fine-tune it.

\subsection{Parameterization of Adjoint Fields} \label{subsec method param 2}
Since the parameterizations in \Cref{eq GP u,eq GP t} do \textit{not} explicitly encode the dependence of the displacement and temperature fields on the density, we employ the adjoint method to evaluate the sensitivity of the objective function with respect to the design variables. Depending on the specific TO problem, the adjoint fields associated with the displacement or temperature may not be directly available, requiring us to solve the corresponding adjoint equations under a multi-physics setting. Hereafter, we refer to displacement and temperature as the primary fields, and their respective adjoint variables as the adjoint fields. Following the notation in \Cref{eq param}, we express the adjoint fields as:
\begin{equation}\label{eq param adj}
\zb^{(a)}\big(\xb;\zetab^{(a)}\big) := [ u^{(a)}_1(\xb),\dots,u^{(a)}_{n_d}(\xb),T^{(a)}(\xb)]^{\text{T}},
\end{equation}
where $\zetab^{(a)}$ collects all parameters associated with the adjoint-field functions. Similar to the primary fields we assign GP priors to the adjoint displacement and temperature fields, each equipped with an independent kernel and sampled according to its corresponding BCs which are derived from the continuous adjoint analysis (see \Cref{subsec adj analysis}). 
Following the notation in \Cref{eq mean m}, the mean functions are written as  
$\mb^{(a)}_u(\xb;\thetab^{(a)}_u)
:= [m^{(a)}_{u_1}(\xb;\thetab^{(a)}_u),\dots,m^{(a)}_{u_{n_d}}(\xb;\thetab^{(a)}_u)]^{\mathrm{T}}$  
for the adjoint displacement and  
$m^{(a)}_T(\xb;\thetab^{(a)}_t)$  
for the adjoint temperature. These mean functions use parameter sets that are distinct from those employed for the primary state fields.  
Leveraging the reproducing property of GPs, the posterior distribution of the adjoint displacement at $\xb^*$ is:
\begin{subequations} \label{eq GP u adj}
    \begin{align}
    u^{(a)}_i(\xb^*; \thetab^{(a)}_u, \phib_{u})
    &:= \mathbb{E}[u_i^{(a)*} \mid \tilde{\ub}^{(a)}_i, \Xb^{(a)}_{u_i}]
    = m^{(a)}_{u_i}(\xb^*; \thetab^{(a)}_u) + \Wb^{(a)\text{T}}_{u_i} \rb^{(a)}_{u_i}, \label{eq GP u adj 1} \\
    \Wb^{(a)}_{u_i}
    &= c_u^{-1}(\Xb^{(a)}_{u_i}, \Xb^{(a)}_{u_i}; \phib_{u}) \,
       c_u(\Xb^{(a)}_{u_i}, \xb^*; \phib_{u}), \label{eq GP u adj 2} \\
    \rb^{(a)}_{u_i}
    &= \tilde{\ub}^{(a)}_i - m^{(a)}_{u_i}(\Xb^{(a)}_{u_i}; \thetab^{(a)}_u),\quad i = 1,\dots,n_d, \label{eq GP u adj 3}
    \end{align}
\end{subequations}
and for the adjoint temperature:
\begin{subequations} \label{eq GP t adj}
    \begin{align}
    T^{(a)}(\xb^*; \thetab^{(a)}_t, \phib_{t})
    &:= \mathbb{E}[T^{(a)*} \mid \tilde{\Tb}^{(a)}, \Xb^{(a)}_{t}]
    = m^{(a)}_{t}(\xb^*; \thetab^{(a)}_t) + \Wb_{t}^{(a)\text{T}} \rb^{(a)}_{t}, \label{eq GP t adj 1} \\
    \Wb^{(a)}_{t}
    &= c_{t}^{-1}(\Xb^{(a)}_{t}, \Xb^{(a)}_{t}; \phib_{t}) \,
       c_{t}(\Xb^{(a)}_{t}, \xb^*; \phib_{t}), \label{eq GP t adj 2} \\
    \rb^{(a)}_{t}
    &= \tilde{\Tb}^{(a)} - m^{(a)}_{t}(\Xb^{(a)}_{t}; \thetab_t). \label{eq GP t adj 3}
    \end{align}
\end{subequations}
The interpretation of the slightly modified symbols for the adjoint fields in \Cref{eq GP u adj} and \Cref{eq GP t adj} follows directly from the corresponding definitions in \Cref{eq GP u} and \Cref{eq GP t} for the primary fields. For simplicity, we adopt the same kernel type and the same set of \textit{fixed} hyperparameters for the adjoint fields as those used for the primary variables. We emphasize that solving for the adjoint fields is not necessary when the problem is self-adjoint, such as in CM or heat sink optimization.

Using \Cref{eq GP u,eq GP t,eq GP rho,eq GP u adj,eq GP t adj}, we now rewrite \Cref{eq param,eq param adj} as:
\begin{subequations} \label{eq param2}
    \begin{align}
    \zb^{\prime}(\xb;\thetab,\phib) &\coloneqq [ u_1(\xb;\thetab_u,\phib_{u}),\dots,u_{n_d}(\xb;\thetab_u,\phib_{u}),T(\xb;\thetab_{t},\phib_{t}),\notag\\
    & \rho_0(\xb;\thetab_{\rho},\phib_{\rho}),\dots,\rho_{n_m}(\xb;\thetab_{\rho},\phib_{\rho})]^{\text{T}}, \label{eq param2 a}\\
    \zb^{\prime(a)}(\xb;\thetab^{(a)},\phib) &\coloneqq [ u^{(a)}_1(\xb;\thetab^{(a)}_u,\phib_{u}),\dots,u^{(a)}_{n_d}(\xb;\thetab^{(a)}_u,\phib_{u}),T^{(a)}(\xb;\thetab^{(a)}_{t},\phib_{t})]^{\text{T}}.
    \label{eq param2 b}
    \end{align}
\end{subequations}
and introduce the following shorthands 
$\zb^{\prime}_u(\xb;\thetab_u,\phib_u)
\coloneqq [u_1(\xb;\thetab_u,\phib_{u}),\dots,u_{n_d}(\xb;\thetab_u,\phib_{u})]^{\text{T}}$,  
$\zb^{\prime}_{\rho}(\xb;\thetab_{\rho},\phib_{\rho})
\coloneqq [\rho_0(\xb;\thetab_{\rho},\phib_{\rho}),\dots,\rho_{n_m}(\xb;\thetab_{\rho},\phib_{\rho})]^{\text{T}}$,  
$z^{\prime}_t(\xb;\thetab_t,\phib_t) \coloneqq T(\xb;\thetab_t,\phib_t)$,
$\zb^{\prime (a)}_u(\xb;\thetab^{(a)}_u,\phib_u)
\coloneqq [u^{(a)}_1(\xb;\thetab^{(a)}_u,\phib_{u}),\dots,u^{(a)}_{n_d}(\xb;\thetab^{(a)}_u,\phib_{u})]^{\text{T}}$,  
and  
$z^{\prime (a)}_t(\xb;\thetab^{(a)}_t,\phib_t) \coloneqq T^{(a)}(\xb;\thetab^{(a)}_t,\phib_t)$.

By exploiting the reproducing property of GPs, we can eliminate all data-based constraints such as $R_b^{(j)}\big(\zb_m(\xb;\zetab)\big)$ and $C_i\big(z_{\rho}(\xb;\zetab)\big)$ from \Cref{eq penal1}.
Using \Cref{eq param2} with fixed $\hat{\phib} = [\hat{\phib}_u, \hat{\phib}_t,\hat{\phib}_{\rho}]^{\text{T}}$, we now rewrite \Cref{eq penal1} as:
\begin{equation}\label{eq penal2}
\begin{aligned}
\thetab &= \argmin_{\thetab} L\big(\zb^{\prime}(\xb;\thetab),\zb^{\prime(a)}(\xb;\thetab^{(a)})\big)
= \argmin_{\thetab} \bigg[L_C\big(\zb^{\prime}(\cdot),\zb^{\prime(a)}(\cdot)\big) \\
& + \omega_m \bigg(L_M\big(\zb^{\prime}(\cdot)\big) 
+ L^{(a)}_M\big(\zb^{\prime(a)}(\cdot),\zb^{\prime}_{\rho}(\cdot)\big)\bigg) 
+ \omega_t \bigg(L_T\big(z^{\prime}_t(\cdot),\zb^{\prime}_{\rho}(\cdot)\big)
+ L_T^{(a)}\big(z^{\prime(a)}_t(\cdot),\zb^{\prime}_{\rho}(\cdot)\big)\bigg) \\
& + \omega_{v} C_M^2\big(\zb_{\rho}(\cdot)\big)
+ \omega_{p} C_P^2\big(\zb_{\rho}(\cdot)\big)
\bigg].
\end{aligned}
\end{equation}

The objective function $L_C(\cdot)$ in \Cref{eq penal2} includes adjoint terms parameterized with $\zb^{\prime(a)}(\cdot)$ whose detailed formulation is provided in \Cref{subsec adj analysis}. In addition, the potential energy functionals $L^{(a)}_M(\cdot)$ and $L^{(a)}_T(\cdot)$ are incorporated into the unconstrained optimization to solve for the adjoint fields.

\subsection{Neural Representation of Mean Functions} \label{subsec neural mean}
Motivated by our earlier work \citep{yousefpour_IDETC_2025,sun_smo_2025}, we employ the PGCAN architecture \citep{shishehbor_parametric_2024} as the NN parameterization for the mean functions of all GPs. Our previous studies have demonstrated that PGCAN can capture high-frequency design features and dramatically accelerate convergence by suppressing undefined gray regions and reducing PDE residuals. 
This superior performance stems from PGCAN's unique encoder–decoder architecture:
(1) the encoder embeds the design domain in a grid-based feature space where learnable features from the grid vertices are nonlinearly interpolated to obtain feature values for any arbitrary query point inside the design domain, and  
(2) the encoded feature vector at a query point is passed to a decoder, which is composed of a shallow MLP with attention layers, to obtain the desired output (design variables, displacement, temperature, or their corresponding adjoint variables).

\begin{figure*}[!h]
    \centering
    \includegraphics[width = 0.95\textwidth]{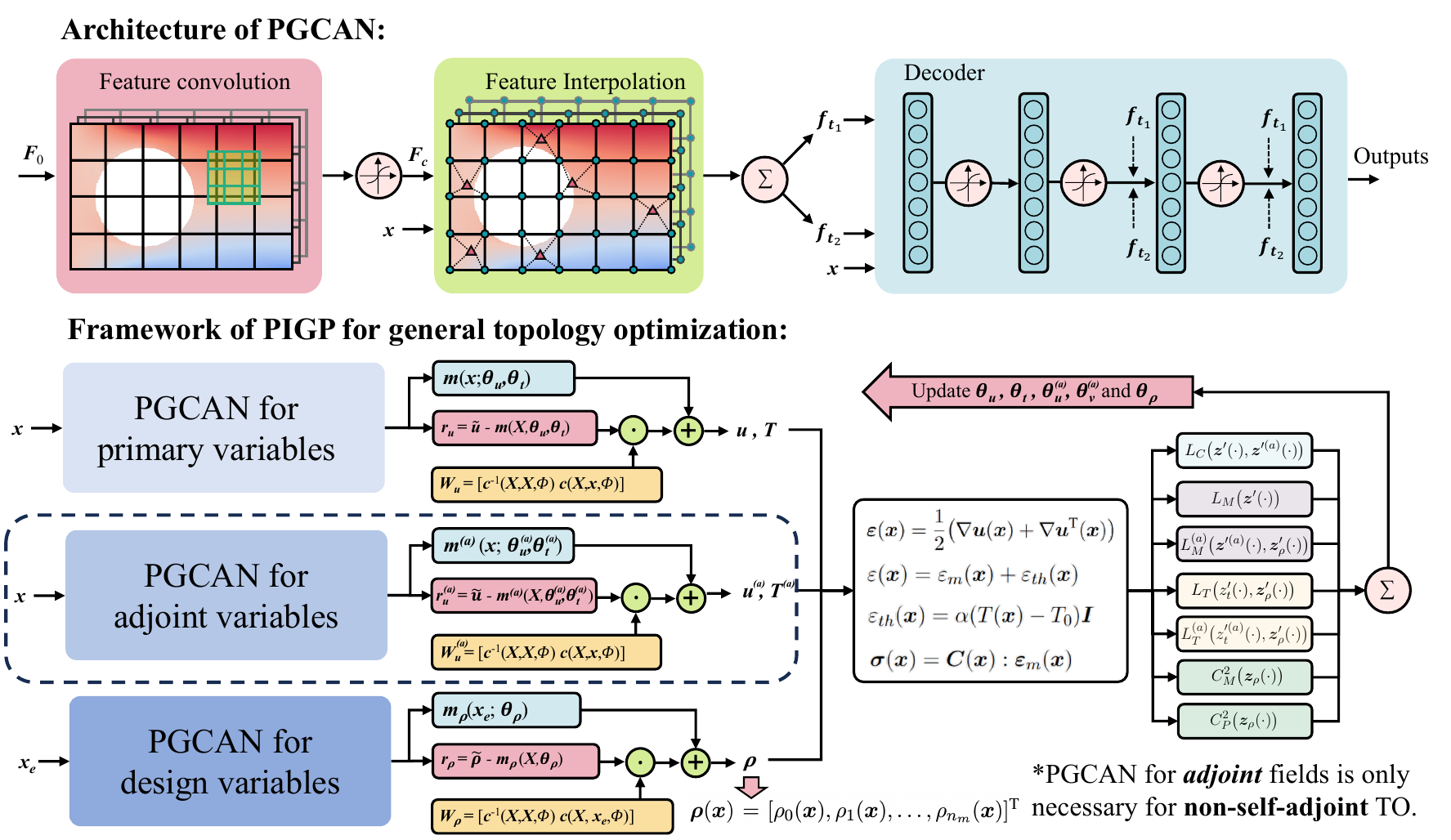}
\caption{\textbf{PIGP framework for multi-physics topology optimization:} The mean functions for the design variable, as well as for the primary and adjoint fields, are parameterized using the multi-output PGCAN architecture, which consists of three modules: (i) feature-space encoding via a convolutional neural network, (ii) feature interpolation at all collocation points, and (iii) decoding through a shallow MLP.
The design variable $\rhob(\xb)$ is a vector of volume fractions for the void and all candidate materials at each domain point $\xb$. The final-layer multi-output from PGCAN for $\rhob(\xb)$ is passed through a $\mathrm{softmax}$ function to enforce the partition of unity of $\rhob(\xb)$.
Gaussian processes are employed on these mean functions to impose the BCs for the primary and adjoint fields, and to enforce the design constraints. }
\label{fig pigp}
\end{figure*}

As illustrated in \Cref{fig pigp} for a 2D example, the design domain is parameterized through a trainable feature tensor $\Fb_0 \in \mathbb{R}^{N_{rep} \times N_f \times N_x^e \times N_y^e}$, where $N_{rep}$ is the number of grid repetitions with small diagonal offsets, $N_f$ is the number of features per grid vertex, and $N_x^e$ and $N_y^e$ denote the numbers of grid vertices along the $x$- and $y$-directions, respectively.  
For the 3D setting, an additional dimension $N_z^e$ is included so that $\Fb_0 \in \mathbb{R}^{N_{rep} \times N_f \times N_x^e \times N_y^e \times N_z^e}$.
To promote information flow across neighboring cells and mitigate overfitting, we apply a $3 \times 3$ convolution with $N_f$ input and output channels to $\Fb_0$, followed by a $\tanh$ activation function, producing the feature map $\Fb_c \in \mathbb{R}^{N_{rep} \times N_f \times N_x^e \times N_y^e}$, which has the same dimensions as $\Fb_0$.

To compute the feature vector at an arbitrary query point $\xb$, we first determine its local (i.e., within-cell) coordinates $\bar{\xb} \in [0,1]^2$ by normalizing the global coordinates and then apply a cosine transformation as:
\begin{equation}\label{eq cosine}
    \xb^{(c)} = \frac{1}{2}\big(1 - \cos(\pi\bar{ \xb})\big).
\end{equation}
For each grid repetition, the feature vector at $\xb$ is obtained via bilinear interpolation using the features at the vertices of the cell enclosing $\xb^{(c)}$:
\begin{equation}\label{eq interpolation}
\fb_{\tb}^m(\xb) = (1 - x^{(c)})(1 - y^{(c)}) \fb^m_{\tb(0,0)} + (1 - x^{(c)}) y^{(c)} \fb^m_{\tb(0,1)} + x^{(c)}(1 - y^{(c)}) \fb^m_{\tb(1,0)} + x^{(c)} y^{(c)} \fb^m_{\tb(1,1)},
\end{equation}
where $\fb^m_{\tb(i,j)} \in \mathbb{R}^{N_f}$ for $i,j \in \{0,1\}$ are the features extracted from $\Fb_c$, and $m \in \{1,\dots,N_{rep}\}$ indexes the grid repetitions. Because the encoder uses only the features located at the vertices of the enclosing cell, it enables PGCAN to effectively mitigate spectral bias and accurately represent high-gradient solutions.
The final feature vector at the query point is computed by summing the contributions across all grid repetitions, i.e., $\fb_{\tb}(\xb) = \sum_{m}\fb_{\tb}^m(\xb)$. 
Once $\fb_{\tb}(\xb)$ is obtained, it is split into two equal-length vectors, $\fb_{\tb_1}(\xb)$ and $\fb_{\tb_2}(\xb)$, and passed to a shallow decoder network with three hidden layers. From \cite{shishehbor_parametric_2024}, these features sequentially modulate the hidden states of the decoder, thereby improving gradient propagation during training.

As shown in \Cref{fig pigp}, we represent the multi-material design at each spatial point $\xb$ by the vector-valued design variable $\rhob(\xb)=[\rho_0(\xb),\rho_1(\xb),\ldots,\rho_{n_m}(\xb)]^{\mathsf T}$. The PGCAN for $\rhob(\xb)$ produces a multi-output field, and we apply a $\mathrm{softmax}$ mapping to the final-layer outputs to obtain nonnegative fractions that satisfy the partition-of-unity constraint.
To further enhance robustness and accelerate convergence of the thermal field in the coupled multi-physics setting, we also adjust the bias of the final PGCAN layer for temperature before the training, so that the initial mean temperature is consistent with the imposed thermal BCs. This initialization reduces early-stage imbalances and accelerates thermal equilibrium enforcement during optimization. 

\subsection{Numerical Approximation} \label{subsec method sf}
Multiple terms in \Cref{eq penal2} involve gradient and integration operations which must be calculated in each optimization iteration. Although automatic differentiation (AD) provides a convenient mechanism for computing derivatives, it is computationally expensive and becomes impractical for large-scale problems. Numerical schemes such as finite difference have been successfully used in place of AD \citep{sun_smo_2025, yousefpour_simultaneous_2025} to reduce the costs but FD may provide insufficient robustness or accuracy, particularly near domain boundaries.

To improve cost and accuracy we draw inspiration from the FEM \citep{hughes_FEM_2012} as follows. In our framework the solution and design variables can be evaluated at any arbitrary point in the domain. So, we choose a set of CPs such that they coincide with the FE nodes of a pre-selected element type where “element’’ refers to the subdomain $\Omega^{(e)}$ spanned by a small group of neighboring CPs. Within each element, the displacement field is interpolated via shape functions as:
\begin{equation}\label{eq sf 1}
    u_i^{(e)}(\xb) = \sum_{j=1}^{n_s}N_j\big(\xib^{(e)}(\xb)\big) u_i(\xb^{(e)}_j;\thetab_u,\phib_u), \quad i=1,\dots,n_d, \quad e = 1,\dots,n_e,\quad \xb\in\Omega^{(e)}
\end{equation}
where $n_s$ is the number of shape functions per element, and $n_e$ is the total number of elements in the design domain, $N_j\big(\xib^{(e)}(\xb)\big)$ denotes the $j$-th shape function evaluated at the mapped local coordinate $\xib^{(e)}(\xb):=[\xi^{(e)}_1(\xb),\dots,\xi^{(e)}_{n_d}(\xb)]^{\text{T}}$, and $\xb^{(e)}_j=[x^{(e)}_{1,j},\dots,x^{(e)}_{n_d,j}]^{\text{T}}$ is the global coordinates of the $j$-th node. 

In this work we adopt bilinear quadrilateral elements in 2D and trilinear hexahedral elements in 3D. The 2D shape functions are:
\begin{subequations} \label{eq sf 2}
    \begin{align}
    N_1\big(\xib^{(e)}(\xb)\big)
    &:= \frac{1}{4}\bigg(1-\xi^{(e)}_1(\xb)\bigg)\bigg(1-\xi^{(e)}_2(\xb)\bigg), \label{eq sf 1 a} \\
    N_2\big(\xib^{(e)}(\xb)\big)
    &:= \frac{1}{4}\bigg(1+\xi^{(e)}_1(\xb)\bigg)\bigg(1-\xi^{(e)}_2(\xb)\bigg), \label{eq sf 1 b} \\
    N_3\big(\xib^{(e)}(\xb)\big)
    &:= \frac{1}{4}\bigg(1+\xi^{(e)}_1(\xb)\bigg)\bigg(1+\xi^{(e)}_2(\xb)\bigg), \label{eq sf 1 c} \\
    N_4\big(\xib^{(e)}(\xb)\big)
    &:= \frac{1}{4}\bigg(1-\xi^{(e)}_1(\xb)\bigg)\bigg(1+\xi^{(e)}_2(\xb)\bigg). \label{eq sf 1 d}
    \end{align}
\end{subequations}
Other shape function families may also be used and their expressions follow forms analogous to \Cref{eq sf 2}. From \Cref{eq sf 1}, the gradient with respect to the local coordinate $\xib^{(e)}$ is written as:
\begin{equation}\label{eq sf grad 1}
 \frac{\partial u_i^{(e)}(\xb)}{\partial \xib^{(e)}} = \sum_{j=1}^{n_s}\frac{\partial N_j\big(\cdot\big)}{\partial \xib^{(e)}} u_i(\xb^{(e)}_j;\thetab_u,\phib_u), \quad i=1,\dots,n_d,
\end{equation}
where $\partial u_i^{(e)}(\xb)/\partial \xib^{(e)}$ is a column vector. Computing gradients in global coordinates $\partial u_i^{(e)}(\xb)/\partial\xb$ requires the mapping function $\xib^{(e)}(\xb)$, which is difficult to obtain explicitly. Instead, we adopt the isoparametric mapping from $\xib^{(e)}$ to $\xb$ using the same shape functions:
\begin{equation}\label{eq sf map}
    \xb\big(\xib^{(e)}\big) = \sum_{j=1}^{n_s}N_j\big(\xib^{(e)}\big) \xb^{(e)}_j,\quad \xb\in\Omega^{(e)},
\end{equation}
and further define the Jacobian as:
\begin{equation}\label{eq sf J 1}
    \Jb\big(\xib^{(e)}\big) = \frac{\partial\xb}{\partial\xib^{(e)}},
\end{equation}
which has the matrix form explicitly written as:
\begin{equation}\label{eq sf J 2}
    \Jb\big(\xib^{(e)}\big)
    =
    \begin{bmatrix}
        \displaystyle\sum_{j=1}^{n_s}\dfrac{\partial N_j}{\partial \xi^{(e)}_1} x^{(e)}_{1,j} 
        & \cdots &
        \displaystyle\sum_{j=1}^{n_s}\dfrac{\partial N_j}{\partial \xi^{(e)}_{n_d}} x^{(e)}_{1,j} \\[2ex]
        \vdots & \ddots & \vdots \\[1ex]
        \displaystyle\sum_{j=1}^{n_s}\dfrac{\partial N_j}{\partial \xi^{(e)}_1} x^{(e)}_{n_d,j}
        & \cdots &
        \displaystyle\sum_{j=1}^{n_s}\dfrac{\partial N_j}{\partial \xi^{(e)}_{n_d}} x^{(e)}_{n_d,j}
    \end{bmatrix}.
\end{equation}
Applying the chain rule to any displacement component $u_i^{(e)}(\xb)$, we can have the gradient relation:
\begin{equation}\label{eq sf grad 2}
    \frac{\partial u_i^{(e)}(\xb)}{\partial \xib^{(e)}} = \Jb^{\text{T}}\big(\xib^{(e)}\big)\frac{\partial u_i^{(e)}(\xb)}{\partial \xb},
\end{equation}
where $\partial u_i^{(e)}(\cdot)/\partial \xb$ are the gradient functions in the form of a column vector. 
Combining \Cref{eq sf grad 1} and \Cref{eq sf grad 2}, we can estimate the gradient of displacement in a specific element as:
\begin{equation}\label{eq sf grad 3}
 \frac{\partial u_i^{(e)}(\xb)}{\partial \xb} = \Jb^{-\text{T}}\big(\xib^{(e)}\big) \sum_{j=1}^{n_s}\frac{\partial N_j\big(\cdot\big)}{\partial \xib^{(e)}} u_i(\xb^{(e)}_j;\thetab_u,\phib_u), \quad i=1,\dots,n_d.
\end{equation}
During the implementation, we follow a standard FEM practice where a gradient matrix $\Bb_u(\xib^{(e)})$ is assembled using \Cref{eq kinematic,eq sf grad 3} and cached for subsequent optimization iterations. The strain vector $\epsilonb^{(e)}(\xb)=[\varepsilon^{(e)}_1,\dots,\varepsilon^{(e)}_{n_{sr}}]^{\text{T}}$ is then obtained as:
\begin{equation}\label{eq sf strain}
 \epsilonb^{(e)}(\xb) = \Bb_u(\xib^{(e)})\ub^{(e)},
\end{equation}
where $n_{sr}$ denotes the number of \textit{independent} strain or stress components and $\ub^{(e)} = [u_1(\Xb^{(e)};\thetab_u,\phib_u),\dots,u_{n_d}(\Xb^{(e)};\thetab_u,\phib_u)]^{\text{T}}$ is the stacked displacement vector with $n_r=n_s\times n_d$ entries and $\Xb^{(e)}=\{\xb^{(e)}_1,\dots,\xb^{(e)}_{n_s}\}$ being the set of element nodes. Similarly, the temperature gradient $\partial T^{(e)}(\xb)/\partial \xb$ is computed as
\begin{equation}\label{eq sf dT}
 \frac{\partial T^{(e)}(\xb)}{\partial \xb} = \Bb_T(\xib^{(e)})\Tb^{(e)},
\end{equation}
where $\Bb_T(\xib^{(e)})$ is the temperature gradient matrix and $\Tb^{(e)} = T(\Xb^{(e)};\thetab_t,\phib_t)$ represents the vector of nodal temperatures associated with the element. This shape-function-based approach is highly efficient for both forward evaluations and backward parameter updates, as gradient computations reduce to cached matrix–vector multiplications that are well optimized on GPUs.

Once the gradients of the primary and adjoint variables are obtained, each loss term on the RHS of \Cref{eq penal2} is evaluated using Gauss integration. In \Cref{eq sf strain,eq sf dT} we take $\xb$ as the weighted center of each element with corresponding local coordinate $\xib^{(e)*}=[0,\dots,0]^{\text{T}}$. The internal energy integral is approximated as:
\begin{equation}\label{eq sf int}
\int_{\Omega} \sigmab(\xb) : \epsilonb(\xb) \, dV = \sum_{e=1}^{n_e} \omega_g \sigmab^{(e)\text{T}}\big(\xib^{(e)*}\big)\epsilonb^{(e)}\big(\xib^{(e)*}\big)\det\Jb\big(\xib^{(e)*}\big),
\end{equation}
where $\omega_g=4$ is the Gauss weight and the shear strains in $\epsilonb^{(e)}$ represent true shear strains. This corresponds to the reduced integration scheme in FEM, which we find substantially more efficient than full integration with multiple Gauss points. Based on our observation, full integration yields negligible accuracy gains while incurring significantly higher computational cost. The estimations of other integrals have the similar form as shown in \Cref{eq sf int}. To estimate integrals during TO, the design variables $\rhob(\xb)$ as well as the distribution of material parameters such as $\Cb(\xb)$, $\kappa(\xb)$, and $\alpha(\xb)$, etc., are all estimated on the weighted center of each element.

We highlight that although our framework incorporates concepts from FEM such as the elements, shape functions, and Gauss integration, it fundamentally differs from traditional FEM-based TO (e.g., SIMP). In our approach, the primary fields, adjoint fields, and design variables are represented by continuous parameterized functions which are evaluated at a finite number of CPs for gradient and integral calculations. As shown in our previous works \citep{sun_smo_2025}, the spatial distribution of these CPs do not affect the designed topology as long as they yield sufficient accuracy for estimating gradients and integrals. That is, once a sufficient number of CPs is used, further increasing CP count does not affect the optimized topology (note that the cost would expectedly go up by using more CPs and hence we aim to use the smallest number of CPs that provides sufficient accuracy). In contrast to our approach, FEM-based TO directly solves for discrete nodal unknowns and therefore requires additional mechanisms (e.g., filtering) to mitigate checkerboard patterns and mesh dependency issues. 

\subsection{Stability and Curriculum Training} \label{subsec method training}
To efficiently estimate spatial derivatives using shape functions, we can place the CPs on a fixed FE grid that remains unchanged throughout the optimization. This setup enables us to pre-compute and store the covariance matrices (and their inverses) appearing in \Cref{eq GP u,eq GP t,eq GP rho} for the primary fields and design variables, as well as in \Cref{eq GP u adj,eq GP t adj} for the adjoint fields. By eliminating repeated matrix inversions, this strategy significantly reduces computational cost and enhances numerical robustness.

However, combining a fixed FE grid with the PGCAN architecture can introduce overfitting \citep{sun_smo_2025}. To address this issue, we adopt a multi-grid approach where each grid is randomly selected at an optimization iteration. Specifically, we first mesh the design domain (e.g., via a commercial meshing tool) and define a coarse and fine element size where the latter produces twice the number of elements than the former. The intermediate FE grids are then generated by linearly interpolating the element size between these two extremes, forming a set of all FE grids denoted by $\boldsymbol{\Sigma} = [\Xb_1,\dots,\Xb_{n_g}]$, where $\Xb_1$ and $\Xb_{n_g}$ correspond to the coarse and fine grids, respectively. For each grid in $\boldsymbol{\Sigma}$, we cache all relevant geometric and structural information, including CP coordinates, element-wise weighted center coordinates, node–element connectivity, and the gradient matrix. 

To further increase the stability of the minimization problem in \Cref{eq penal2} we leverage curriculum training \citep{soviany_curriculum_2022}. Without such strategies, the training can exhibit abrupt oscillations in the loss histories, often resulting in rough boundaries and suboptimal topologies as reported in \citep{sun_smo_2025}. 
For multi-material and multi-physics problems studied in our works, we have found it useful to apply the curriculum schedule to $\psi_m$ and $\psi_p$.  Taking the target mass fraction as an example, it can be scheduled as:
\begin{equation}\label{eq vf schedule}
\psi_m =
\begin{cases}
\displaystyle \frac{\psi_{ml} - \psi_{m0}}{\gamma n_{tol}}\, n + \psi_{m0}, & n \leq \gamma n_{tol}, \\
\psi_{ml}, &  n > \gamma n_{tol},
\end{cases}
\end{equation}
where $\psi_{m0}$ and $\psi_{ml}$ denote the initial and final mass fractions, $n_{tol}$ is the total number of training epochs, and $\gamma = 50\%$ controls the ramping interval. The initial mass fraction $\psi_{m0}$ is estimated from the design variable distribution from the first iteration before training. 
By gradually enforcing the mass and cost constraints, the optimization undergoes smoother topological transitions, leading to improved training stability. 

\subsection{Continuous Adjoint Analysis} \label{subsec adj analysis}
The conventional adjoint-based sensitivity analysis developed for FE formulations is not directly applicable to our mesh-free approach. In this section, we derive the sensitivity of a general objective function $f_{obj}(\cdot)$ with respect to the design field $\rhob(\cdot)$ using the continuous adjoint method. This approach (1) provides meaningful gradient information for topology updates, (2) accommodates both self-adjoint and non-self-adjoint systems through appropriately constructed boundary value problems, and (3) offers analytical adjoint formulations suitable for multi-physics coupling.

The design variables in our multi-material TO framework are denoted as $\rhob(\xb) = [\rho_0(\xb), \dots, \rho_{n_m}(\xb)]^{\text{T}}$, and the sensitivity must be evaluated for each component $\rho_i(\xb),\,i\in 0,\dots,n_m$. For a linear elastic mechanical system without body forces, we introduce the augmented functional $L^{(a)}_C(\cdot)$ as follows:
\begin{equation}\label{eq adjoint 1}
    L^{(a)}_C(\cdot) = f_{obj}(\cdot) + \alpha_u\int_{\Omega} \vb_{u}(\xb) \cdot \big(\nabla \cdot \sigmab(\xb)\big)\, dV,
\end{equation}
where $\vb_{\ub}(\xb)$ is the adjoint displacement field, which can be chosen arbitrarily, and $\alpha_u$ is a weighting factor that controls the contribution of the adjoint term. Applying the divergence theorem to \Cref{eq adjoint 1} yields:
\begin{equation}\label{eq adjoint 2}
    L^{(a)}_C(\cdot) = f_{obj}(\cdot) 
    - \alpha_u \bigg(\int_{\Omega} \sigmab(\xb) : \nabla\vb_{\ub}(\xb)\, dV 
    - \int_{\partial\Omega} \vb_{\ub}(\xb)\cdot \big( \sigmab(\xb)\nb\big)\, ds\bigg).
\end{equation}
By differentiating both sides of \Cref{eq adjoint 2} and using the constitutive relation in \Cref{eq constitutive}, we have:
\begin{equation}\label{eq adjoint 3}
\begin{split}
    \frac{d L^{(a)}_C(\cdot)}{d\rho_i} 
    &= \frac{d f_{obj}(\cdot)}{d\rho_i} 
    - \alpha_u \bigg(\int_{\Omega} \big(\frac{\partial\Cb(\xb)}{\partial\rho_i} : \nabla \ub(\xb)\big):\nabla\vb_{\ub}(\xb)\, dV \\
    &\quad 
    + \int_{\Omega} \big(\Cb(\xb):\nabla(\tfrac{\partial\ub}{\partial\rho_i})\big):\nabla \vb_{\ub}(\xb)\, dV 
    - \int_{\partial\Omega} \vb_{\ub}(\xb)\cdot \bigg(\frac{d\sigmab(\xb)}{d\rho_i}\nb\bigg)\, ds\bigg).
\end{split}
\end{equation}
Using the symmetry of $\Cb(\xb)$, the second term can be rewritten as:
\begin{equation}\label{eq adjoint 4}
    \int_{\Omega} \big(\Cb(\xb):\nabla(\frac{\partial\ub}{\partial\rho_i})\big):\nabla \vb_{\ub}(\xb)\,dV = \int_{\Omega} \sigmab^{\vb}(\xb):\nabla(\frac{\partial\ub}{\partial\rho_i})\,dV
\end{equation}
where $\sigmab^{\vb}(\xb)=\Cb(\xb):\nabla \vb_{\ub}(\xb)$ denotes the stress field associated with $\vb_{\ub}(\xb)$. Applying the divergence theorem again:
\begin{equation}\label{eq adjoint 5}
    \int_{\Omega} \sigmab^{\vb}(\xb):\nabla(\frac{\partial\ub}{\partial\rho_i})\,dV = -\int_{\Omega}\frac{\partial\ub}{\partial\rho_i}\cdot\big(\nabla\cdot\sigmab^{\vb}(\xb)\big)\,dV + \int_{\partial\Omega}\frac{\partial\ub}{\partial\rho_i}\cdot\big(\sigmab^{\vb}(\xb)\nb\big)\,ds,
\end{equation}
Since $\vb_{\ub}(\xb)$ is arbitrary, we choose it to satisfy the same governing PDEs as in \Cref{eq pde}, leading to the adjoint equations:
\begin{equation}\label{eq adjoint pde}
    \nabla \cdot \sigmab^{\vb}(\xb) = 0, \quad \forall \xb \in \Omega.
\end{equation}
Using \Cref{eq adjoint 4,eq adjoint 5,eq adjoint pde}, we can simplify \Cref{eq adjoint 3} given as:
\begin{equation}\label{eq adjoint 6}
\begin{split}
    \frac{d L^{(a)}_C(\cdot)}{d\rho_i} 
    &= \frac{d f_{obj}(\cdot)}{d\rho_i} 
    - \alpha_u \bigg( \int_{\Omega} \big(\frac{\partial\Cb(\xb)}{\partial\rho_i} : \nabla \ub(\xb)\big):\nabla\vb_{\ub}(\xb) \, dV \\
    &\quad 
    + \int_{\partial\Omega}\frac{\partial\ub}{\partial\rho_i}\cdot\big(\sigmab^{\vb}(\xb)\nb\big)\,ds 
    - \int_{\partial\Omega} \vb_{\ub}(\xb) \cdot \big(\frac{d\sigmab(\xb)}{d\rho_i}\nb \big)\, ds\bigg).
\end{split}
\end{equation}
Depending on the design objective and the prescribed BCs, we derive the corresponding adjoint equations for (1) CM, (2) compliant mechanism design, (3) heat sink optimization, and (4) multi-physics compliant design with thermo-mechanical coupling. 

\subsubsection{Compliance Minimization} \label{subsubsec exam comp}
For CM, the objective function is defined as a volume integral of internal energy density:
\begin{equation}\label{eq comp obj}
 f_{obj}(\cdot) = \int_{\Omega}\big(\Cb(\xb):\nabla\ub(\xb)\big):\nabla\ub(\xb)\,dV.
\end{equation}
Substituting this expression into \Cref{eq adjoint 6} yields:
\begin{equation}\label{eq comp 1}
\begin{split}
    \frac{d L^{(a)}_C(\cdot)}{d\rho_i} 
    & = \int_{\Omega}\big(\frac{\partial\Cb(\xb)}{\partial\rho_i}:\nabla\ub(\xb)\big):\nabla\big(\ub(\xb) - \alpha_u \vb_{\ub}(\xb)\big)\,dV 
    + 2 \int_{\Omega} \sigmab(\xb):\nabla(\frac{\partial\ub}{\partial\rho_i}) \, dV \\
    &\quad 
    - \alpha_u \bigg(\int_{\partial\Omega}\frac{\partial\ub}{\partial\rho_i}\cdot\big(\sigmab^{\vb}(\xb)\nb\big)\,ds 
    - \int_{\partial\Omega} \vb_{\ub}(\xb) \cdot \big(\frac{d\sigmab(\xb)}{d\rho_i}\nb \big)\, ds\bigg).
\end{split}
\end{equation}
Similar to \Cref{eq adjoint 5}, application of the divergence theorem gives the relation:
\begin{equation}\label{eq comp 2}
    \int_{\Omega} \sigmab(\xb):\nabla(\frac{\partial\ub}{\partial\rho_i})\,dV =\int_{\partial\Omega}\frac{\partial\ub}{\partial\rho_i}\cdot\big(\sigmab(\xb)\nb\big)\,ds.
\end{equation}
Using \Cref{eq comp 2}, \Cref{eq comp 1} simplifies to:
\begin{equation}\label{eq comp 3}
\begin{split}
    \frac{d L^{(a)}_C(\cdot)}{d\rho_i} 
    & = \int_{\Omega}\big(\frac{\partial\Cb(\xb)}{\partial\rho_i}:\nabla\ub(\xb)\big):\nabla\big(\ub(\xb) - \alpha_u \vb_{\ub}(\xb)\big)\,dV  \\
    &\quad 
    - \int_{\partial\Omega}\frac{\partial\ub}{\partial\rho_i}\cdot\big((\alpha_u\sigmab^{\vb}(\xb) - 2\sigmab(\xb))\nb\big)\,ds 
    + \alpha_u \int_{\partial\Omega} \vb_{\ub}(\xb) \cdot \big(\frac{d\sigmab(\xb)}{d\rho_i}\nb \big)\, ds.
\end{split}
\end{equation}
For CM, no spring supports are present on the domain boundary. We therefore choose $\vb_{\ub}(\xb)$ to satisfy homogeneous BCs on $\partial\Omega_h$, which eliminates the last boundary term in \Cref{eq comp 3}. Moreover, by selecting the adjoint field such that $\alpha_u \vb_{\ub}(\xb) = 2\ub(\xb)$, the term involving $\partial\ub/\partial\rho_i$ also vanishes. The final sensitivity expression becomes:
\begin{equation}\label{eq comp 4}
    \frac{d L^{(a)}_C(\cdot)}{d\rho} = -\alpha_u\int_{\Omega}\big(\frac{\partial\Cb(\xb)}{\partial\rho}:\nabla\ub(\xb)\big):\nabla\ub(\xb)\,dV.
\end{equation}
This result shows that the gradient does not depend on the adjoint field $\vb_{\ub}(\xb)$, indicating that CM is a self-adjoint problem and does not require solving the separate adjoint PDEs. We also note that \Cref{eq comp 4} is consistent with our previous formulation in \citep{sun_smo_2025} by choosing $\alpha_u = 1$.

\subsubsection{Compliant Mechanism Design} \label{subsubsec exam cmpt}
For compliant mechanism design, the design objective is to \textit{maximize} a prescribed displacement component at a specific boundary point $\xb_{out}\in\partial\Omega$. Accordingly, the continuous objective to be minimized can be written as the negative of a surface integral:
\begin{equation}\label{eq cmpt obj}
 f_{obj}(\cdot) = -u_{out} = -\int_{\partial\Omega}\ub(\xb)\cdot\big(\delta(\xb - \xb_{out})\eb_n\big)\,ds.
\end{equation}
where $\eb_n$ is denotes the desired displacement direction. Substituting this expression into \Cref{eq adjoint 6}, we have:
\begin{equation}\label{eq cmpt 1}
\begin{split}
    \frac{d L^{(a)}_C(\cdot)}{d\rho_i} 
    &= - \int_{\partial\Omega}\frac{\partial\ub}{\partial\rho_i}\cdot\big(\delta(\xb - \xb_{out})\eb_n\big)\,ds - \alpha_u \bigg( \int_{\Omega} \big(\frac{\partial\Cb(\xb)}{\partial\rho_i} : \nabla \ub(\xb)\big):\nabla\vb_{\ub}(\xb) \,dV 
    \\
    &\quad 
    + \int_{\partial\Omega}\frac{\partial\ub}{\partial\rho_i}\cdot\big(\sigmab^{\vb}(\xb)\nb\big)\,ds  
    - \int_{\partial\Omega} \vb_{\ub}(\xb) \cdot \big(\frac{d\sigmab(\xb)}{d\rho_i}\nb \big)\, ds\bigg).
\end{split}
\end{equation}
To construct the adjoint problem consistent with \Cref{eq bc}, we choose $\vb_{\ub}(\xb)$ to satisfy the \textit{same} homogeneous BCs and free-surface conditions as $\ub(\xb)$. We also enforce $\sigmab^{\vb}(\xb)\nb = 0$ on $\partial\Omega_t$. Under these choices, the last two terms of \Cref{eq cmpt 1} vanish on $\partial\Omega_h$, $\partial\Omega_0$, and $\partial\Omega_t$.
For the spring-supported points $\xb_s$ defined in \Cref{eq bc 4}, the final boundary term can be written as:
\begin{equation}\label{eq cmpt 2}
 \int_{\partial\Omega_s} \vb_{\ub}(\xb) \cdot \big(\frac{d\sigmab(\xb)}{d\rho_i}\nb \big)\, ds = -\int_{\partial\Omega_s} \frac{\partial\ub}{\partial\rho_i} \cdot K_s \delta(\xb - \xb_s)\big(\vb_{\ub}(\xb)\cdot\eb_s\big)\eb_s\, ds .
\end{equation}
Using \Cref{eq cmpt 2} and the imposed conditions on $\vb_{\ub}(\xb)$, the augmented sensitivity expression becomes:
\begin{equation}\label{eq cmpt 3}
\begin{split}
    \frac{d L^{(a)}_C(\cdot)}{d\rho_i} 
    &= - \int_{\partial\Omega}\frac{\partial\ub}{\partial\rho_i}\cdot\big(\alpha_u\sigmab^{\vb}(\xb)\nb + \delta(\xb - \xb_{out})\eb_n \big)\,ds 
    - \alpha_u \bigg(\int_{\Omega} \big(\frac{\partial\Cb(\xb)}{\partial\rho_i} : \nabla \ub(\xb)\big):\nabla\vb_{\ub}(\xb) \,dV\\
    &\quad 
    + \int_{\partial\Omega_s}\frac{\partial\ub}{\partial\rho_i}\cdot\bigg[\sigmab^{\vb}(\xb)\nb +  K_s \delta(\xb - \xb_s)\big(\vb_{\ub}(\xb)\cdot\eb_s\big)\eb_s \bigg]\, ds\bigg).
\end{split}
\end{equation}
We now select the traction BCs for $\sigmab^{\vb}(\xb)\nb$ on $\partial\Omega_s$ and $\partial\Omega$ such that the two boundary integrals in \Cref{eq cmpt 3} disappear, and the final sensitivity expression reduces to:
\begin{equation}\label{eq cmpt 4}
    \frac{d L^{(a)}_C(\cdot)}{d\rho} = - \alpha_u \int_{\Omega} \big(\frac{\partial\Cb(\xb)}{\partial\rho} : \nabla \ub(\xb)\big):\nabla\vb_{\ub}(\xb) \,dV 
\end{equation}
where the adjoint field $\vb_{\ub}(\xb)$ is determined from the following BCs:
\begin{subequations} \label{eq bc adj}
    \begin{align}
    \vb_{\ub}(\xb) &= 0, \quad \forall \xb \in \partial \Omega_{\hb}, \label{eq bc adj 1} \\
    \sigmab^{\vb}(\xb) \cdot \nb(\xb) &= 0, \quad \forall \xb \in \partial \Omega_t, \label{eq bc adj 2}\\
    \sigmab^{\vb}(\xb) \cdot \nb(\xb) &= 0, \quad \forall \xb \in \partial \Omega_0, \label{eq bc adj 3}\\
    \sigmab^{\vb}(\xb) \cdot \nb(\xb) &= -K_s\delta(\xb - \xb_s)\big(\ub(\xb)\cdot\eb_s\big)\eb_s, \quad \forall \xb \in \partial \Omega_s, \label{eq bc adj 4}\\
    \sigmab^{\vb}(\xb) \cdot \nb(\xb) &= -\frac{1}{\alpha_u}\delta(\xb - \xb_{out})\eb_n, \quad \forall \xb \in \partial \Omega. \label{eq bc adj 5}
    \end{align}
\end{subequations}
We observe that $\vb_{\ub}(\xb)$ shares the same boundary regions as $\ub(\xb)$ for the homogeneous, free-surface, and spring supports. However, unlike $\ub(\xb)$, the traction BC on $\partial\Omega_t$ vanishes for $\vb_{\ub}(\xb)$ and a new traction condition is imposed on $\partial\Omega$ to account for the objective functional. The magnitude of the applied external traction is scaled by $1/\alpha_u$ due to the weighting factor introduced in the formulation. Adjusting $\alpha_u$ affects both the convergence behavior and the stability of the optimization. In our implementation, we found $\alpha_u = 10$ to provide the most stable training and the best overall performance.
In general, compliant mechanism design is not a self-adjoint problem and therefore $\vb_{\ub}(\xb)$ must be solved explicitly.

\subsubsection{Heat Sink Optimization} \label{subsubsec exam heat}
We consider homogeneous temperature BCs together with zero external heat flux ($q(\xb)=0$) and no thermal convection ($h_v(\xb)=0$). Under these assumptions, the augmented objective functional is written as:
\begin{equation}\label{eq adjoint heat 1}
    L^{(a)}_C\big(T(\xb,\rhob)\big) = \frac{1}{2} \int_{\Omega} \kappa(\xb) |\nabla T(\xb,\rhob)|^2 \, dV + \alpha_T\int_{\Omega} v_T(\xb,\rhob)\big(\nabla\cdot (\kappa(\xb)\nabla T(\xb,\rhob)) + s\big) \, dV,
\end{equation}
where $v_T(\xb)$ is the adjoint temperature field and the heat source term $s$ is taken as a constant. $\alpha_T$ is a weighting factor that adjusts the contribution of the added thermal term. Applying the divergence theorem yields:
\begin{equation}\label{eq adjoint heat 2}
\begin{split}
    L^{(a)}_C\big(T(\xb,\rhob)\big) 
    &= \frac{1}{2} \int_{\Omega} \kappa(\xb) |\nabla T(\xb,\rhob)|^2 \, dV 
    + \alpha_u \bigg( \int_{\partial\Omega} v_T(\xb)\kappa(\xb)\nabla T(\xb,\rhob)\cdot \nb \, ds\\
    &\quad 
    -\int_{\Omega} \kappa(\xb) \nabla v_T(\xb) \cdot \nabla T(\xb,\rhob) \, dV + \int_{\Omega} v_T(\xb)s \, dV\bigg).
\end{split}
\end{equation}
We choose $v_T(\xb)$ to satisfy the same homogeneous BCs as in \Cref{eq bc heat 2}. Additionally, since the other boundary regions are thermally insulated, the boundary term in \Cref{eq adjoint heat 2} vanishes. Differentiating $L_C(\cdot)$ with respect to $\rho_i(\xb)$ yields:
\begin{equation}\label{eq adjoint heat 3}
\begin{split}
    \frac{d L^{(a)}_C(\cdot) }{d \rho_i} 
    &= \frac{1}{2} \int_{\Omega} \frac{\partial\kappa(\xb)}{\partial\rho_i} \nabla T(\xb,\rhob) \cdot \nabla T(\xb,\rhob) \, dV 
    + \int_{\Omega} \kappa(\cdot) \nabla T(\xb,\rhob) \cdot \nabla \frac{\partial T}{\partial\rho_i} \, dV\\
    &\quad 
    -\alpha_T\bigg(\int_{\Omega} \frac{\partial\kappa(\xb)}{\partial\rho_i} \nabla v_T(\xb) \cdot \nabla T(\xb,\rhob) \, dV 
    + \int_{\Omega} \kappa(\cdot) \nabla v_T(\xb) \cdot \nabla \frac{\partial T}{\partial\rho_i} \, dV\bigg).
\end{split}
\end{equation}
By choosing the adjoint field as $\alpha_T v_T(\xb) = T(\xb,\rhob)$, the sensitivity expression above simplifies to:
\begin{equation}\label{eq adjoint heat 4}
    \frac{d L^{(a)}_C(\cdot) }{d \rho_i} = -\frac{1}{2} \int_{\Omega} \frac{\partial\kappa(\xb)}{\partial\rho_i} \nabla T(\xb,\rhob) \cdot \nabla T(\xb,\rhob) \, dV.
\end{equation}
Since the adjoint field $v_T(\xb)$ is directly obtained from $T(\xb,\rhob)$ and does not require solving an additional PDE system, the heat transfer optimization problem is self-adjoint, similar to CM.

\subsubsection{Thermo-mechanical Design} \label{subsubsec exam coup}
As in the compliant mechanism design, the objective is to maximize the displacement at a particular location of a component as defined in \Cref{eq cmpt obj}. Instead of relying on externally applied forces, thermo-mechanical compliant design is driven by thermal expansion or contraction resulting from the heat transfer within the design domain. The augmented objective functional $L^{(a)}_C\big(\ub(\xb),T(\xb,\rhob)\big)$ is written as:
\begin{equation}\label{eq coup obj 1}
\begin{aligned}
L^{(a)}_C(\cdot) 
&= 
-\int_{\partial\Omega}\ub(\xb)\cdot\big(\delta(\xb - \xb_{out})\eb_n\big)\,ds + \alpha_u \int_{\Omega} \vb_{\ub}(\xb)\big(\nabla\cdot \sigmab\big) \, dV   \\
&\quad
+ \alpha_T\int_{\Omega} v_T(\xb)
\Big[\nabla\cdot \big(\kappa(\xb)\nabla T(\xb,\rhob)\big)
- h_v \big(T(\xb,\rhob) - T_\infty\big)
+ s(\xb) \Big]\, dV,
\end{aligned}
\end{equation}
where the two adjoint terms correspond to the mechanical and thermal governing equations, weighted by $\alpha_u$ and $\alpha_T$, respectively. We assume that the body convection coefficient $h_v$ is constant across all material phases. Applying the divergence theorem to the mechanical adjoint term gives:
\begin{equation}\label{eq mech adj 1}
\int_{\Omega} \vb_{\ub}(\xb)\big(\nabla\cdot\sigmab(\xb)\big)\,dV
=
\int_{\partial\Omega} \vb_{\ub}(\xb)\cdot\big(\sigmab(\xb)\cdot\nb(\xb)\big)\,ds
- \int_{\Omega} \sigmab(\xb):\nabla\vb_{\ub}(\xb)\,dV.
\end{equation}
Using the thermo-mechanical constitutive relation in \Cref{eq constitutive tm}, we obtain the mechanical adjoint expansion given as:
\begin{equation}\label{eq mech adj 2}
\begin{aligned}
\int_{\Omega} \vb_{\ub}(\xb)\big(\nabla\cdot\sigmab(\xb)\big)\,dV
&=
\int_{\partial\Omega} \vb_{\ub}(\xb)\cdot\big(\sigmab(\xb)\cdot\nb(\xb)\big)\,ds\\
&- \int_{\Omega} \Cb(\xb):\big(\nabla\ub(\xb) - \alphab(\xb)\Delta T(\xb,\rhob) \big):\nabla\vb_{\ub}(\xb)\,dV 
\end{aligned}
\end{equation}
Similarly, applying the divergence theorem to the thermal adjoint term yields:
\begin{equation}\label{eq thermo adj 1}
\begin{aligned}
\int_{\Omega} v_T(\xb)
\Big[\nabla\cdot\big(\kappa(\xb)\nabla T(\xb,\rhob)\big)\Big]\,dV
&=
\int_{\partial\Omega} v_T(\xb)\,\kappa(\xb)\big(\nabla T(\xb,\rhob)\cdot\nb(\xb)\big)\,ds \\
&\quad
- \int_{\Omega} \kappa(\xb)\nabla T(\xb,\rhob)\cdot\nabla v_T(\xb)\,dV.
\end{aligned}
\end{equation}
Substituting \Cref{eq mech adj 2} and \Cref{eq thermo adj 1} back into \Cref{eq coup obj 1} leads to the expanded form written as:
\begin{equation}\label{eq coup obj 2}
\begin{aligned}
L^{(a)}_C(\cdot)
&=
-\int_{\partial\Omega}\ub(\xb)\cdot\big(\delta(\xb-\xb_{out})\eb_n\big)\,ds 
+ \alpha_u \bigg(\int_{\partial\Omega} \vb_{\ub}(\xb)\cdot\sigmab(\xb)\nb(\xb)\,ds \\
&\quad
- \int_{\Omega} \bigg[\Cb(\xb):\big(\nabla\ub(\xb) - \alphab(\xb)\Delta T(\xb,\rhob)\big)\bigg]:\nabla\vb_{\ub}(\xb)\,dV\bigg) \\
&\quad
+ \alpha_T \bigg(\int_{\partial\Omega} v_T(\xb)\,\kappa(\xb)\big(\nabla T(\xb,\rhob)\cdot\nb(\xb)\big)\,ds 
- \int_{\Omega}\kappa(\xb)\nabla T(\xb,\rhob)\cdot\nabla v_T(\xb)\,dV\\
&\quad
- \int_{\Omega} h_v\Delta T(\xb,\rhob) v_T(\xb)\,dV
+ \int_{\Omega} s(\xb)\,v_T(\xb)\,dV\bigg).
\end{aligned}
\end{equation}
For simplicity, we assume that both $\Cb(\cdot)$ and $\alphab(\cdot)$ are independent of temperature. Differentiating $L_C(\cdot)$ with respect to $\rho_i$ yields:
\begin{equation}\label{eq coup sens 1}
\begin{aligned}
\frac{d L^{(a)}_C(\cdot)}{d\rho_i}
&= - \alpha_u \int_{\Omega} \bigg(\frac{\partial\Cb(\xb)}{\partial\rho_i}:\big(\nabla\ub(\xb) - \alphab(\xb) \Delta T(\xb,\rhob)\big)\bigg):\nabla\vb_{\ub}(\xb)\,dV \\ 
&\quad 
- \alpha_T \int_{\Omega}\frac{\partial\kappa(\xb)}{\partial\rho_i}\nabla T(\xb,\rhob)\cdot\nabla v_T(\xb)\,dV \\
&\quad
- \alpha_u \bigg( \int_{\Omega} \bigg[\Cb(\xb):\big(\nabla\frac{\partial\ub}{\partial\rho_i} - \frac{\partial\alphab(\xb)}{\partial\rho_i}\Delta T - \alphab(\xb)\frac{\partial T}{\partial\rho_i}\big)\bigg]:\nabla\vb_{\ub}(\xb)\,dV \\
&\quad
- \int_{\partial\Omega} \vb_{\ub}(\xb)\cdot\frac{d \sigmab}{d\rho_i}\nb(\xb)\,ds\bigg)
-\int_{\partial\Omega}\frac{\partial\ub}{\partial\rho_i}\cdot\big(\delta(\xb-\xb_{out})\eb_n\big)\,ds \\
&\quad
- \alpha_T \bigg(\int_{\Omega}\kappa(\xb)\nabla \frac{\partial T}{\partial \rho_i}\cdot\nabla v_T(\xb)\,dV
+ \int_{\Omega} h_v\frac{\partial T}{\partial \rho_i}v_T(\xb)\,dV 
 \\
&\quad
- \int_{\partial\Omega} v_T(\xb)\,\frac{\partial\kappa(\xb)}{\partial\rho_i}\big(\nabla T(\xb,\rhob) \cdot\nb(\xb)\big)\,ds
- \int_{\partial\Omega} v_T(\xb)\,\kappa(\xb)\big(\nabla \frac{\partial T}{\partial\rho_i}\cdot\nb(\xb)\big)\,ds\\
&\quad
- \int_{\Omega} \frac{\partial s(\xb)}{\partial \rho_i}v_T(\xb)\,dV\bigg).
\end{aligned}
\end{equation}
Because $\Cb(\xb)$ is symmetric, and using \Cref{eq adjoint 6}, the mechanical gradient term expands as:
\begin{equation}\label{eq coup sens 2}
\begin{aligned}
    \int_{\Omega} \bigg(\Cb(\xb):\big(\nabla\frac{\partial\ub}{\partial\rho_i} & - \frac{\partial\alphab(\xb)}{\partial\rho_i}\Delta T - \alphab(\xb)\frac{\partial T}{\partial\rho_i}\big)\bigg) :\nabla\vb_{\ub}(\xb)\,dV =
    \int_{\partial\Omega}\frac{\partial\ub}{\partial\rho_i}\cdot\big(\sigmab^{\vb}(\xb)\nb(\xb)\big)\,ds \\
    &
    - \int_{\Omega}(\nabla\cdot\sigmab^{\vb})\frac{\partial \ub}{\partial \rho_i}\, dV - \int_{\Omega} \sigmab^{\vb}(\xb) : \big(\frac{\partial\alphab(\xb)}{\partial\rho_i}\Delta T + \alphab(\xb)\frac{\partial T}{\partial\rho_i}\big)\,dV,
\end{aligned}
\end{equation}
where $\sigmab^{\vb}(\xb) = \Cb(\xb):\nabla\vb_{\ub}(\xb)$ is the adjoint stress field satisfying \Cref{eq adjoint pde}. Applying the divergence theorem to the thermal gradient term, we have:
\begin{equation}\label{eq coup sens 3}
    \int_{\Omega}\kappa(\xb)\nabla \frac{\partial T}{\partial \rho_i}\cdot\nabla v_T(\xb)\,dV = \int_{\partial\Omega}\kappa(\xb)\frac{\partial T}{\partial\rho_i}\big(\nabla v_T(\xb)\cdot\nb(\xb)\big)\, ds - \int_{\Omega}\frac{\partial T}{\partial\rho_i}\nabla\cdot\big(\kappa(\xb)\nabla v_T(\xb)\big)\,dV.
\end{equation}
Substituting \Cref{eq coup sens 2} and \Cref{eq coup sens 3} into \Cref{eq coup sens 1}, and the full sensitivity expression after rearranging terms can be given as:
\begin{equation}\label{eq coup sens 4}
\begin{aligned}
\frac{d L^{(a)}_C(\cdot)}{d\rho_i}
&= - \alpha_u \int_{\Omega} \bigg(\frac{\partial\Cb(\xb)}{\partial\rho_i}:\big(\nabla\ub(\xb) - \alphab(\xb) \Delta T(\xb,\rhob)\big)\bigg):\nabla\vb_{\ub}(\xb)\,dV \\
&\quad
- \alpha_T\int_{\Omega}\frac{\partial\kappa(\xb)}{\partial\rho_i}\nabla T(\xb,\rhob)\cdot\nabla v_T(\xb)\,dV \\
&\quad
- \alpha_u \bigg(\int_{\partial\Omega}\frac{\partial\ub}{\partial\rho_i}\cdot\big(\sigmab^{\vb}(\xb)\nb(\xb)\big)\,ds 
- \int_{\partial\Omega} \vb_{\ub}(\xb)\cdot\frac{d \sigmab}{d\rho_i}\nb(\xb)\,ds\bigg) \\
&\quad
+ \alpha_T \int_{\Omega}\frac{\partial T}{\partial\rho_i}\bigg( \nabla\cdot\big(\kappa(\xb)\nabla v_T(\xb)\big) - h_v v_T(\xb) + \frac{\alpha_u}{\alpha_T}\sigmab^{\vb}(\xb) : \alphab(\xb) \bigg)\,dV \\
&\quad
-\int_{\partial\Omega}\frac{\partial\ub}{\partial\rho_i}\cdot\big(\delta(\xb-\xb_{out})\eb_n\big)\,ds
+ \alpha_u \int_{\Omega}\sigmab^{\vb}(\xb):\frac{\partial\alphab(\xb)}{\partial\rho_i}\Delta T\, dV\\
&\quad
- \alpha_T \bigg(\int_{\partial\Omega}\kappa(\xb)\frac{\partial T}{\partial\rho_i}\big(\nabla v_T(\xb)\cdot\nb(\xb)\big)\, ds
- \int_{\partial\Omega} v_T(\xb)\,\frac{\partial\kappa(\xb)}{\partial\rho_i}\big(\nabla T(\xb,\rhob) \cdot\nb(\xb)\big)\,ds\\
&\quad
- \int_{\partial\Omega} v_T(\xb)\,\kappa(\xb)\big(\nabla \frac{\partial T}{\partial\rho_i}\cdot\nb(\xb)\big)\,ds\bigg)
+ \alpha_T \int_{\Omega} \frac{\partial s(\xb)}{\partial \rho_i}v_T(\xb)\,dV.
\end{aligned}
\end{equation}
We observe that the third and fifth boundary terms on the RHS of \Cref{eq coup sens 4} match those in \Cref{eq cmpt 1} and therefore they can be eliminated. As a result, the adjoint displacement $\vb_{\ub}(\xb)$ satisfies the same governing PDEs and BCs as specified in \Cref{eq adjoint pde} and \Cref{eq bc adj}.
To eliminate the thermal volumetric integration, we choose the adjoint temperature field such that:
\begin{equation}\label{eq pde heat adj}
    -\nabla \cdot \big(\kappa(\xb) \nabla v_T(\xb)\big) 
    + h_v v_T(\xb) = \frac{\alpha_u}{\alpha_T}\sigmab^{\vb}(\xb):\alphab(\xb), 
    \quad \forall \xb \in \Omega,
\end{equation}
where it is still a heat equation with homogeneous reference temperature for volume convection while the internal heat generation is governed by $\sigmab^{\vb}(\xb):\alphab(\xb)$. From the BCs for $T(\xb,\rhob)$ in \Cref{eq bc heat}, the adjoint temperature BCs are chosen as:
\begin{subequations} \label{eq bc heat adj}
    \begin{align}
    -\kappa(\xb)\nabla v_T(\xb)\cdot \nb(\xb) &= 0, \quad \forall \xb \in \partial \Omega_{F}, \label{eq bc heat adj 1} \\
    v_T(\xb) &= 0, \quad \forall \xb \in \partial \Omega_D, \label{eq bc heat adj 2}
    \end{align}
\end{subequations}
These BCs ensure that all boundary terms involving $v_T(\xb)$ in \Cref{eq coup sens 4} vanish. The final expression for the sensitivity of thermo-mechanical TO is therefore given as:
\begin{equation}\label{eq coup sens 5}
\begin{aligned}
\frac{d L^{(a)}_C(\cdot)}{d\rho_i}
&= - \alpha_u\int_{\Omega} \bigg(\frac{\partial\Cb(\xb)}{\partial\rho_i}:\big(\nabla\ub(\xb)
- \alphab(\xb) \Delta T(\xb,\rhob)\big)\bigg):\nabla\vb_{\ub}(\xb)\,dV \\
&\quad
- \alpha_T \int_{\Omega}\frac{\partial\kappa(\xb)}{\partial\rho_i}\nabla T(\xb,\rhob)\cdot\nabla v_T(\xb)\,dV 
+ \alpha_u \int_{\Omega}\sigmab^{\vb}(\xb):\frac{\partial\alphab(\xb)}{\partial\rho_i}\Delta T\, dV \\
& \quad
+ \alpha_T \int_{\Omega} \frac{\partial s(\xb)}{\partial \rho_i}v_T(\xb)\,dV.
\end{aligned}
\end{equation}
For isothermal expansion problems, the temperature field is independent of $\rhob(\cdot)$ (i.e., $\partial T(\xb,\rhob)/\partial\rho_i = 0$). The sensitivity then simplifies to:
\begin{equation}\label{eq coup sens 6}
\begin{aligned}
\frac{d L^{(a)}_C(\cdot)}{d\rho_i}
&= -\alpha_u\bigg( \int_{\Omega} \bigg(\frac{\partial\Cb(\xb)}{\partial\rho_i}:\big(\nabla\ub(\xb)
- \alphab \Delta T(\xb,\rhob)\big)\bigg):\nabla\vb_{\ub}(\xb)\,dV \\
& \quad + \int_{\Omega}\sigmab^{\vb}(\xb):\frac{\partial\alphab(\xb)}{\partial\rho_i}\Delta T\, dV\bigg).
\end{aligned}
\end{equation}

\begin{table*}[!b]
    \centering
    \renewcommand{\arraystretch}{1.5}
    \small
    \setlength\tabcolsep{8pt}
    \caption{\textbf{Parameters of artificial materials:}
    The Young's modulus $E$ ($\niM\niPa$), thermal conductivity $\kappa$ ($\niW/(\nim\niK)$), discrete physical density $\hat{\rho}_i$, and unit-volume price $\hat{p}_i$ are reported for four phases.}
    \begin{tabular}{l|c|c|c|c} 
        \hline
        \textbf{Parameter} 
        & \textbf{Void} 
          \matbox{VoidColor} \quad
        & \textbf{Material 1}
          \matbox{MatOne} \quad
        & \textbf{Material 2} 
          \matbox{MatTwo} \quad
        & \textbf{Material 3} 
          \matbox{MatThree} \quad \\ 
        \hline
        $E_i$            
            & $10^{-5}$ & $0.4$ & $0.6$ & $1.0$ \\
        $\kappa_i$            
            & $10^{-5}$ & $0.2$ & $0.5$ & $1.0$ \\
        $\hat{\rho}_i$ 
            & $0$ & $0.5$ & $0.7$ & $1.0$ \\
        $\hat{p}_i$    
            & $0$       & $1.6$ & $1.2$ & $1.0$ \\
        \hline
    \end{tabular}
    \label{tab cm mat}
\end{table*}

The penalized loss in \Cref{eq penal1} involves an implicit dependence of the solution variables on the design variable through the governing PDEs. 
During implementation, we obtain adjoint-consistent gradients by augmenting \Cref{eq penal1} with the adjoint potential energy functionals, where \(L_M^{(a)}\) and \(L_T^{(a)}\) correspond to the adjoint displacement and adjoint temperature fields, respectively. 
In addition, the design objective \(L_C(\cdot)\) is replaced by its adjoint-consistent form \(L_C^{(a)}(\cdot)\) to match the sensitivity expression in \Cref{eq coup sens 5}. 
This yields the updated penalized loss in \Cref{eq penal2} which simultaneously enforces the primal physics constraints and the adjoint equations, enabling efficient and consistent gradient evaluation for the design update.

\section{Results and Discussions} \label{sec result}
We consider CM, heat conduction optimization, compliant mechanism design, and thermo-mechanical device design in \Cref{subsec result cm,subsec result heat,subsec result cmpt,subsec result thermo}, respectively. The first three problems are well-established benchmarks whereas the last one represents a more challenging multi-physics design task. For the multi-material studies in \Cref{subsec result cm,subsec result heat,subsec result cmpt} we adopt three artificial materials as candidates from \citep{zuo_mm_2017} whose mechanical properties and corresponding designated colors for visualization are summarized in \Cref{tab cm mat}. In \Cref{subsec result thermo} we use real engineering materials, see \Cref{tab thermo prop}.

We compare our approach to open-source codes for the three benchmark problems and use COMSOL as the baseline for the thermo-mechanical device design. The open-source codes include PolyMat \citep{sanders_polymat_2018} (which only accommodates 2D CM problems and hence is only used in \Cref{subsec result cm}) and SIMP \citep{da_mm_2022,liu_efficient_2014}. Since these implementations \Citep{liu_efficient_2014,sanders_polymat_2018} do not include projection functions \Citep{wang_projection_2011} to remove gray areas, we also do not use any projection function in our approach for a fair comparison.

All simulations based on SIMP, PolyMat, and COMSOL are run on an Intel\textsuperscript{\textregistered} Core\textsuperscript{TM} i7-11700K CPU. Our framework is implemented in Python and executed on an NVIDIA A100 GPU. We repeat each experiment 10 times with random initializations to assess robustness and use FEM to evaluate the objective function of the final topologies designed by all approaches in all problems. 

As described in \Cref{subsec method training}, our framework leverages a collection of regular grids $\boldsymbol{\Sigma} = [\Xb_1, \dots, \Xb_{n_g}]$ to evaluate the loss function in \Cref{eq penal2}. We use $n_g = 51$ in all our studies and specify the grid resolutions via CP counts and adopt the same node-count notation to report the FE grid resolution for the SIMP baselines. Detailed node counts are included in each subsection. 

\begin{figure*}[!b]
    \centering
    \begin{subfigure}[t]{0.24\textwidth}
        \captionsetup{justification=raggedright, singlelinecheck=false, skip=0pt, position=top}
        \caption{\textbf{MBB beam 2D}}
        \includegraphics[width=\linewidth]{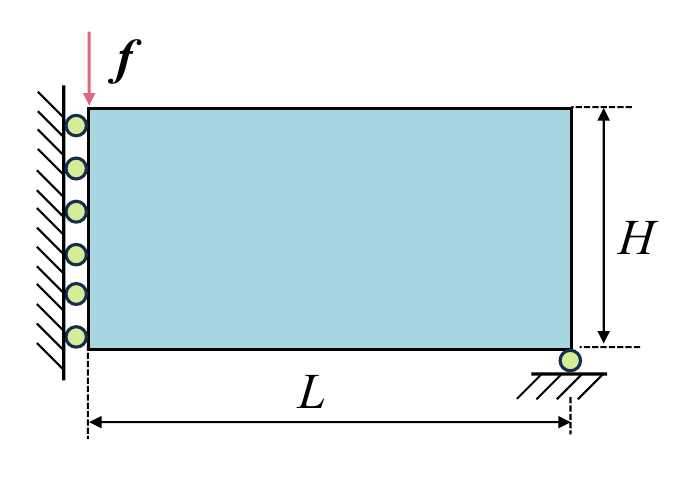}
        \label{fig exam2D 1}
    \end{subfigure}
    \begin{subfigure}[t]{0.24\textwidth}
        \captionsetup{justification=raggedright, singlelinecheck=false, skip=0pt, position=top}
        \caption{\textbf{Cantilever beam 2D}}
        \includegraphics[width=\linewidth]{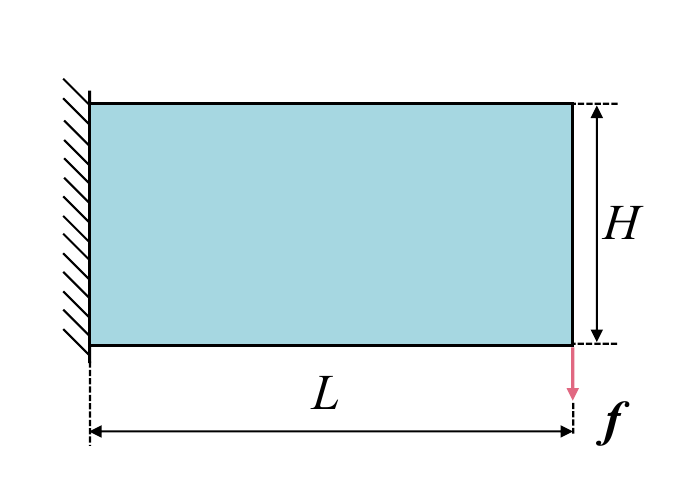}
        \label{fig exam2D 2}
    \end{subfigure}
    \begin{subfigure}[t]{0.24\textwidth}
        \captionsetup{justification=raggedright, singlelinecheck=false, skip=0pt, position=top}
        \caption{\textbf{Bridge beam 2D}}
        \includegraphics[width=\linewidth]{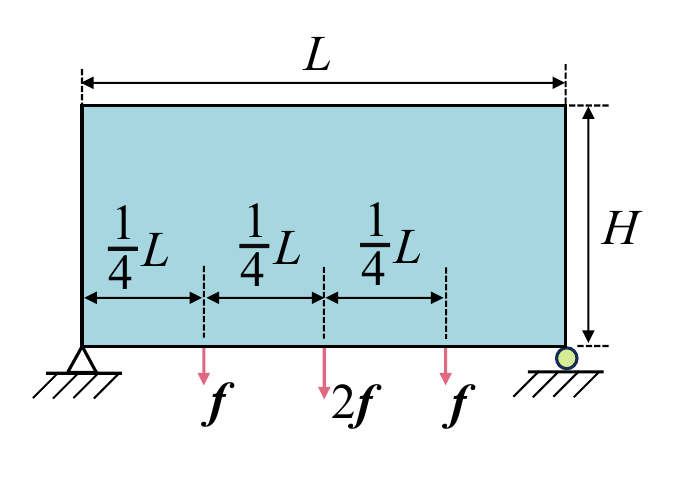}
        \label{fig exam2D 3}
    \end{subfigure}
    \begin{subfigure}[t]{0.24\textwidth}
        \captionsetup{justification=raggedright, singlelinecheck=false, skip=0pt, position=top}
        \caption{\textbf{L-shape beam 2D}}
        \includegraphics[width=\linewidth]{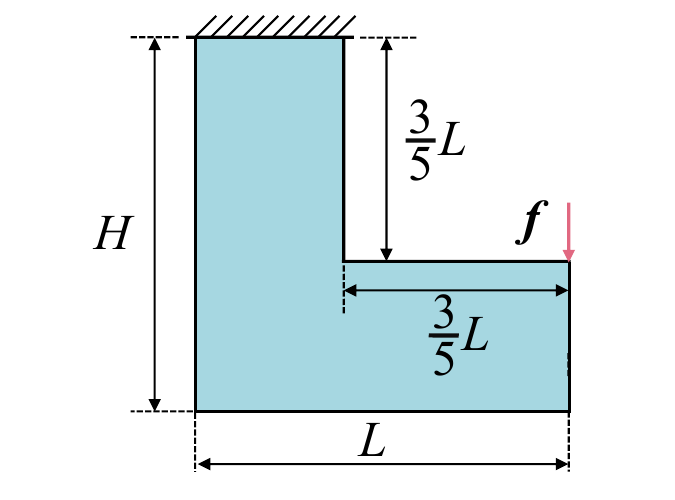}
        \label{fig exam2D 4}
    \end{subfigure}
    \begin{subfigure}[t]{0.24\textwidth}
        \captionsetup{justification=raggedright, singlelinecheck=false, skip=0pt, position=top}
        \caption{\textbf{MBB beam 3D}}
        \includegraphics[width=\linewidth]{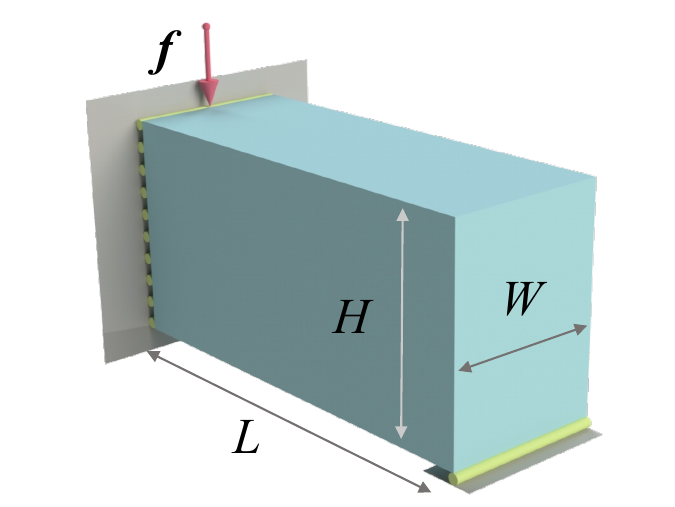}
        \label{fig exam3D 1}
    \end{subfigure}
    \begin{subfigure}[t]{0.24\textwidth}
        \captionsetup{justification=raggedright, singlelinecheck=false, skip=0pt, position=top}
        \caption{\textbf{Cantilever beam 3D}}
        \includegraphics[width=\linewidth]{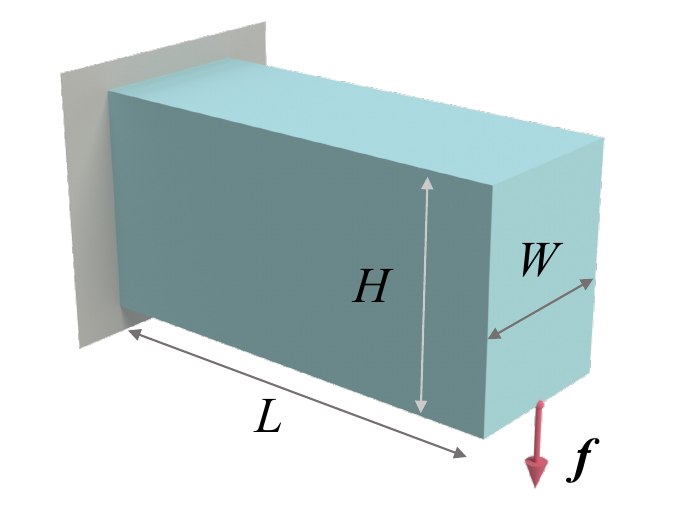}
        \label{fig exam3D 2}
    \end{subfigure}
    \begin{subfigure}[t]{0.24\textwidth}
        \captionsetup{justification=raggedright, singlelinecheck=false, skip=0pt, position=top}
        \caption{\textbf{Bridge beam 3D}}
        \includegraphics[width=\linewidth]{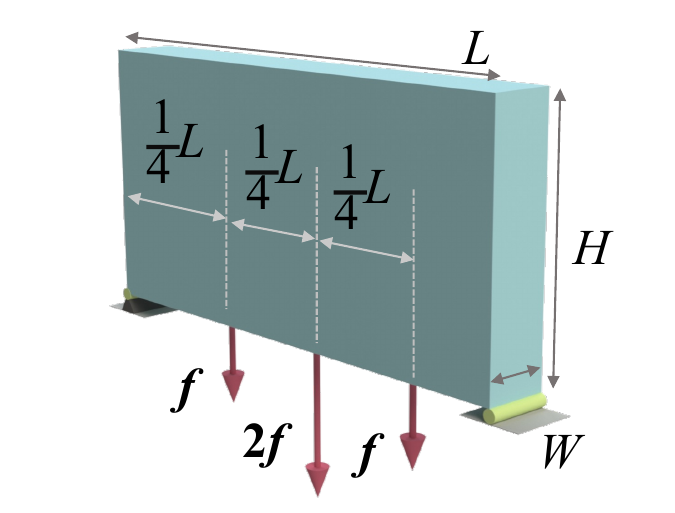}
        \label{fig exam3D 3}
    \end{subfigure}
    \begin{subfigure}[t]{0.24\textwidth}
        \captionsetup{justification=raggedright, singlelinecheck=false, skip=0pt, position=top}
        \caption{\textbf{L-shape beam 3D}}
        \includegraphics[width=\linewidth]{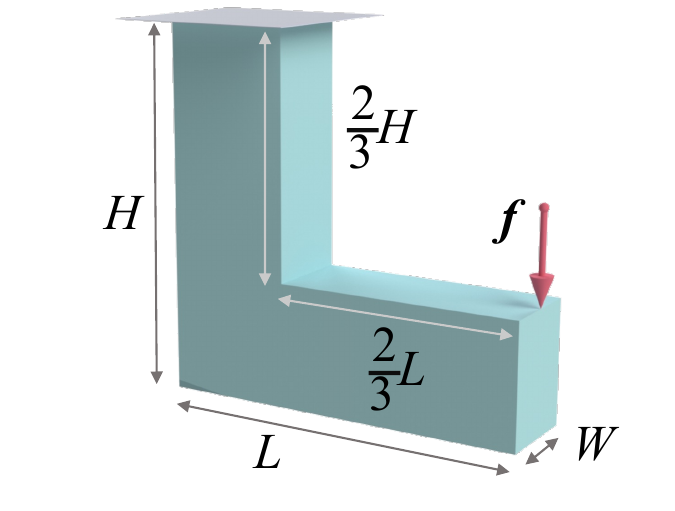}
        \label{fig exam3D 4}
    \end{subfigure}
    \vspace{-2em}
    \caption{\textbf{Benchmark examples for CM:} Four canonical benchmark problems are considered, including the Messerschmitt--Bölkow--Blohm (MBB) beam, cantilever beam, bridge beam, and L-shaped beam. The configurations of the 2D examples are shown in (a)–(d) and their 3D counterparts are presented in (e)–(h).}
    \label{fig example}
\end{figure*}

We use PGCAN to parameterize the mean functions of independent GPs associated with all state and design variables. Both the convolution-based encoder and the feature decoder employ the $\tanh$ activation function. 
For the encoder, the number of grid repetitions and features per grid vertex are set to $N_{\mathrm{rep}} = 3$ and $N_f = 128$, respectively. The initial feature tensor $\Fb_0 \in \mathbb{R}^{N_{\mathrm{rep}} \times N_f \times N_x^e \times N_y^e}$ is defined by the number of grid vertices $N_x^e$ and $N_y^e$, which are determined using the resolution parameter $\Res = 36$ \citep{sun_smo_2025}. 
The decoder is built as a shallow multilayer perceptron with 3 hidden layers. 
All models are trained using the Adam optimizer for up to $10{,}000$ epochs, starting from an initial learning rate of $10^{-3}$ that is reduced by a factor of $0.75$ at four evenly spaced stages during training. The loss weighting coefficients in \Cref{eq penal2} are chosen as $\omega_m = \omega_t = 1$ and $\omega_v = \omega_p = 100$ for all examples.

\subsection{Multi-Material Compliance Minimization in 2D and 3D}\label{subsec result cm}
We consider four canonical examples in both two and three dimensions as shown in \Cref{fig example}. The corresponding domain sizes, applied external loads, target mass fractions $\psi_m$, and cost fractions $\psi_p$ are summarized in \Cref{tab cm dimension}. For the bridge beam example, half of the design domain is considered due to symmetry. The sizes of the design domains and the magnitudes of the applied loads are selected to ensure the validity of the small-deformation assumption and linear elasticity.

\begin{table*}[!t]
    \centering
    \renewcommand{\arraystretch}{1.5}
    \small
    \setlength\tabcolsep{8pt}
    \small
    \begin{tabular}{l|c|c|c|c|c|c|c|c} 
    \hline
    \textbf{Example} & $H$ ($\nimm$) & $L$ ($\nimm$) & $W$ ($\nimm$) & $f$ ($\niN$) & \textbf{$\psi_m$} & $M_0$ & \textbf{$\psi_p$} & $P_0$ \\ 
    \hline
    MBB beam 2D           & 100   & 200  & --   & 0.1    & 0.3 & 2$\times10^4$ & 0.4 & 3.2$\times10^4$ \\
    Cantilever beam 2D    & 100   & 200  & --   & 0.1    & 0.3 & 2$\times10^4$ & 0.4 & 3.2$\times10^4$ \\
    Bridge beam 2D        & 100   & 200  & --   & 0.1    & 0.3 & 2$\times10^4$ & 0.4 & 3.2$\times10^4$ \\
    L-shape beam 2D       & 100   & 100  & --   & 0.1    & 0.2 & 6.4$\times10^3$ & 0.3 & 1.024$\times10^4$ \\
    MBB beam 3D           & 30    & 60   & 20   & 0.1    & 0.1 & 3.6$\times10^4$  & 0.2 & 5.76$\times10^4$\\
    Cantilever beam 3D    & 30    & 60   & 20   & 0.1    & 0.1 & 3.6$\times10^4$ & 0.2 & 5.76$\times10^4$\\
    Bridge beam 3D        & 60    & 120   & 15   & 0.1   & 0.1 & 1.08$\times10^5$ & 0.2 & 1.728$\times10^4$\\
    L-shape beam 3D       & 60    & 60    & 15   & 0.1   & 0.1 & 3$\times10^4$ & 0.2 & 4.8$\times10^4$ \\
    \hline
    \end{tabular}
    \caption{\textbf{Dimensions and parameters for CM:}
    The geometric dimensions, applied external forces, prescribed mass or cost fractions, and the corresponding reference mass $M_0$ and cost $P_0$ for the four benchmark examples are summarized. All dimensions are given in $\nimm$ and $f$ denotes the magnitude of the applied load ($\mathrm{N}$).}
    \label{tab cm dimension}
\end{table*}

\begin{table*}[!b]
    \centering
    \renewcommand{\arraystretch}{1.5}
    \small
    \setlength\tabcolsep{6pt}
    \caption{\textbf{Grid resolutions for CM:}
    For the ordered SIMP, the FE node counts $(N_x, N_y)$ along $L$ and $H$ dimensions are reported. 
    For the PolyMat, only the total number of FE nodes is provided due to its use of unstructured polygonal meshes. 
    For the SIMP 3D, the FE node counts $(N_x, N_y, N_z)$ along the domain dimensions $L$, $H$, and $W$ are reported. 
    For our PIGP framework, the numbers of CPs are listed for both coarse and fine grids: $(N_x^{(1)}, N_y^{(1)})$ and $(N_x^{(n_g)}, N_y^{(n_g)})$ for the 2D cases, and $(N_x^{(1)}, N_y^{(1)}, N_z^{(1)})$ and $(N_x^{(n_g)}, N_y^{(n_g)}, N_z^{(n_g)})$ for the 3D cases.}

    \begin{tabular}{l|c|c|cc||c|cc}
        \hline
        \multirow{2}{*}{\textbf{Example}}
        & \multicolumn{1}{c|}{\textbf{Ordered SIMP}} 
        & \multicolumn{1}{c|}{\textbf{PolyMat}} 
        & \multicolumn{2}{c||}{\textbf{PIGP 2D}} 
        & \multicolumn{1}{c|}{\textbf{SIMP 3D}}
        & \multicolumn{2}{c}{\textbf{PIGP 3D}} \\
        \cline{2-8}
        & \textbf{FE Nodes} 
        & \textbf{Total Nodes} 
        & \textbf{Coarse} 
        & \textbf{Fine}
        & \textbf{FE Nodes} 
        & \textbf{Coarse} 
        & \textbf{Fine} \\
        \hline
        MBB
            & (301, 151) & 45413
            & (201, 101) & (401, 201)
            & (77, 39, 27)
            & (61, 31, 21) & (92, 47, 32) \\
        Cantilever
            & (301, 151) & 45413
            & (201, 101) & (401, 201)
            & (77, 39, 27)
            & (61, 31, 21) & (92, 47, 32) \\
        Bridge
            & (151, 151) & 22734
            & (101, 101) & (201, 201)
            & (76, 76, 20)
            & (61, 61, 16) & (92, 92, 24) \\
        L-shape
            & (151, 151) & 14701
            & (101, 101) & (201, 201)
            & (76, 76, 20)
            & (61, 61, 16) & (92, 92, 24) \\
        \hline
    \end{tabular}
    \label{tab cm grid}
\end{table*}

For 2D problems, we compare our PIGP framework against ordered SIMP \citep{da_mm_2022} and PolyMat which uses polygonal FEs \citep{sanders_polymat_2018}. In 3D, we use the single-material SIMP 3D implementation from \citep{liu_efficient_2014} for comparison since an open-source implementation of 3D multi-material CM is not readily available. 
The coarse and fine grid resolutions for the benchmark CM problems are summarized in \Cref{tab cm grid}. For ordered SIMP, we report the number of FE nodes along each domain edge while for PolyMat the total number of FE nodes is enumerated since it uses unstructured polygonal meshes. For ordered SIMP a filter \citep{da_mm_2022} of radius $6$ is applied whereas PolyMat employs a density filter with a radius of $2\,\nimm$, leading to final designs with comparable structural complexity. We also implemented an NN-based multi-material TO algorithm (TOuNN) from \citep{chandrasekhar_MMTOuNN_2021} for the MBB and Bridge examples using the same material properties and geometric dimensions as specified in \Cref{tab cm mat} and \Cref{tab cm dimension}. However, the resulting topologies were overly simplified with the majority of the design domain dominated by a single material phase. Hence, we have not included TOuNN in our comparative studies.

\begin{figure*}[!t]
    \centering
    \includegraphics[width = 1.0\textwidth]{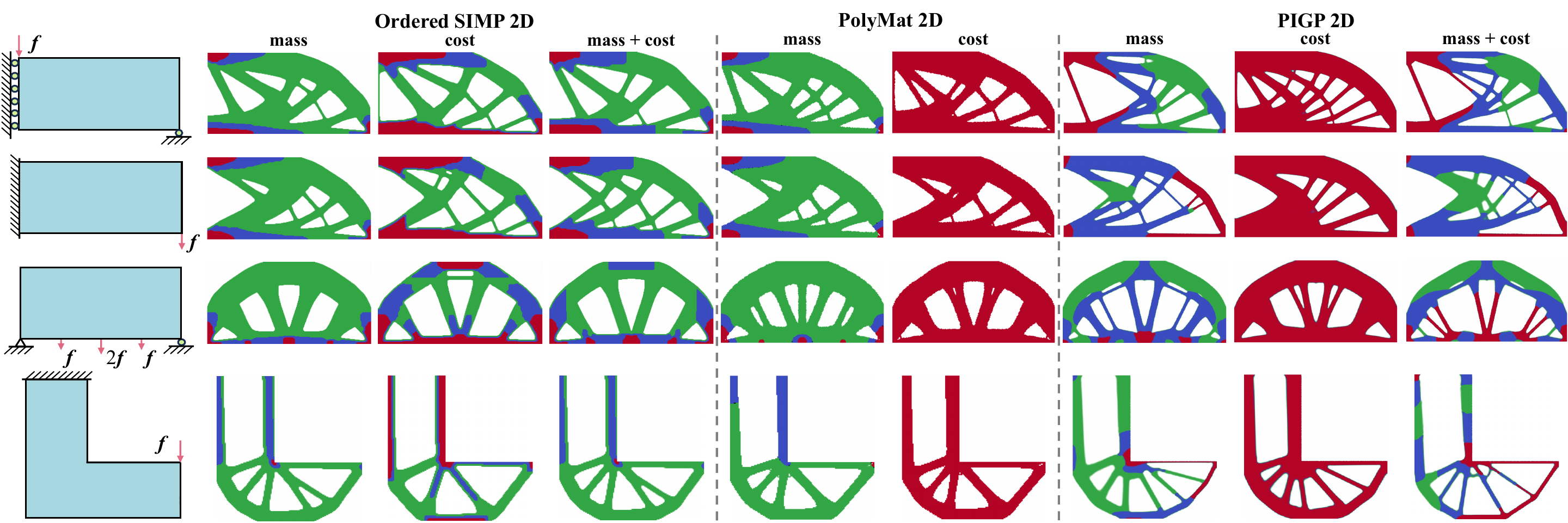} 
    \caption{\textbf{Designed multi-material topologies for 2D CM:} Representative optimized topologies are shown for each method under different constraints. PolyMat accommodates only mass or cost constraint but not both.}
    \label{fig cm 2D}
\end{figure*}
\begin{table*}[!b]
    \centering
    \renewcommand{\arraystretch}{1.5}
    \footnotesize
    \setlength\tabcolsep{6pt}
    \caption{\textbf{Median compliance and constraint values of the designed multi-material 2D topologies:} Compliance values are computed by FEM for the final topologies designed by each method under different constraints: mass, cost, or (mass, cost). Each value represents the median of 10 independent runs.}
     \begin{adjustbox}{width=\linewidth}
    \begin{tabular}{l|l|ccc|cc|ccc}
        \hline
        \multirow{2}{*}{\textbf{Example}}
        & \multirow{2}{*}{\textbf{Metric}}
        & \multicolumn{3}{c|}{\textbf{Ordered SIMP 2D}} 
        & \multicolumn{2}{c|}{\textbf{PolyMat 2D}} 
        & \multicolumn{3}{c}{\textbf{PIGP 2D}} \\
        \cline{3-5}\cline{6-7}\cline{8-10}
        & 
        & \textbf{Mass} 
        & \textbf{Cost} 
        & (\textbf{Mass, Cost}) 
        & \textbf{Mass} 
        & \textbf{Cost} 
        & \textbf{Mass} 
        & \textbf{Cost} 
        & (\textbf{Mass, Cost}) \\ 
        \hline

        \multirow{2}{*}{MBB}
        & Compliance & 0.982 & 0.991 & 1.010 & 0.987 & 0.652 & 1.049 & 0.651 & 1.104 \\
        & Constraint &  5990.8   &  12792.9 & (5989.4, 8835.2)    & 5988.0    & 12802.5    & 6006.3    & 12805.5    & (6000.2, 12798.4)    \\
        \hline

        \multirow{2}{*}{Cantilever}
        & Compliance & 0.890 & 0.886 & 0.907 & 0.899 & 0.587 & 0.976 & 0.586 & 0.987 \\
        & Constraint & 5993.8    & 12791.9    & (5995.6, 8768.0)    & 5996.3    & 12797.0    & 6012.2    & 12807.5    & (6003.5, 12692.6)    \\
        \hline

        \multirow{2}{*}{Bridge}
        & Compliance & 1.041 & 1.035 & 1.076 & 1.050 & 0.771 & 1.133 & 0.769 & 1.154 \\
        & Constraint & 5989.2    & 12788.9    & (5988.4, 8914.0)    & 5999.2    & 12786.0    & 5997.0    & 12801.0    & (6009.0, 11950.0)    \\
        \hline

        \multirow{2}{*}{L-shape}
        & Compliance & 3.194 & 2.764 & 3.194 & 3.303 & 1.805 & 3.529 & 1.799 & 3.817 \\
        & Constraint & 1278.4    & 3067.8    & (1278.4, 1904.6)    & 1272.8    & 3054.0    & 1272.6    & 3065.5    & (1255.7, 2412.7)    \\
        \hline

    \end{tabular}
    \end{adjustbox}
    \label{tab cm 2D obj}
\end{table*}

The designed topologies by our PIGP framework, ordered SIMP, and PolyMat are shown in \Cref{fig cm 2D} and the corresponding compliance and constraint values (median across 10 independent runs) are summarized in \Cref{tab cm 2D obj}. 
We observe that our framework achieves comparable results to ordered SIMP and PolyMat methods under the mass-only constraint, while producing slightly finer design features. The resulting topologies from our approach are physically consistent: the stiffest material (red) is preferentially placed in stress-localized regions (e.g., near external load points or boundary corners) whereas softer materials occupy the bulk of the domain. 
As shown in \Cref{tab cm 2D obj}, our PIGP approach achieves compliance values that are slightly higher yet comparable to those obtained using ordered SIMP and PolyMat under the mass-only constraint.

Under the cost-only constraint, both PIGP and PolyMat methods converge to topologies composed exclusively of the stiffest material, which is expected since this phase has the lowest unit-volume price and the highest Young’s modulus simultaneously. In contrast, the ordered SIMP method produces topologies involving all three solid phases, which result in noticeably higher compliance values, see \Cref{tab cm 2D obj}. The inability of ordered SIMP is most likely due to the embedded nonlinear material interpolation scheme which prevents the gradient-based optimizer to converge to the global optimum. However, our approach benefits from a built-in continuation mechanism that can prevent convergence to local optima. 

When both mass and cost constraints are imposed, the resulting topologies are similar to those obtained under the mass-only constraint. This behavior arises because our selected cost targets are not very aggressive and satisfying the mass constraint often inherently satisfies the cost constraint as well. 

\begin{figure*}[!b]
    \centering
    \includegraphics[width = \textwidth]{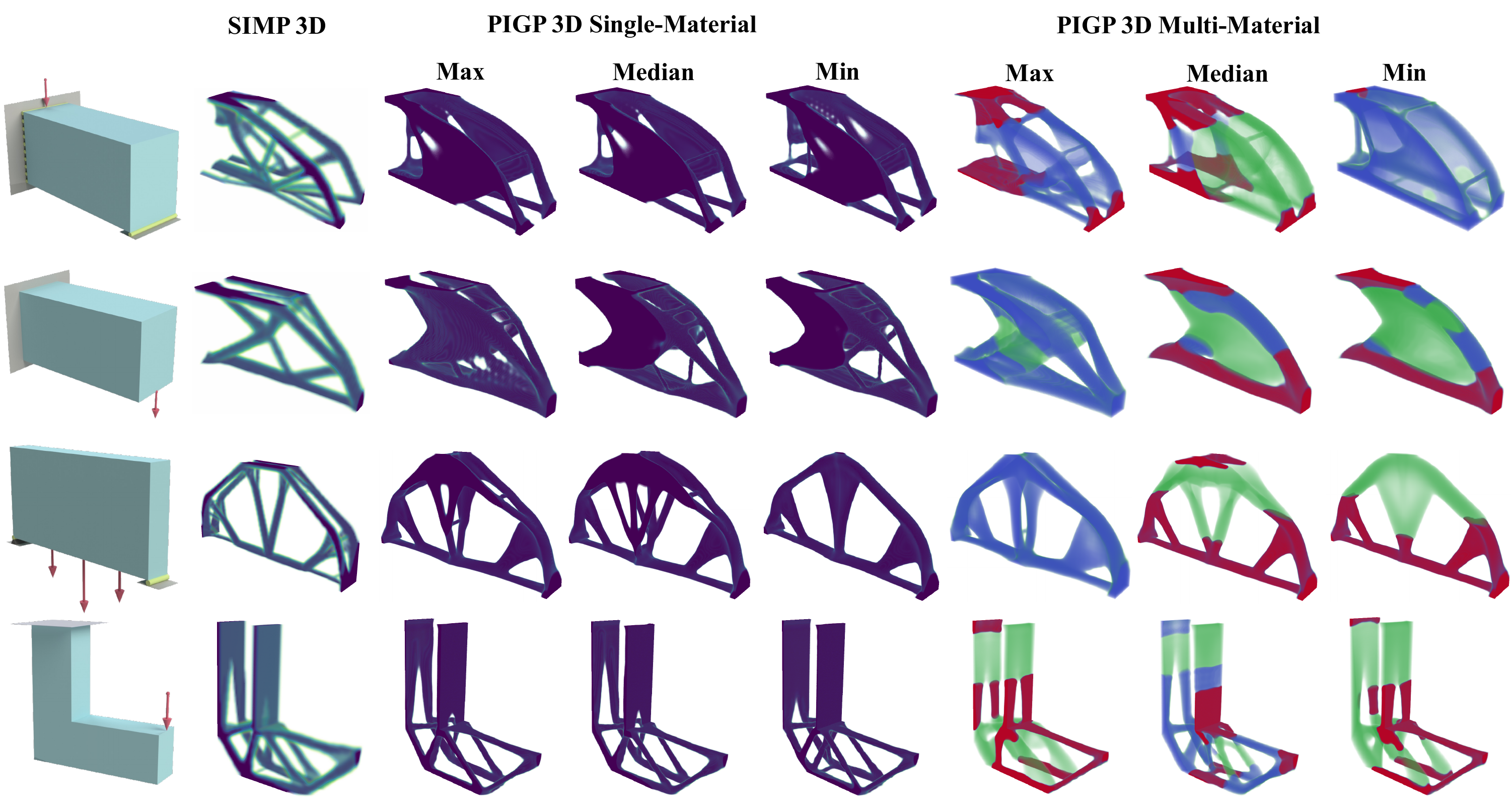} 
    \caption{\textbf{Representative designs for 3D CM:}
    For each benchmark, the topologies corresponding to the maximum, median, and minimum compliance values are shown, comparing SIMP 3D in the single-material setting with our PIGP 3D framework under both single- and multi-material settings. The three colors denote the three candidate artificial materials defined in \Cref{tab cm mat}. The multi-material designs are subject to both mass and cost constraints.}
    \label{fig cm 3D}
\end{figure*}

In \Cref{fig cm 3D} we compare 3D designs produced by SIMP 3D and our PIGP framework under a mass-only constraint. We also visualize multi-material designs generated by PIGP under mass and cost constraints. For PIGP, we visualize three designs corresponding to the maximum, minimum, and median compliance values obtained in the 10 runs. Inspecting the three samples across each problem, we observe that PIGP produces similar overall topologies whose fine features can change based on the initialization. 
The median, mean, and standard deviation of the final compliance are enumerated in \Cref{tab cm 3D} where it can be seen that for the MBB and cantilever beams, our PIGP framework yields slightly higher median compliance than SIMP 3D, whereas the opposite trend is observed with the bridge and L-shape beam. Overall, the final compliance and its standard deviation are comparable between the two single-material approaches across all 3D examples.

\begin{table*}[!t]
    \centering
    \renewcommand{\arraystretch}{1.1}
    \footnotesize
    \setlength\tabcolsep{6pt} 
    \caption{\textbf{Compliance and constraint statistics of 3D designs:} 
    The median, mean, and standard deviation of the final compliance and mass values are reported for each method. 
    Values are computed via FEM on the optimized topologies, and statistics are obtained using 10 independent runs. Single-material designs are subject to a mass constraint while the multi-material one is subject to both mass and cost (only mass is reported in the table).}
    \begin{adjustbox}{width=\linewidth}
    \begin{tabular}{l|l|ccc|ccc|ccc}
        \hline
        \multirow{2}{*}{\textbf{Example}}
        & \multirow{2}{*}{\textbf{Metric}}
        & \multicolumn{3}{c|}{\textbf{SIMP 3D Single-material}} 
        & \multicolumn{3}{c|}{\textbf{PIGP 3D Single-material}} 
        & \multicolumn{3}{c}{\textbf{PIGP 3D Multi-material}} \\
        
        \cline{3-5}\cline{6-8}\cline{9-11}
        & 
        & \textbf{Median} 
        & \textbf{Mean} 
        & \textbf{Std} 
        & \textbf{Median} 
        & \textbf{Mean} 
        & \textbf{Std} 
        & \textbf{Median} 
        & \textbf{Mean} 
        & \textbf{Std}\\ 
        \hline

        \multirow{2}{*}{MBB}
        & Compliance 
        & 0.191 & 0.194 & 7.7$\times10^{-3}$ 
        & 0.198 & 0.206 & 1.8$\times10^{-2}$ 
        & 0.171 & 0.169 & 5.6$\times10^{-3}$\\ 
        
        & Mass 
        & 3742.0 & 3748.2 & 36.20 
        & 3636.1 & 3636.5 & 20.87 
        & 3684.2 & 3692.4 & 28.06\\ 
        \hline

        \multirow{2}{*}{Cantilever}
        & Compliance 
        & 0.169 & 0.172 & 1.3$\times10^{-2}$ 
        & 0.176 & 0.179 & 1.6$\times10^{-2}$ 
        & 0.136 & 0.139 & 7.5$\times10^{-3}$\\
        
        & Mass 
        & 3660.0 & 3655.4 & 43.98 
        & 3649.4 & 3659.5 & 37.72 
        & 3655.8 & 3657.1 & 27.42\\
        \hline

        \multirow{2}{*}{Bridge}
        & Compliance 
        & 0.243 & 0.246 & 7.2$\times10^{-3}$ 
        & 0.234 & 0.232 & 6.3$\times10^{-3}$ 
        & 0.209 & 0.208 & 3.9$\times10^{-3}$\\
        
        & Mass 
        & 10792 & 10787 & 49.81 
        & 10806 & 10807 & 19.71 
        & 10799 & 10793 & 19.66\\
        \hline

        \multirow{2}{*}{L-shape}
        & Compliance 
        & 1.249 & 1.173 & 1.5$\times10^{-1}$ 
        & 1.126 & 1.153 & 1.1$\times10^{-1}$ 
        & 0.941 & 0.991 & 1.1$\times10^{-1}$\\
        
        & Mass 
        & 3022.5 & 3026.0 & 36.41 
        & 2955.8 & 2950.9 & 21.30 
        & 2913.8 & 2922.0 & 39.79\\
        \hline

    \end{tabular}
    \end{adjustbox}
    \label{tab cm 3D}
\end{table*}

In the multi-material 3D CM results, the stiffest material is preferentially allocated to the high-stress regions, consistent with the 2D results in \Cref{fig cm 2D}. In addition, introducing multiple candidate materials can further reduce the achieved compliance while achieving the same mass target, see \Cref{tab cm 3D}. For instance, the median compliance for the MBB example decreases from 0.198 to 0.171 with multi-material PIGP. We highlight that the standard deviation for multi-material PIGP is consistently lower than that of the corresponding single-material SIMP 3D cases, indicating that the proposed framework remains robust and yields more consistent solutions under multi-material settings. 

\begin{figure*}[!b]
    \centering
    \begin{subfigure}[t]{0.48\textwidth}
        \includegraphics[width=\linewidth]{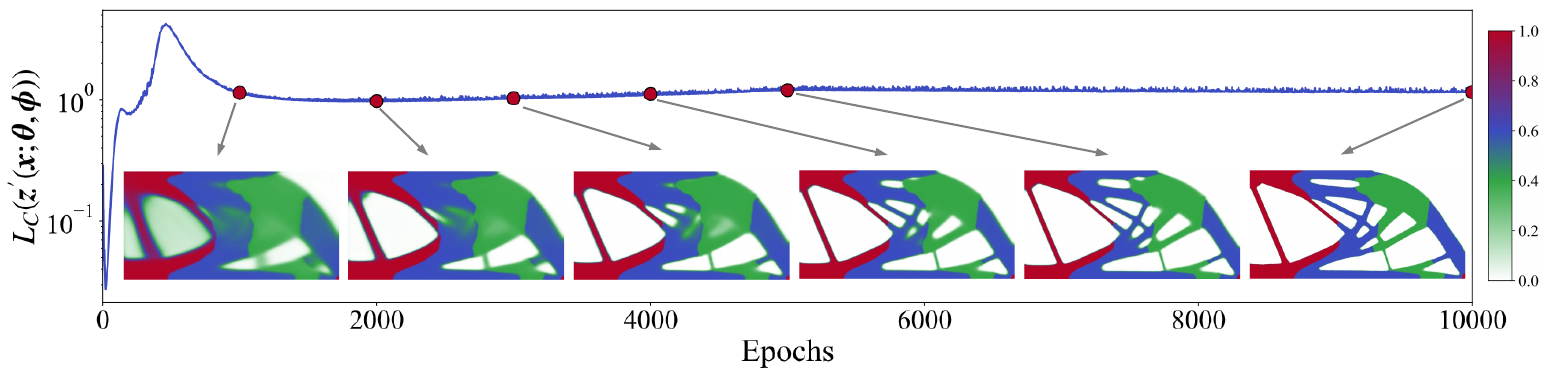}
        \vspace{-7.0em}
        \captionsetup{justification=raggedright, singlelinecheck=false, skip=-3.5pt, position=top}
        \caption[]{}
        \label{fig evo 1}
    \end{subfigure}
    \begin{subfigure}[t]{0.48\textwidth}
        \includegraphics[width=\linewidth]{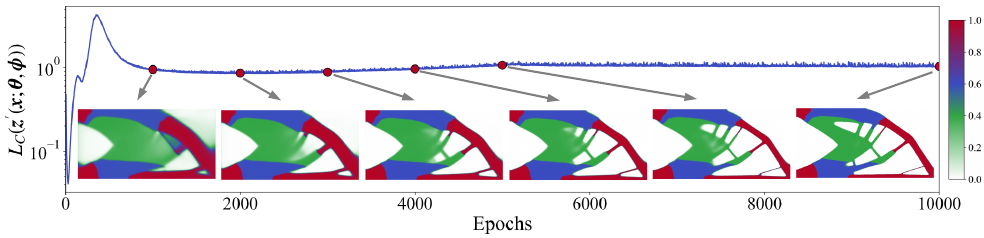}
        \vspace{-7.0em}
        \captionsetup{justification=raggedright, singlelinecheck=false, skip=-3.5pt, position=top}
        \caption[]{}
        \label{fig evo 2}
    \end{subfigure}
    \begin{subfigure}[t]{0.48\textwidth}
        \includegraphics[width=\linewidth]{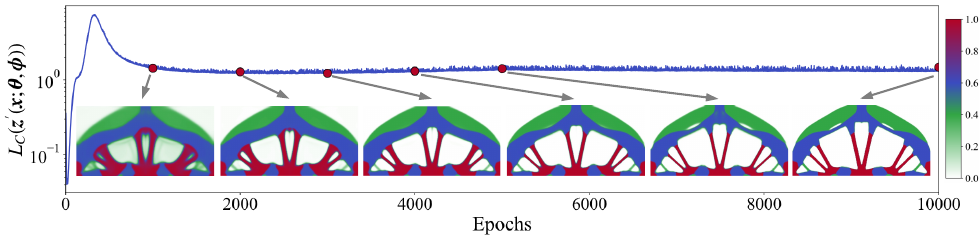}
        \vspace{-7.0em}
        \captionsetup{justification=raggedright, singlelinecheck=false, skip=-3.5pt, position=top}
        \caption[]{}
        \label{fig evo 3}
    \end{subfigure}
    \begin{subfigure}[t]{0.48\textwidth}
        \includegraphics[width=\linewidth]{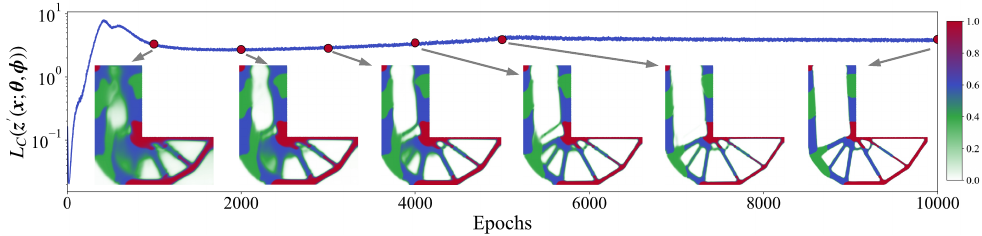}
        \vspace{-7.0em}
        \captionsetup{justification=raggedright, singlelinecheck=false, skip=-3.5pt, position=top}
        \caption[]{}
        \label{fig evo 4}
    \end{subfigure}
    \begin{subfigure}[t]{0.48\textwidth}
        \includegraphics[width=\linewidth]{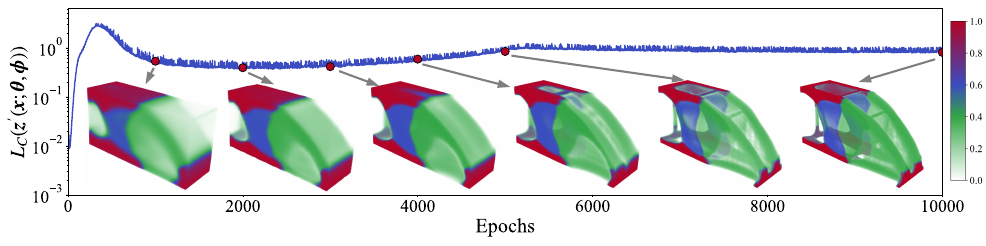}
        \vspace{-7.0em}
        \captionsetup{justification=raggedright, singlelinecheck=false, skip=-3.5pt, position=top}
        \caption[]{}
        \label{fig evo 5}
    \end{subfigure}
    \begin{subfigure}[t]{0.48\textwidth}
        \includegraphics[width=\linewidth]{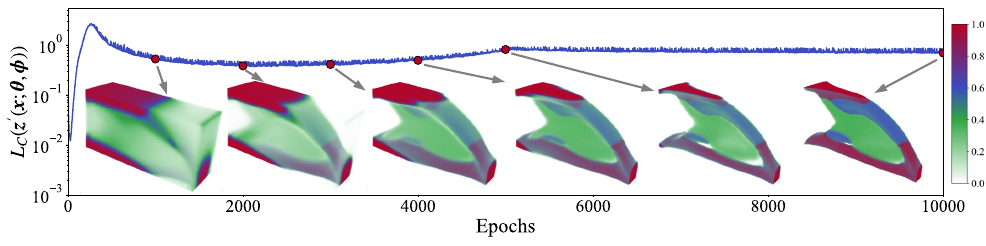}
        \vspace{-7em}
        \captionsetup{justification=raggedright, singlelinecheck=false, skip=-3.5pt, position=top}
        \caption[]{}
        \label{fig evo 6}
    \end{subfigure}
    \begin{subfigure}[t]{0.48\textwidth}
        \includegraphics[width=\linewidth]{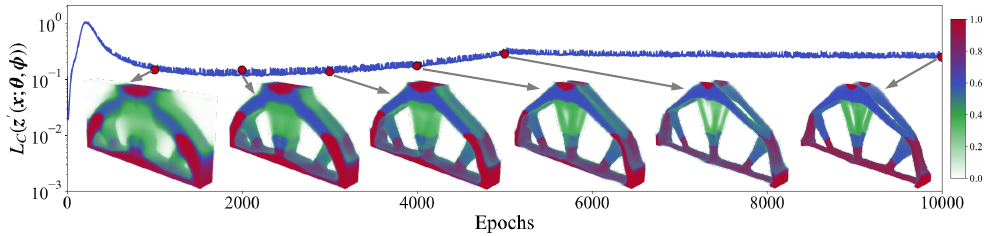}
        \vspace{-7.0em}
        \captionsetup{justification=raggedright, singlelinecheck=false, skip=-3.5pt, position=top}
        \caption[]{}
        \label{fig evo 7}
    \end{subfigure}
    \begin{subfigure}[t]{0.48\textwidth}
        \includegraphics[width=\linewidth]{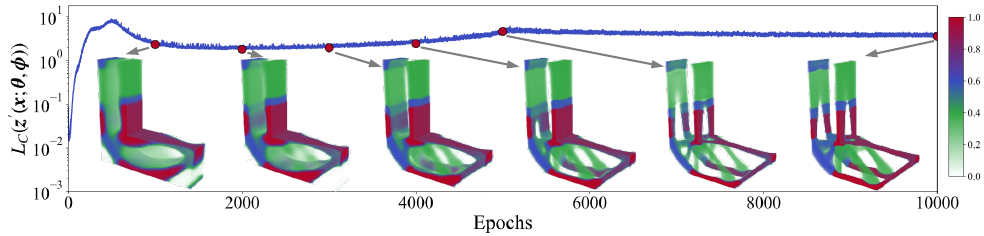}
        \vspace{-7em}
        \captionsetup{justification=raggedright, singlelinecheck=false, skip=-3.5pt, position=top}
        \caption[]{}
        \label{fig evo 8}
    \end{subfigure}
    \caption{\textbf{Topology evolution during multi-material CM training:} Intermediate topologies at selected epochs are shown along with the history of design objective $L_C(\cdot)$ for (a) 2D MBB beam, (b) 2D cantilever beam, (c) 2D bridge beam, and (d) 2D L-shaped beam. Panels (e)--(h) show the corresponding topology evolution for the same four examples in the 3D setting.}
    \label{fig evo}
\end{figure*}

To provide further insight into the performance and training dynamics of PIGP, we visualize the topology evolution for both the 2D and 3D examples at selected epochs and overlay these snapshots on the history of the design objective $L_C(\cdot)$ in \Cref{fig evo}. Across all examples, the objective histories exhibit a consistent evolution pattern. The objective increases rapidly during the early stage of training, reaches a peak at approximately the 500\textsuperscript{th} epoch, and then decreases as the optimizer begins to discover meaningful topologies. The initially small compliance is due to the near-trivial displacement fields. As the training progresses toward nontrivial force-equilibrium solutions, the compliance increases sharply and attains its peak value. After this point, the objective loss decreases as the optimization begins to reduce compliance, and discernible structural features emerge around epoch 1k with stiffer material accumulating in high-stress regions. At this stage, the design field still contains substantial intermediate densities that do not belong to any candidate material. Between 1k and 5k epochs $L_C(\cdot)$ mildly increases which is due to the curriculum training schedule in which the mass or cost fraction is progressively reduced until 50\% of the total epochs, and then held fixed thereafter. Consequently, solid material is gradually removed as shown by the embedded snapshots between the 2k and 5k epochs, leading to a modest increase in compliance before the design objective stabilizes. 
In the late stage of training, intermediate gray regions are effectively eliminated, and the density field approaches a near-binary distribution. Notably, high-quality topologies are often achieved by approximately 5k epochs, suggesting that the overall computational cost can be reduced substantially by adopting an appropriate early-stopping criterion or further improving the optimization efficiency.

\subsubsection{Shape Function vs. Finite Difference}\label{subsubsec result cm sf fd}

\begin{figure*}[!b]
    \centering
    \begin{subfigure}[t]{0.48\textwidth}
        \captionsetup{justification=raggedright, singlelinecheck=false, skip=0pt, position=top}
        \caption{\textbf{MBB beam}}
        \includegraphics[width=\linewidth]{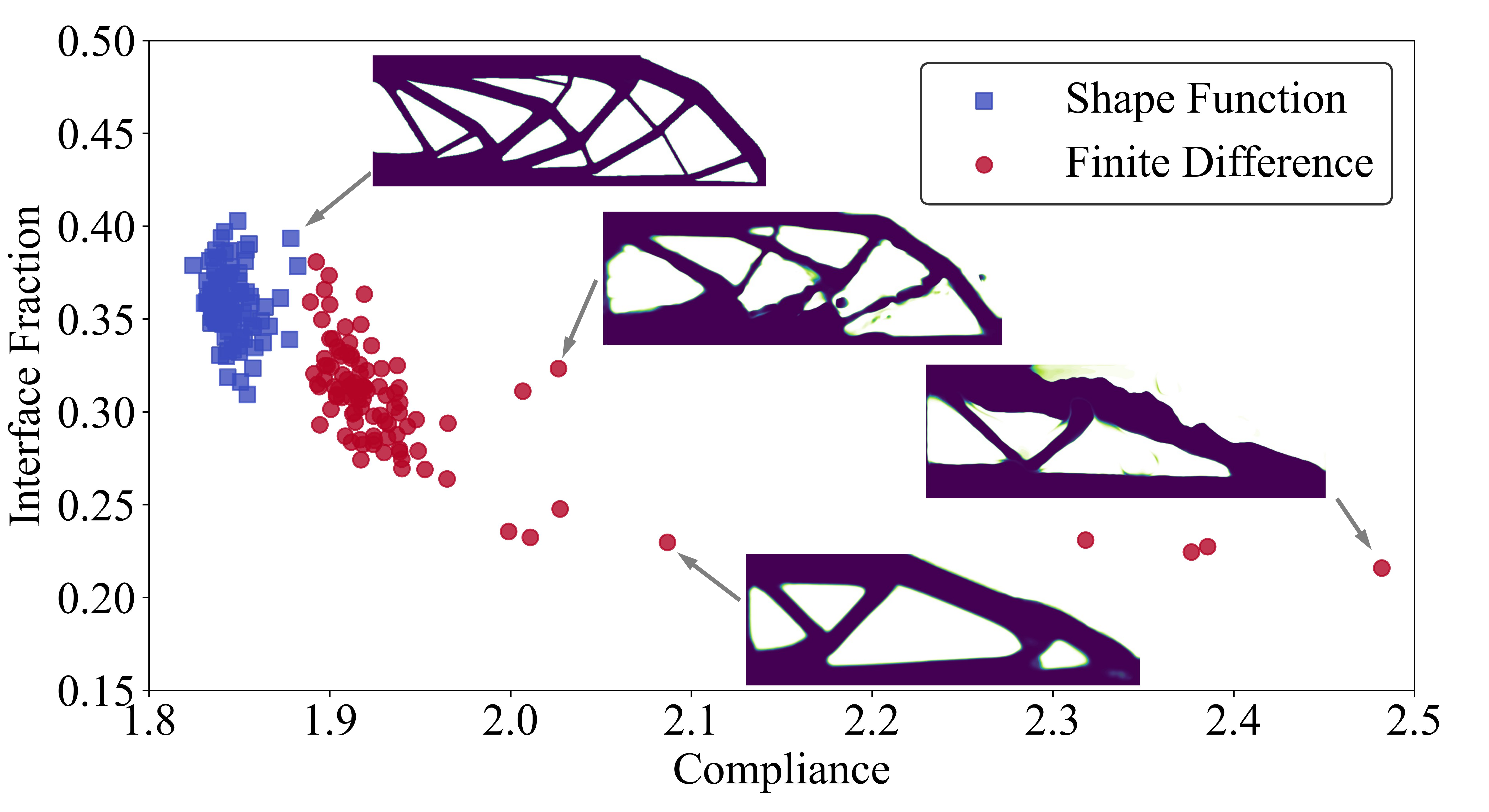}
        \label{fig sf fd 1}
    \end{subfigure}
    \begin{subfigure}[t]{0.48\textwidth}
        \captionsetup{justification=raggedright, singlelinecheck=false, skip=0pt, position=top}
        \caption{\textbf{Cantilever beam}}
        \includegraphics[width=\linewidth]{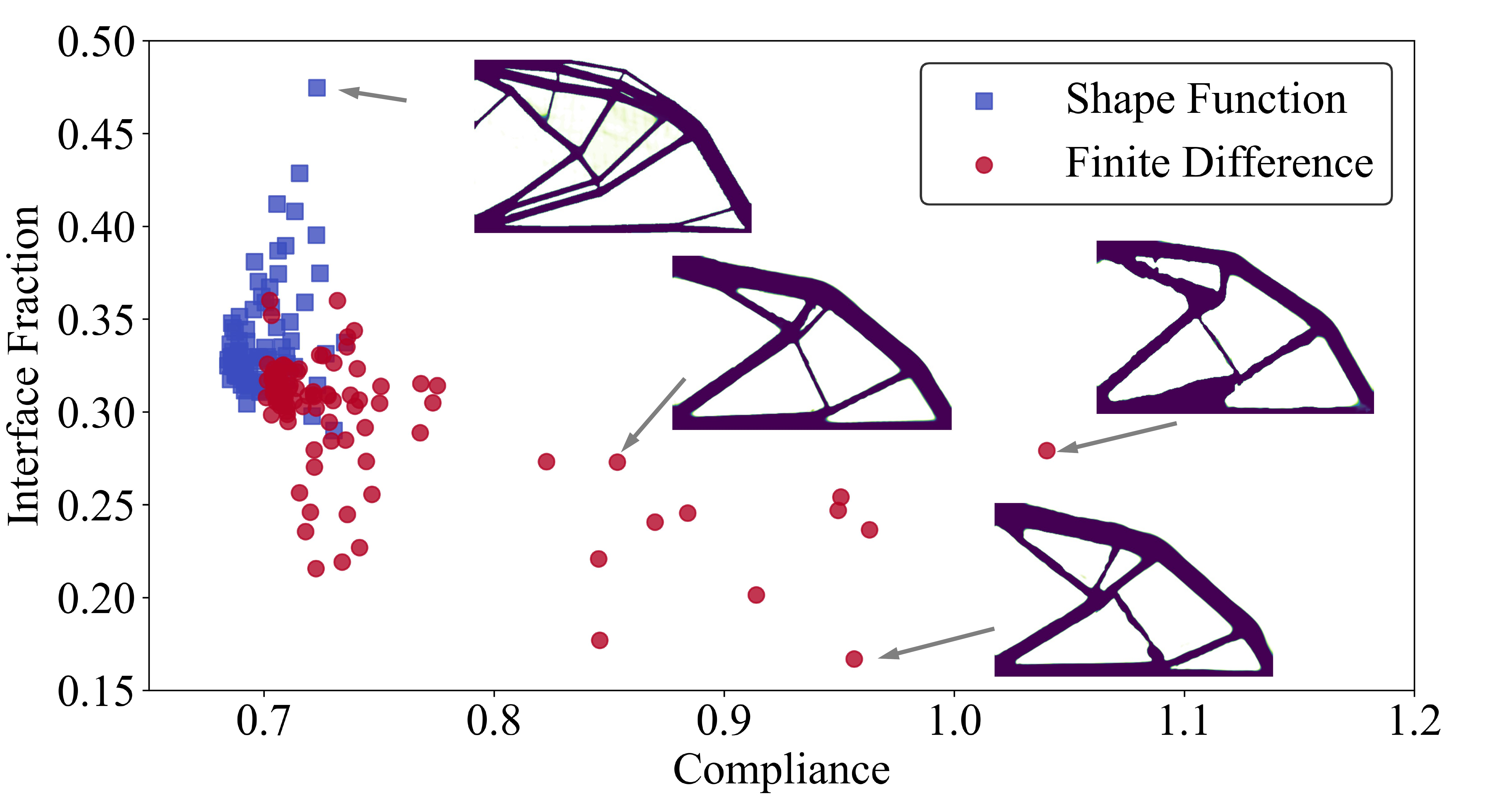}
        \label{fig sf fd 2}
    \end{subfigure}
    \vspace{-2em}
    \caption{\textbf{Effect of different approximation techniques on the robustness of our approach:} We compare the designed topologies when the residuals are approximated using either shape functions or FD. FD is less stable and generally produces topologies with higher compliance. Details of the two examples are provided in \citep{sun_smo_2025}.}
    \label{fig sf fd}
\end{figure*}

We compare the CM results using our current shape function implementation to those from our previous work using FD \citep{sun_smo_2025}. The NN architecture and all other hyperparameters are identical between the two approaches. We conduct 100 independent runs of the two approaches with random initializations and visualize the resulting distributions of interface fraction versus final compliance in \Cref{fig sf fd 1} and \Cref{fig sf fd 2} for the MBB beam and cantilever beam, respectively. The interface fraction \citep{sun_smo_2025} serves as a measure of the overall topological complexity.

For both examples, the shape-function implementation generally achieves lower compliance while producing more complex designs with higher interface fractions compared to FD. 
PIGP with shape functions yields more consistent distributions with most realizations clustering tightly around a narrow region. The obtained topologies have sharp and smooth interfaces between the void and solid phases. 
In contrast, the FD-based implementation produces several outliers in both examples.  As illustrated by the inset topologies, these outliers correspond to substantially higher compliance and rough solid-void interfaces. Consistent with our earlier hypothesis \citep{sun_smo_2025}, we attribute these less-preferable results from FD to training instabilities induced by inaccurate gradient approximations. The proposed shape-function implementation effectively mitigates this issue.

\subsection{Multi-Material Heat Conduction Optimization in 2D and 3D}\label{subsec result heat}

\begin{figure*}[!t]
    \centering
    \begin{subfigure}[t]{0.45\textwidth}
        \includegraphics[width=\linewidth]{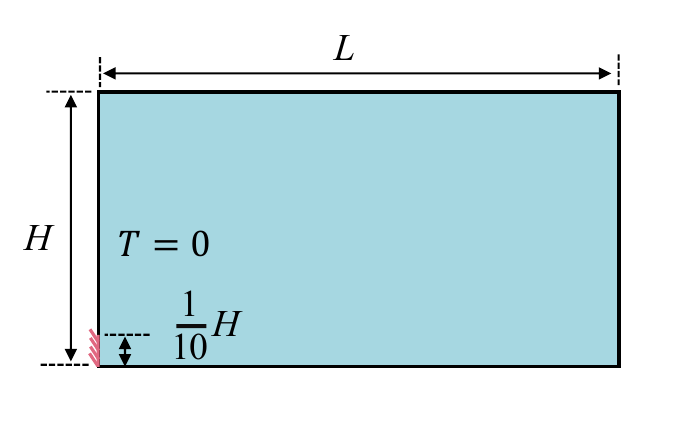}
        \vspace{-14em}
        \captionsetup{justification=raggedright, singlelinecheck=false, skip=-3.5pt, position=top}
        \caption[]{\textbf{Heat Conduction 2D}}
        \label{fig exam heat 1}
    \end{subfigure}
    \begin{subfigure}[t]{0.45\textwidth}
        \includegraphics[width=\linewidth]{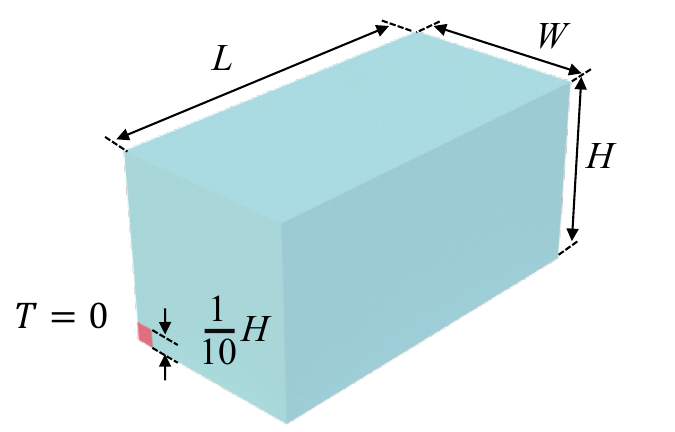}
        \vspace{-14em}
        \captionsetup{justification=raggedright, singlelinecheck=false, skip=-3.5pt, position=top}
        \caption[]{\textbf{Heat Conduction 3D}}
        \label{fig exam heat 2}
    \end{subfigure}
    \caption{\textbf{Benchmark examples for 2D and 3D heat conduction optimization:} The prescribed temperature at the bottom left corner acts as the heat sink while the entire domain is subjected to uniform volumetric heat generation.}
\label{fig exam heat}
\end{figure*}

We present the 2D and 3D benchmark examples for heat conduction optimization in \Cref{fig exam heat} where relevant parameters are summarized in \Cref{tab heat param}. Both examples impose a prescribed zero-temperature BC in the red region, while a uniform volumetric heat-generation source term is applied over the entire domain. The design objective is to minimize the system thermal compliance defined in \Cref{eq adjoint heat 1}. 
For the 2D case, we compare the results obtained by our PIGP approach with those from the ordered SIMP method under both single-material and multi-material settings. 
We adopt the same SIMP 3D implementation from \citet{liu_efficient_2014} as the baseline in the single-material setting for comparison with our PIGP approach. The FE node counts used by the two SIMP baselines, along with the coarse and fine CP grid resolutions employed in PIGP are summarized in \Cref{tab heat grid} for both examples.
The material properties are listed in \Cref{tab cm mat}. 

\begin{table*}[!b]
    \centering
    \renewcommand{\arraystretch}{1.5}
    \small
    \setlength\tabcolsep{6pt}
    \begin{tabular}{l|c|c|c|c|c|c} 
    \hline
    \textbf{Example} 
    & $H$ ($\nimm$) 
    & $L$ ($\nimm$) 
    & $W$ ($\nimm$) 
    & $s$ ($\niW/(\nimm^3)$)
    & \textbf{$\psi_m$}
    & $M_0$\\ 
    \hline
    Heat Conduction  2D & 100 & 200 & -- &  1$\times 10^{-5}$ & 0.3 & 2$\times10^4$\\
    Heat Conduction 3D  & 30 & 60 & 30 &  1$\times 10^{-5}$ & 0.1 & 5.4$\times10^4$\\
    \hline
    \end{tabular}
    \caption{\textbf{Design parameters for heat conduction examples:}
    Geometric dimensions, heat source intensity, prescribed mass fraction, and the reference mass $M_0$ for heat conduction problems shown in \Cref{fig exam heat}. }
    \label{tab heat param}
\end{table*}

\begin{table*}[!b]
    \centering
    \renewcommand{\arraystretch}{1.5}
    \small
    \setlength\tabcolsep{6pt}
    \caption{\textbf{Grid resolutions for the heat conduction benchmarks:} 
    We report the FE node counts and CP resolutions of the dynamic grids for SIMP and PIGP, respectively.}

    \begin{tabular}{l|c|cc||c|cc}
        \hline
        \multirow{2}{*}{\textbf{Example}}
        & \multicolumn{1}{c|}{\textbf{Ordered SIMP 2D}} 
        & \multicolumn{2}{c||}{\textbf{PIGP 2D}} 
        & \multicolumn{1}{c|}{\textbf{SIMP 3D}}
        & \multicolumn{2}{c}{\textbf{PIGP 3D}} \\
        \cline{2-7}
        & \textbf{FE Nodes} 
        & \textbf{Coarse} 
        & \textbf{Fine}
        & \textbf{FE Nodes} 
        & \textbf{Coarse} 
        & \textbf{Fine} \\
        \hline
        Heat Conduction 2D
            & (301, 151)
            & (201, 101) & (401, 201)
            & --
            & -- & -- \\
        Heat Conduction 3D
            & --
            & -- & --
            & (61, 31, 31)
            & (41, 21, 21) & (81, 41, 41) \\
        \hline
    \end{tabular}

    \label{tab heat grid}
\end{table*}

\begin{figure*}[!t]
    \centering
    \includegraphics[width = 1.0\textwidth]{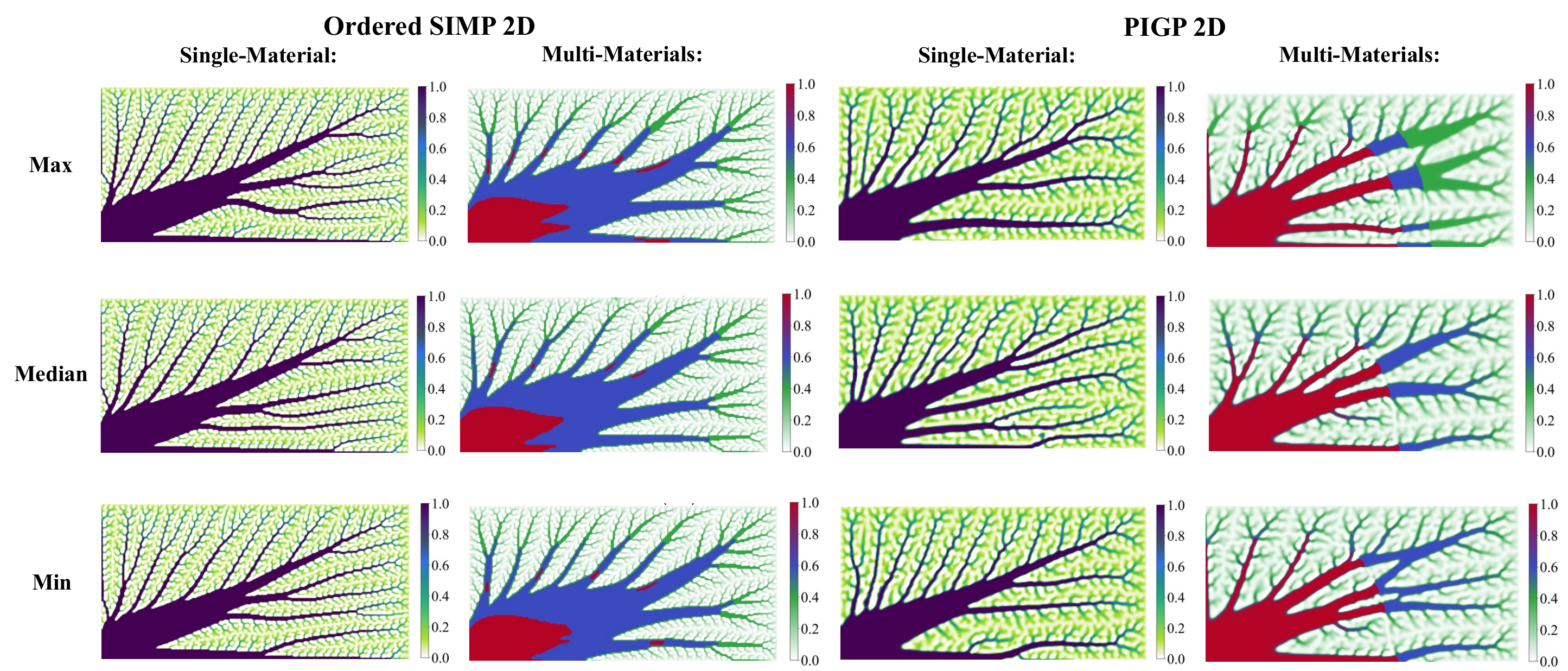} \caption{\textbf{Representative designs for 2D heat conduction optimization:} Both approaches are used for single- and multi-material cases. For each case, we observe the designs corresponding to the maximum, median, and minimum thermal compliance values.}
    \label{fig heat 2D}
\end{figure*}

\begin{table*}[!b]
    \centering
    \renewcommand{\arraystretch}{1.5}
    \footnotesize
    \setlength\tabcolsep{10pt}
    \caption{\textbf{Summary of objective and mass statistics for heat conduction optimization:}
    The values are computed across 10 runs. FEM is used to evaluate the thermal compliance of the optimized topology for each method. The 3D SIMP implementation \citep{liu_efficient_2014} only accommodates single-material problems.}
    \begin{adjustbox}{width=\linewidth}
    \begin{tabular}{l l l | c c c c c}
        \hline
        \textbf{Method} & \textbf{Materials} & \textbf{Metric}
        & \textbf{Median} & \textbf{Mean} & \textbf{Std} &
        \textbf{Min} & \textbf{Max} \\
        \hline
        \multirow{4}{*}{\textbf{Ordered SIMP 2D}}
            & \multirow{2}{*}{\textbf{Single-Material}} & \textbf{Compliance} & 1.986 & 1.988 & 0.306 & 1.650 & 2.613 \\
            &                                            & \textbf{Mass}       & 6002.0 & 6009.2 & 28.41 & 5964.0 & 6061.0 \\
            & \multirow{2}{*}{\textbf{Multi-Material}}    & \textbf{Compliance} & 0.487 & 0.481 & 0.026 & 0.443 & 0.520 \\
            &                                            & \textbf{Mass}       & 6013.9 & 6010.4 & 14.96 & 5978.2 & 6036.8 \\
        \hline
        \multirow{4}{*}{\textbf{PIGP 2D}}
            & \multirow{2}{*}{\textbf{Single-Material}} & \textbf{Compliance} & 1.775 & 1.781 & 0.200 & 1.457 & 2.156 \\
            &                                            & \textbf{Mass}       & 6037.5 & 6026.8 & 30.21 & 5949.0 & 6059.0 \\
            & \multirow{2}{*}{\textbf{Multi-Material}}    & \textbf{Compliance} & 0.577 & 0.577 & 0.035 & 0.534 & 0.647 \\
            &                                            & \textbf{Mass}       & 6031.3 & 6032.9 & 8.75 & 6021.4 & 6054.0 \\
        \hline
        \multirow{2}{*}{\textbf{SIMP 3D}}
            & \multirow{2}{*}{\textbf{Single-Material}} & \textbf{Compliance} & 29.954 & 30.264 & 1.078 & 29.400 & 33.008 \\
            &                                           & \textbf{Mass}       & 5399.8 & 5399.9 & 0.35 & 5399.4 & 5400.6 \\
        \hline
        \multirow{4}{*}{\textbf{PIGP 3D}}
            & \multirow{2}{*}{\textbf{Single-Material}} & \textbf{Compliance} & 25.421 & 25.870 & 1.068 & 24.911 & 28.242 \\
            &                                            & \textbf{Mass}       & 5434.7 & 5435.8 & 13.92 & 5406.9 & 5454.9 \\
            & \multirow{2}{*}{\textbf{Multi-Material}}    & \textbf{Compliance} & 9.442 & 9.848 & 2.223 & 7.139 & 13.176 \\
            &                                            & \textbf{Mass}       & 5453.1 & 5448.4 & 16.84 & 5406.1 & 5471.5 \\
        \hline
    \end{tabular}
    \end{adjustbox}
    \label{tab heat obj}
\end{table*}

Representative designed topologies by ordered SIMP and our PIGP methods are shown in \Cref{fig heat 2D} in the 2D case. In the single-material setting, PIGP achieves final designs that are broadly consistent with those produced by ordered SIMP, although SIMP typically exhibits finer geometric features. This difference is due to the discrepancy between the parameters that control the feature lengthscales in each method (i.e., filters in SIMP and grid resolution $\Res$ in PGCAN of PIGP), e.g., increasing the filter size in SIMP or decreasing $\Res$ in PIGP will reduce the complexity of the designs. 
In the multi-material setting in \Cref{fig heat 2D} , PIGP yields a consistent material-allocation pattern in which the highly conductive material (red) is preferentially placed near the prescribed temperature BC, in agreement with the SIMP solutions. The ordered-SIMP multi-material designs are also consistent across runs and tend to follow a red-blue-green progression from the lower-left to the upper-right of the domain with occasional small regions occupied by the red material. A similar progression is observed for PIGP which tends to use far less blue material than ordered SIMP. 

\begin{figure*}[!h]
    \centering
    \includegraphics[width = 1.0\textwidth]{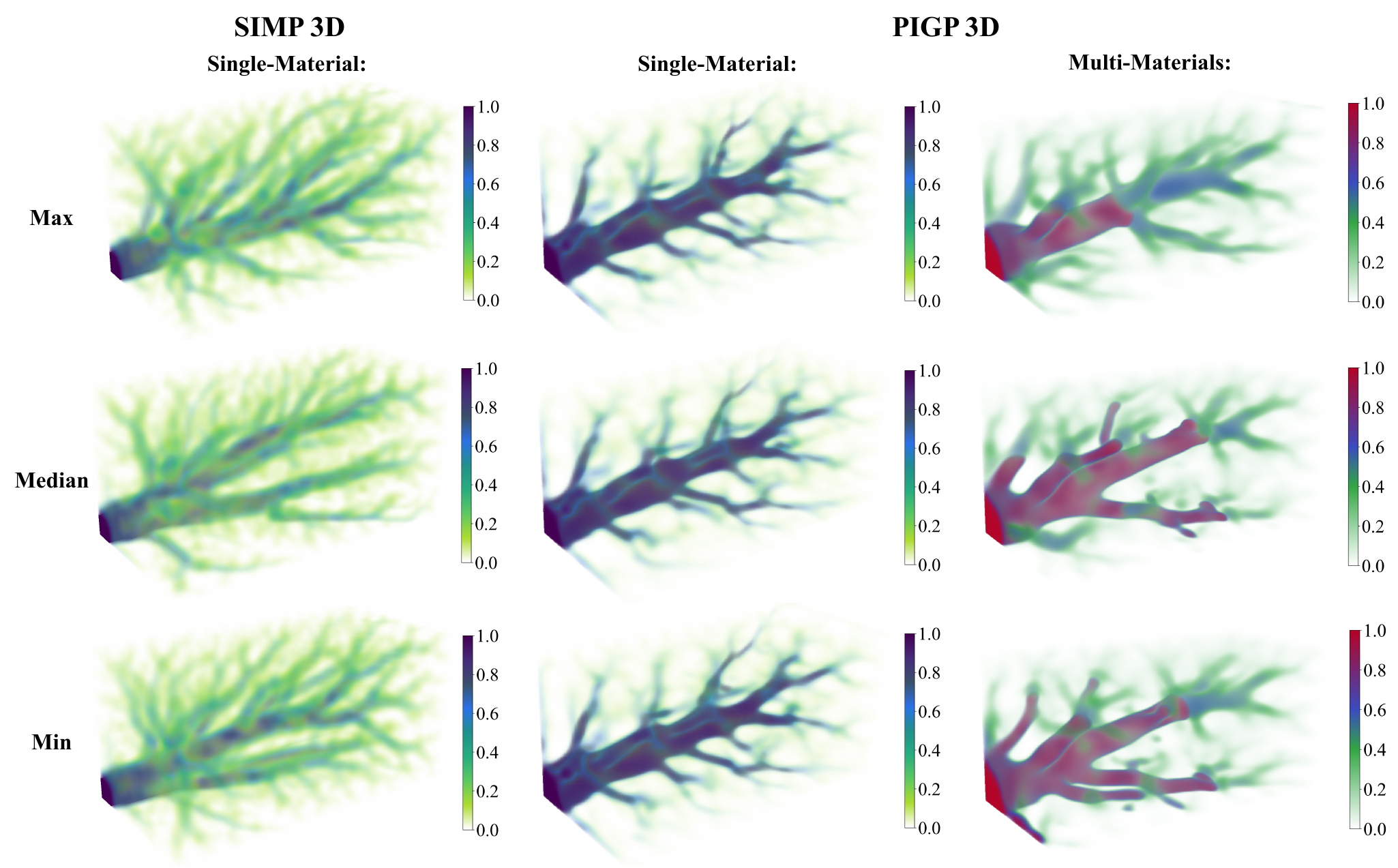}
    \caption{\textbf{Representative designs for 3D heat conduction optimization:}
    For each setting, the designs corresponding to the maximum, median, and minimum thermal compliance values are shown. SIMP 3D \Citep{liu_efficient_2014} is used only in the single-material setting, whereas the proposed framework accommodates both single- and multi-material settings.}
    \label{fig heat 3D}
\end{figure*}

\Cref{tab heat obj} quantitatively compares the thermal compliance and mass of the designs produces by each approach. 
In the single-material case, PIGP attains a lower median thermal compliance (1.775) than ordered SIMP (1.986) but this trend is reversed in the multi-material case. Both methods achieve similar constraint values (i.e., mass).

The 3D topologies obtained using SIMP 3D and PIGP 3D approaches are shown in \Cref{fig heat 3D}. In the single-material setting, PIGP produces a more binarized design than SIMP, whereas SIMP retains more structural features. 
We emphasize that the pronounced gray regions especially in the low-density material phase (green) in both SIMP baselines and PIGP arise from the diffusive nature of the heat conduction problem. 
In the multi-material setting, PIGP yields consistent trends as with the 2D cases where the highly conductive material concentrates near the applied thermal BC area, which serves as the primary outlet for heat dissipation. 

The final thermal compliance values for this 3D heat conduction example are reported in \Cref{tab heat obj}. Under the single-material setting, PIGP achieves a lower median thermal compliance (25.42) than SIMP 3D (29.95). For both the 2D and 3D heat conduction examples, the multi-material designs yield substantially smaller thermal compliance than their single-material counterparts for both ordered SIMP and PIGP, which indicates the advantage of incorporating multiple candidates in practical engineering design tasks. 

\subsection{Multi-Material 2D Compliant Mechanism Design}\label{subsec result cmpt}
\begin{figure*}[!t]
    \centering
    \begin{subfigure}[t]{0.45\textwidth}
        \includegraphics[width=\linewidth]{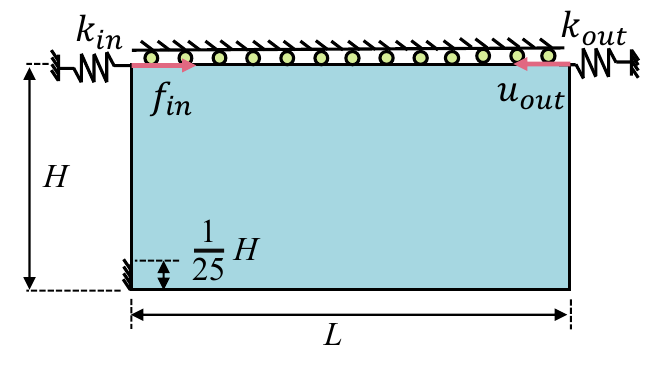}
        \vspace{-13.5em}
        \captionsetup{justification=raggedright, singlelinecheck=false, skip=-3.5pt, position=top}
        \caption[]{\textbf{Displacement Invertor}}
        \label{fig exam cmpt 1}
    \end{subfigure}
    \begin{subfigure}[t]{0.47\textwidth}
        \includegraphics[width=\linewidth]{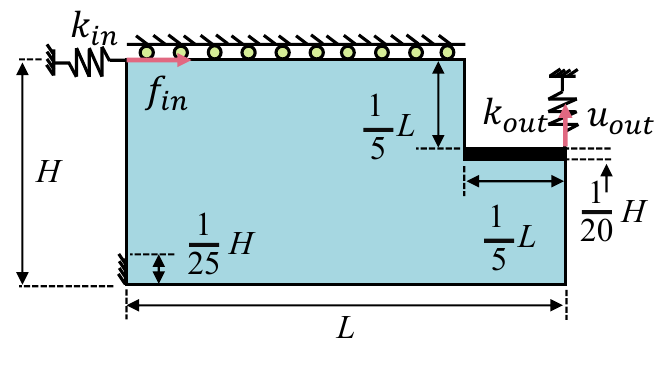}
        \vspace{-13.5em}
        \captionsetup{justification=raggedright, singlelinecheck=false, skip=-3.5pt, position=top}
        \caption[]{\textbf{Compliant Gripper}}
        \label{fig exam cmpt 2}
    \end{subfigure}
    \vspace{-5mm}
    \caption{\textbf{Benchmark examples for compliant mechanism design:} Design domains and BCs for (a) the displacement inverter and (b) the compliant gripper. In both examples, the objective is to maximize the output displacement $u_{\mathrm{out}}$.}
    \label{fig exam cmpt}
\end{figure*}

Unlike in \Cref{subsec result cm,subsec result heat}, compliant mechanism design is not self-adjoint and hence adjoint displacement fields must be solved to evaluate sensitivities. 
We consider the two benchmarks shown in \Cref{fig exam cmpt} where the design objective is to maximize the output displacement at a designated node along a prescribed direction subject to a mass constraint. The first example is a displacement inverter where the output displacement is in the opposite direction to the applied input force. The second example is a compliant gripper where the desired displacement is oriented in the positive vertical direction. The parameters for both examples are summarized in \Cref{tab cmpt param}. 
Both single-material and multi-material settings are considered, with the Young’s moduli and discrete density levels for the three artificial materials given in \Cref{tab cm mat}. 
We compare our approach to ordered SIMP and report their grid settings in \Cref{tab cmpt grid}.

\begin{table*}[!b]
    \centering
    \renewcommand{\arraystretch}{1.5}
    \small
    \setlength\tabcolsep{6pt}
    \begin{adjustbox}{width=\linewidth}
    \begin{tabular}{l|c|c|c|c|c|c|c|c} 
    \hline
    \textbf{Example} 
    & $H$ ($\nimm$) 
    & $L$ ($\nimm$) 
    & Thickness ($\nimm$) 
    & $k_{in}$ ($\mathrm{N/mm}$) 
    & $k_{out}$ ($\mathrm{N/mm}$) 
    & $f_{in}$ ($\niN$)
    & \textbf{$\psi_m$}
    & $M_0$\\ 
    \hline
    Displacement Inverter & 100 & 200 & 1 &  $1\times 10^{-1}$ & $1\times 10^{-3}$ & 0.1 & 0.3 & 2$\times10^4$\\
    Compliant Gripper     & 100 & 200 & 1 &  $1\times 10^{-1}$ & $1\times 10^{-1}$ & 0.1 & 0.3 & 1.84$\times10^4$\\
    \hline
    \end{tabular}
    \end{adjustbox}
    \caption{\textbf{Design parameters for compliant mechanism examples:}
    Geometric dimensions, connected spring stiffness, prescribed mass fraction, reference mass and applied external force corresponding to the problems shown in \Cref{fig exam cmpt}. All geometric dimensions are given in $\mathrm{\mu m}$. The parameter $k_{in}$ and $k_{out}$ denote the stiffness of the connected spring elements, respectively. }
    \label{tab cmpt param}
\end{table*}

\begin{table*}[!b]
    \centering
    \renewcommand{\arraystretch}{1.5}
    \small
    \setlength\tabcolsep{8pt}
    \begin{tabular}{l|c|cc} 
    \hline
    \multirow{2}{*}{\textbf{Example}} 
    & \multicolumn{1}{c|}{\textbf{Ordered SIMP 2D}}
    & \multicolumn{2}{c}{\textbf{PIGP 2D}} \\ 
    \cline{2-4}
    & \textbf{FE nodes}
    & \textbf{Coarse}
    & \textbf{Fine} \\
    \hline
    Displacement Inverter 
        & \multirow{2}{*}{(201,101)}
        & \multirow{2}{*}{(201,101)}
        & \multirow{2}{*}{(401,201)} \\
    Compliant Gripper 
        & 
        & 
        & \\
    \hline
    \end{tabular}
    \caption{\textbf{Grid settings for the compliant mechanism benchmarks:} We report the FE node counts and CP resolutions of the dynamic grids for SIMP and PIGP, respectively.}
    \label{tab cmpt grid}
\end{table*}

\begin{figure*}[!t]
    \centering
    \begin{subfigure}[t]{\textwidth}
        \includegraphics[width=0.95\linewidth]{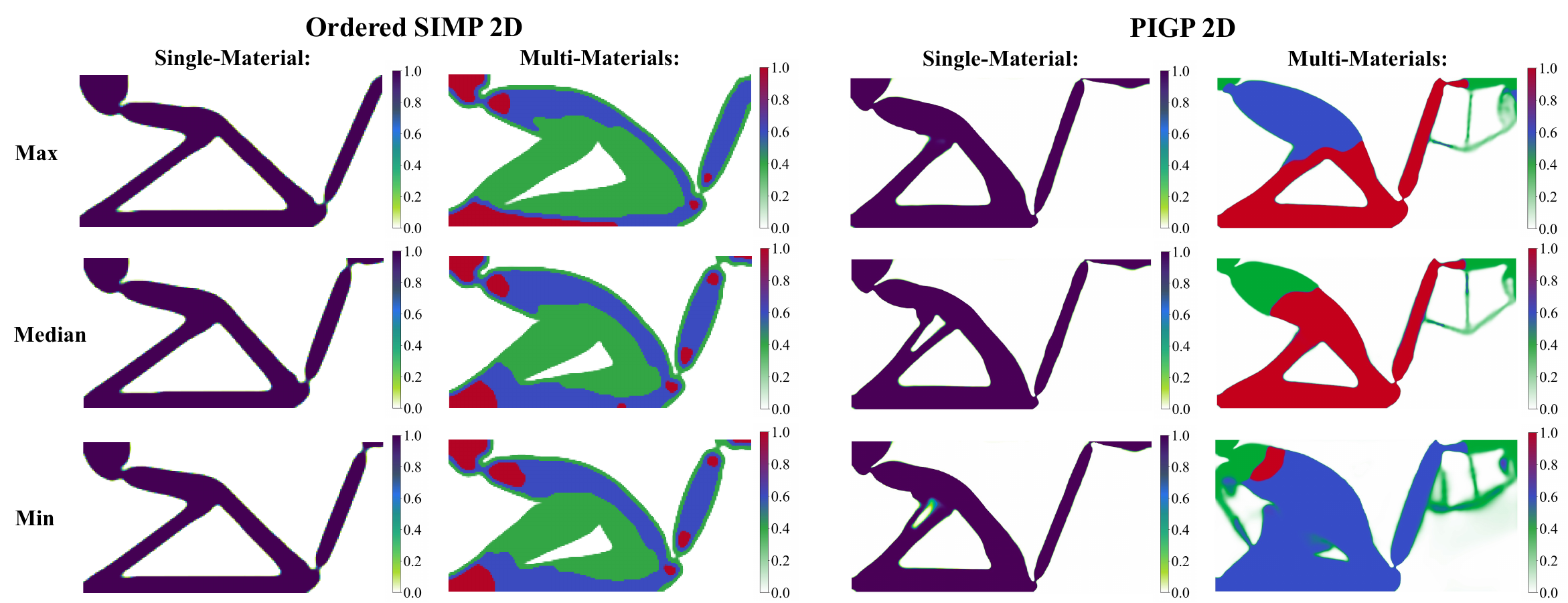}
        \vspace{-17em}
        \captionsetup{justification=raggedright, singlelinecheck=false, skip=-1.5pt, position=top}
        \caption[]{\textbf{Displacement Inverter:}}
        \label{fig cmpt top 1}
    \end{subfigure}
    \par\addvspace{1.5em}
    \begin{subfigure}[t]{\textwidth}
        \includegraphics[width=0.95\linewidth]{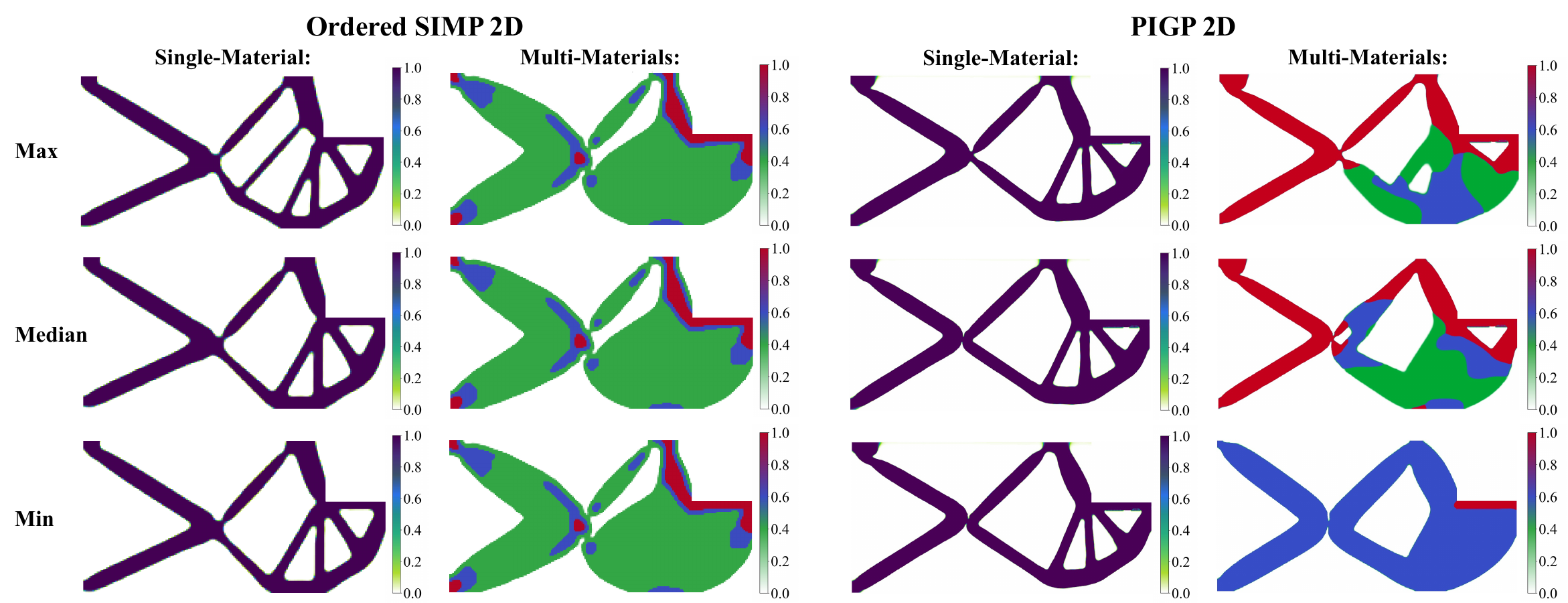}
        \vspace{-17em}
        \captionsetup{justification=raggedright, singlelinecheck=false, skip=-1.5pt, position=top}
        \caption[]{\textbf{Compliant Gripper:}}
        \label{fig cmpt top 2}
    \end{subfigure}
    \caption{\textbf{Representative topologies for compliant mechanism design:}
    Three independent runs corresponding to the maximum, median, and minimum objective values are shown for both the single-material and multi-material settings.}
    \label{fig cmpt top}
\end{figure*}

The optimized topologies for the displacement inverter and compliant gripper are shown in \Cref{fig cmpt top 1} and \Cref{fig cmpt top 2}, respectively. In the single-material setting, the final designs from the two methods are quite similar with PIGP producing slightly more fine-features across runs in the displacement inverter case. 
In the multi-material version of both benchmarks, across different runs PIGP produces designs with larger structural variability and smaller output displacements than those obtained by ordered SIMP, see also \Cref{tab cmpt obj}. We also observe that PIGP produces more gray areas in the displacement inverter case which indicate that the 10k training epochs was insufficient for this example. 



\begin{table*}[!t]
    \centering
    \renewcommand{\arraystretch}{1.5}
    \small
    \setlength\tabcolsep{8pt}
    \caption{\textbf{Statistics of objective and mass values for compliant mechanism designs:}
    The values are obtained across 10 runs where FEM is used to evaluate the final designs from both approaches.}
    \begin{adjustbox}{width=\linewidth}
    \begin{tabular}{l l l | c c c | c c c}
        \hline
        \multirow{2}{*}{\textbf{Method}} 
        & \multirow{2}{*}{\textbf{Materials}} 
        & \multirow{2}{*}{\textbf{Metric}}
        & \multicolumn{3}{c|}{\textbf{Displacement Inverter}} 
        & \multicolumn{3}{c}{\textbf{Compliant Gripper}} \\
        \cline{4-9}
        & 
        & 
        & \textbf{Median} & \textbf{Mean} & \textbf{Std}
        & \textbf{Median} & \textbf{Mean} & \textbf{Std} \\
        \hline
        \multirow{4}{*}{\textbf{Ordered SIMP 2D}}
            & \multirow{2}{*}{\textbf{Single-Material}}
            & \textbf{Objective} 
            & 2.197 & 2.197 & 1.8$\times10^{-2}$ & 0.099 & 0.099 & 2.7$\times10^{-4}$ \\
            & 
            & \textbf{Mass}
            & 5983.0 & 5982.8 & 3.54 & 5501.5 & 5500.1 & 3.96 \\
            & \multirow{2}{*}{\textbf{Multi-Material}}
            & \textbf{Objective}
            & 2.243 & 2.247 & 1.3$\times10^{-2}$ & 0.109 & 0.109 & 2.1$\times10^{-4}$ \\
            &
            & \textbf{Mass}
            & 5987.1 & 5987.5 & 2.03 & 5508.2 & 5508.1 & 0.99 \\
        \hline
        \multirow{4}{*}{\textbf{PIGP 2D}}
            & \multirow{2}{*}{\textbf{Single-Material}}
            & \textbf{Objective}
            & 2.163 & 2.156 & 2.8$\times10^{-2}$ & 0.094 & 0.094 & 1.6$\times10^{-3}$ \\
            &
            & \textbf{Mass}
            & 6000.5 & 5997.3 & 12.07 & 5490.0 & 5489.9 & 4.46  \\
            & \multirow{2}{*}{\textbf{Multi-Material}}
            & \textbf{Objective}
            & 1.979 & 1.910 & 1.8$\times10^{-1}$ & 0.097 & 0.096 & 2.1$\times10^{-3}$ \\
            &
            & \textbf{Mass}
            & 5952.6 & 5944.8 & 51.27 & 5503.1 & 5501.7 & 4.55  \\
        \hline
    \end{tabular}
    \end{adjustbox}
    \label{tab cmpt obj}
\end{table*}

\subsection{Thermo-mechanical Device Design}\label{subsec result thermo}
\begin{figure*}[!b]
    \centering
    \begin{subfigure}[t]{0.45\textwidth}
        \includegraphics[width=\linewidth]{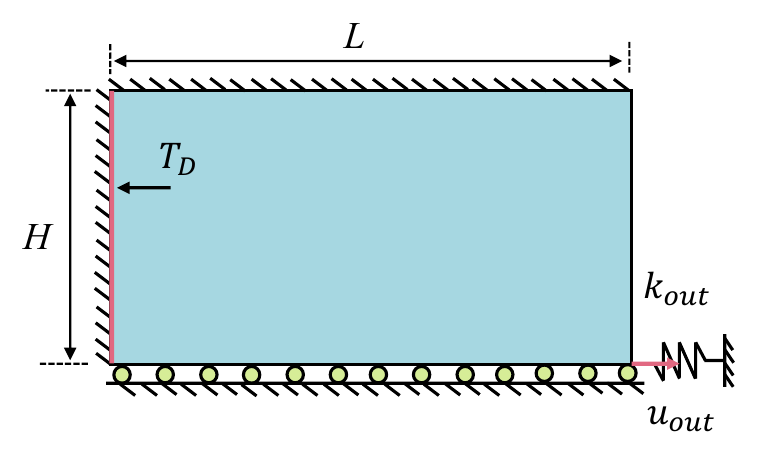}
        \vspace{-13.5em}
        \captionsetup{justification=raggedright, singlelinecheck=false, skip=-3.5pt, position=top}
        \caption[]{\textbf{Thermal Actuator}}
        \label{fig exam thermo 1}
    \end{subfigure}
    \begin{subfigure}[t]{0.45\textwidth}
        \includegraphics[width=\linewidth]{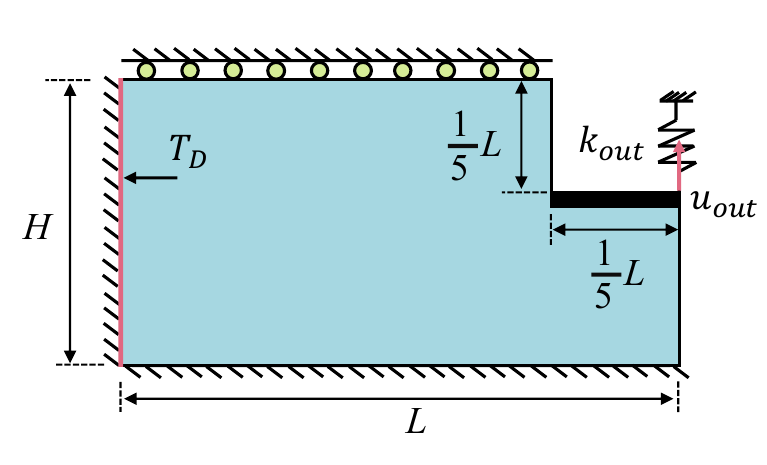}
        \vspace{-13.5em}
        \captionsetup{justification=raggedright, singlelinecheck=false, skip=-3.5pt, position=top}
        \caption[]{\textbf{Thermal Gripper}}
        \label{fig exam thermo 2}
    \end{subfigure}
    \caption{\textbf{Thermo-mechanical compliant device design with the objective of maximizing output displacement:} Design domains and BCs for \textbf{(a)} the thermal actuator and \textbf{(b)} the thermal gripper. For both examples, a high-temperature BC is applied along the left edge.}
    \label{fig exam thermo}
\end{figure*}

To demonstrate the capability of our PIGP framework on more complex applications, we study thermo-mechanical devices that are not self-adjoint and couple thermal and mechanical fields. We consider a thermal actuator and a thermal gripper which are shown in \Cref{fig exam thermo} with the corresponding design parameters summarized in \Cref{tab thermo param}. In both cases, the objective is to maximize the output displacement $u_{\mathrm{out}}$ subject to a prescribed mass fraction constraint. 
Unlike compliant mechanism design where actuation is generated by an external mechanical load, actuation here is driven by a prescribed high-temperature BC with $T_D = 673\niK$ applied along the left edge. The reference temperature is chosen as $T_0 = 293\niK$ for the entire design domain. 
To model volumetric thermal convection, we include a constant heat-sink term $s<0$ for all candidate materials, which balances the thermal energy input from the hot boundary.

Motivated by prior multi-physics TO studies on micro-actuator design \citep{sigmund_design_2001,du_topology_2009}, we adopt Nickel (Ni) as the candidate material for the single-material setting. For the multi-material version we consider four additional engineering metals whose properties including thermal expansion coefficient, thermal conductivity, density, and Young’s modulus are listed in \Cref{tab thermo prop}. Among these, Aluminum (Al) and Titanium (Ti) have relatively low density compared to Copper (Cu) or Iron (Fe) and we aim to understand the effect of this difference on the designed topologies. To this end, we consider two cases with (Al, Fe, Cu) and (Ti, Fe, Cu) as the candidate materials. 

\begin{table*}[!t]
    \centering
    \renewcommand{\arraystretch}{1.5}
    \small
    \setlength\tabcolsep{8pt}
    \begin{adjustbox}{width=\linewidth}
    \begin{tabular}{l|c|c|c|c|c|c|c|c} 
    \hline
    \textbf{Example} 
    & $H$ ($\nimum$) 
    & $L$ ($\nimum$) 
    & Thickness ($\nimum$) 
    & $k_{out}$ ($\niN/\nimum$) 
    & \textbf{$\psi_m$} 
    & $M_0$ 
    & \textbf{$T_D$} ($\niK$)
    & \textbf{$T_0$} ($\niK$) \\ 
    \hline

    Thermal Actuator 
        & \multirow{2}{*}{250}
        & \multirow{2}{*}{500}
        & \multirow{2}{*}{15}
        & \multirow{2}{*}{$2\times 10^{-3}$}
        & \multirow{2}{*}{0.25}
        & $1.25\times10^5$
        & \multirow{2}{*}{673}
        & \multirow{2}{*}{293} \\

    Thermal Gripper
        & 
        & 
        & 
        & 
        & 
        & $1.15\times10^5$
        & 
        & \\

    \hline
    \end{tabular}
    \end{adjustbox}
    \caption{\textbf{Design parameters for thermomechanical device examples:}
    Geometric dimensions, connected spring stiffness, prescribed mass fraction, reference mass and thermal BCs used in the thermal actuator and thermal gripper design problems shown in \Cref{fig exam thermo}. All geometric dimensions are given in $\mathrm{\mu m}$. The parameter $k_{out}$ denotes the stiffness of the connected spring element, while $T_D$ and $T_0$ represent the prescribed high-temperature BCs and the reference ambient temperature, respectively.}
    \label{tab thermo param}
\end{table*}

\begin{table*}[!b]
    \centering
    \renewcommand{\arraystretch}{1.4}
    \small
    \setlength\tabcolsep{10pt}
    \caption{\textbf{Material properties for multi-material thermo-mechanical design:}
    Coefficient of thermal expansion $\alpha$, thermal conductivity $\kappa$, physical density $\rho$, Young’s modulus $E$, and heat source $s$ for examples in \Cref{fig exam thermo} are presented.}
    \begin{tabular}{l|c|c|c|c|c}
        \hline
        \textbf{Material} 
        & $\alpha\;(\niK^{-1})$
        & $\kappa\;(\niW/\nimum\niK)$
        & $\rho\;(\nig/\nic\nim^3)$
        & $E\;(\niG\niPa)$
        & $s\;(\niW/\nimum^3\,\niK)$\\
        \hline
        Ni    & $1.5\times 10^{-5}$  & $9.07\times 10^{-5}$ & 8.9  & 200 & -$4.5\times 10^{-8}$\\
        Fe & $1.2\times 10^{-5}$  & $6.00\times 10^{-5}$ & 7.8  & 200 & -$4.5\times 10^{-8}$\\
        Al    & $2.3\times 10^{-5}$  & $2.37\times 10^{-4}$ & 2.7  & 70  & -$4.5\times 10^{-8}$\\
        Cu    & $1.7\times 10^{-5}$  & $4.00\times 10^{-4}$ & 8.96 & 128 & -$4.5\times 10^{-8}$\\
        Ti    & $8.6\times 10^{-6}$  & $2.59\times 10^{-5}$ & 4.5  & 120 & -$4.5\times 10^{-8}$\\
        \hline
    \end{tabular}
    \label{tab thermo prop}
\end{table*}

Since open-source implementations for multi-physics TO remain limited, we compare PIGP to the commercial software COMSOL for the single-material cases (we were unable to successfully implement a multi-phase design setup in COMSOL). Unlike PIGP, a nested workflow is used in COMSOL which solves the coupled temperature and displacement fields first and then updates the density field on a fixed FE mesh using the global convergent MMA optimizer. The FE mesh resolution for COMSOL is reported in \Cref{tab thermo grid} along with the corresponding coarse and fine grid resolutions PIGP. To further improve training stability, we increase the number of training epochs to 20k in PIGP.

To promote near-binary designs and reduce gray regions, COMSOL leverages the projection function:
\begin{equation}\label{eq proj rho}
    P\big(\rho(\xb)\big) =  \frac{\tanh(\beta \rho_{t}) + \tanh\big(\beta (\rho(\xb) - \rho_{t})\big)}{\tanh(\beta \rho_{t}) + \tanh\big(\beta (1 - \rho_{t})\big)},
\end{equation}
where $\rho_{t}=0.5$ is the threshold density and $\beta$ controls the sharpness of the transition between void and solid. We adopt a three-stage continuation scheme with $\beta \in \{8,12,16\}$ where each stage runs for 500 iterations. A Helmholtz density filter of radius $5\nimum$ is also applied to overcome mesh dependency.
To assess the performance of both COMSOL and PIGP, we first binarize the optimized design fields produced by each approach and then use ABAQUS to simulate the resulting deformation under static thermo-mechanical loading.

\begin{table*}[!h]
    \centering
    \renewcommand{\arraystretch}{1.5}
    \small
    \setlength\tabcolsep{8pt}
    \begin{tabular}{l|c|cc} 
    \hline
    \multirow{2}{*}{\textbf{Example}} 
    & \multicolumn{1}{c|}{\textbf{COMSOL}}
    & \multicolumn{2}{c}{\textbf{PIGP 2D}} \\ 
    \cline{2-4}
    & \textbf{FE nodes}
    & \textbf{Coarse}
    & \textbf{Fine} \\
    \hline
    Thermal Actuator 
        & \multirow{2}{*}{(135, 68)}
        & \multirow{2}{*}{(217, 109)}
        & \multirow{2}{*}{(277, 139)} \\
    Thermal Gripper 
        & 
        & 
        & \\
    \hline
    \end{tabular}
    \caption{\textbf{Grid resolution for the thermo-mechanical design applications:} We report the FE node counts and CP resolutions of the dynamic grids for COMSOL and PIGP, respectively.}
    \label{tab thermo grid}
\end{table*}

\begin{figure*}[!b]
    \centering
    \begin{subfigure}[t]{0.4\textwidth}
        \includegraphics[width=\linewidth]{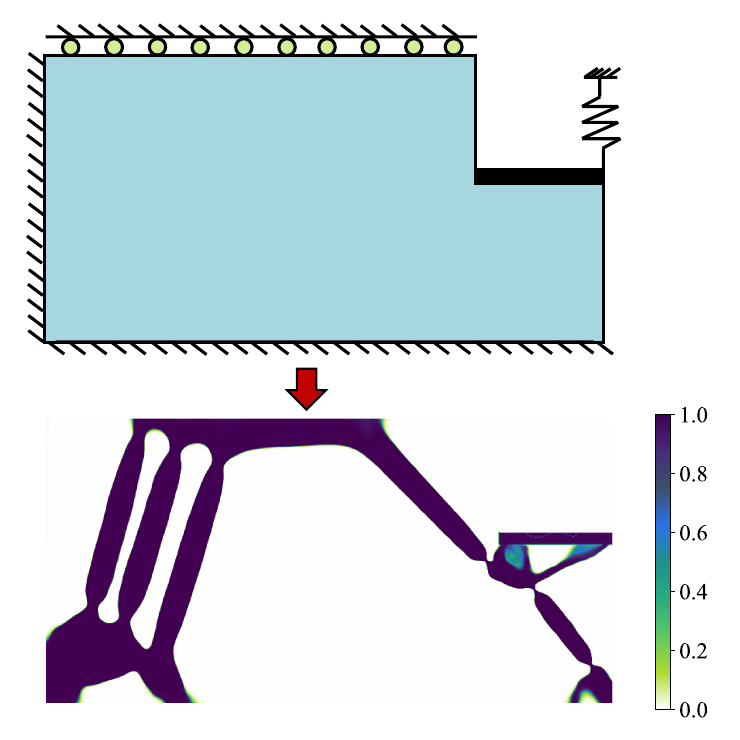}
        \vspace{-19em}
        \captionsetup{justification=raggedright, singlelinecheck=false, skip=-3.5pt, position=top}
        \caption[]{}
        \label{fig thermo sol 1}
    \end{subfigure}
    \begin{subfigure}[t]{0.514\textwidth}
        \includegraphics[width=\linewidth]{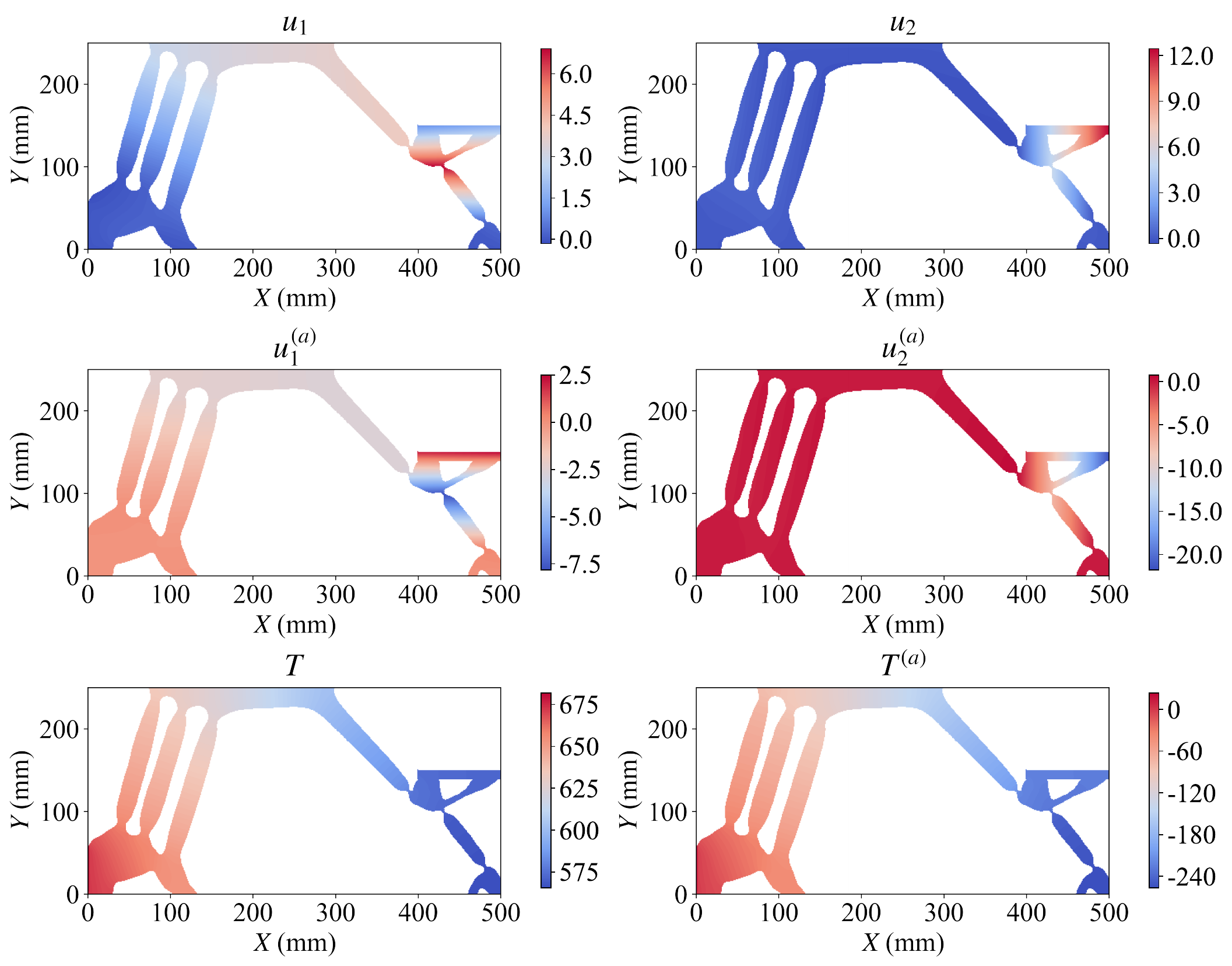}
        \vspace{-19em}
        \captionsetup{justification=raggedright, singlelinecheck=false, skip=-3.5pt, position=top}
        \caption[]{}
        \label{fig thermo sol 2}
    \end{subfigure}
    \caption{\textbf{Thermo-mechanical gripper design:} A representative solution for the thermal gripper showing \textbf{(a)} the design domain and the optimized topology and \textbf{(b)} the corresponding displacement and temperature fields along with their adjoint counterparts.}
    \label{fig thermo sol}
\end{figure*}

We present a representative single-material design solution for the thermal gripper obtained with PIGP in \Cref{fig thermo sol}. The optimized topology exhibits super-resolution structural features with minimal gray regions and multiple beam-like elements that connect the heated left edge to the right-side output point--forming effective thermal conduction pathways. The corresponding displacement and temperature solutions along with their adjoint fields are shown in \Cref{fig thermo sol 2}. At the output point, the vertical displacement $u_2$ is positive, whereas the adjoint displacement $u_2^{(a)}$ points downward, consistent with the sensitivity relation in \Cref{eq bc adj}. 
The temperature field $T$ satisfies the prescribed thermal BCs: the left edge is maintained at $T_D$, and the temperature gradually decreases toward the right edge. The adjoint temperature $T^{(a)}$ is predominantly negative from solving the adjoint thermal PDEs in \Cref{eq pde heat adj}.

\begin{figure*}[!t]
    \centering
    \begin{subfigure}[t]{\textwidth}
        \includegraphics[width=0.95\linewidth]{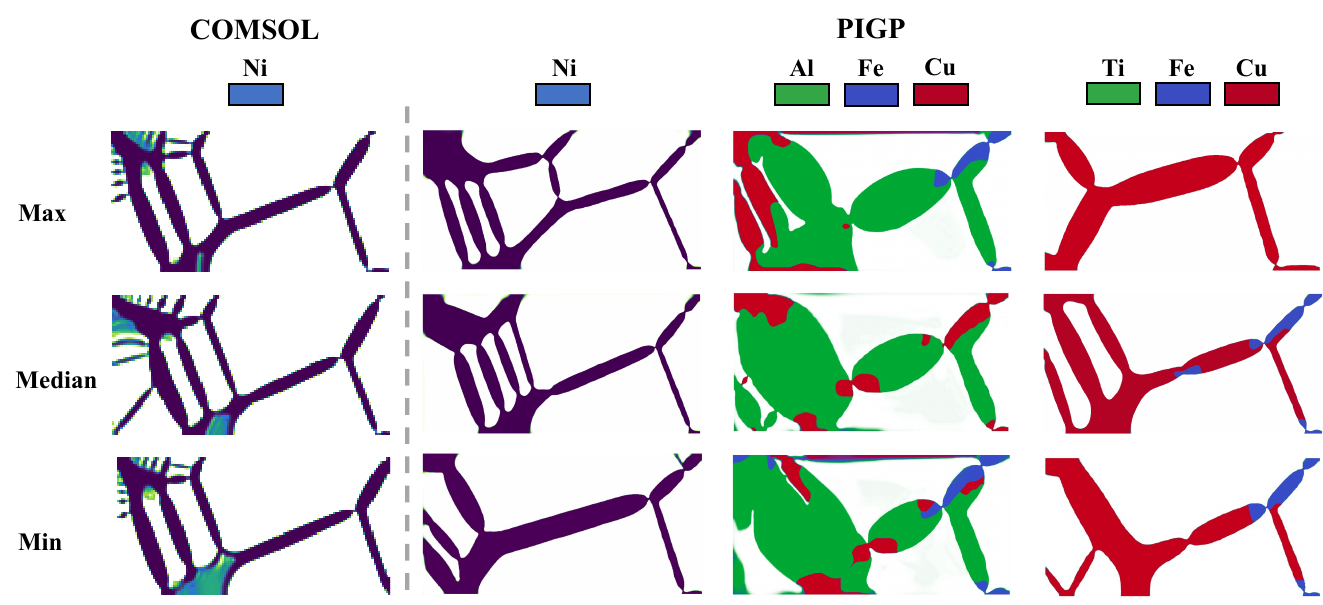}
        \vspace{-18em}
        \captionsetup{justification=raggedright, singlelinecheck=false, skip=-3.5pt, position=top}
        \caption[]{}
        \label{fig thermo comp 1}
    \end{subfigure}
    \begin{subfigure}[t]{\textwidth}
        \includegraphics[width=0.95\linewidth]{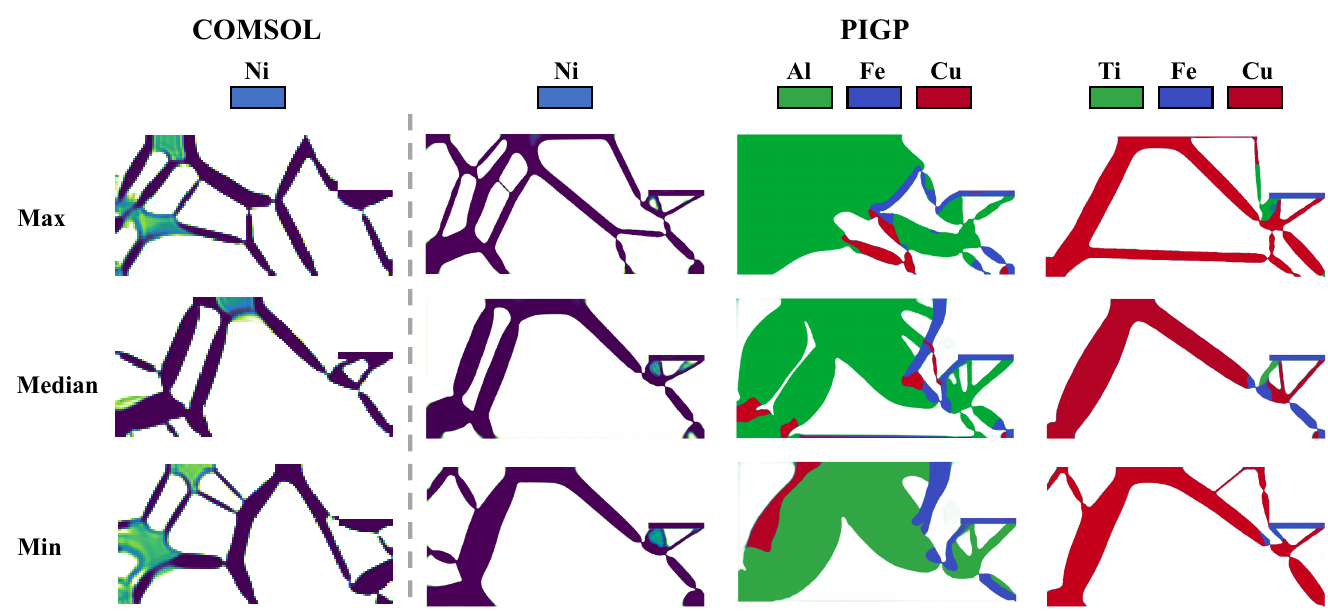}
        \vspace{-18em}
        \captionsetup{justification=raggedright, singlelinecheck=false, skip=-3.5pt, position=top}
        \caption[]{}
        \label{fig thermo comp 2}
    \end{subfigure}
    \caption{\textbf{Thermo-mechanical designs of PIGP vs COMSOL:}
    Optimized topologies for (a) the thermal actuator and (b) the thermal gripper. COMSOL is evaluated under the single-material setting using Ni, whereas PIGP is evaluated under both single- and multi-material settings where in the latter case two material combinations are considered. The visualized topologies correspond to simulations associated with the maximum, median, and minimum objective values.}
    \label{fig thermo comp}
\end{figure*}

The optimized topologies obtained with COMSOL and PIGP are compared in \Cref{fig thermo comp}, with the thermal actuator designs shown in \Cref{fig thermo comp 1} and the thermal gripper designs shown in \Cref{fig thermo comp 2}. 
In the single-material setting with Ni as the candidate material, the two approaches yield roughly similar topologies for both examples. However, PIGP offers several clear advantages over COMSOL: (1) it provides super-resolution which produces smoother void-solid interfaces; (2) it achieves designs with much less gray area despite the fact that COMSOL employs a threshold projection while PIGP does not; and (3) it exhibits improved robustness with respect to random initialization. 
Additionally, inspecting the output displacements summarized in \Cref{tab thermo obj} we observe that the final objective values calculated via ABAQUS are noticeably higher for PIGP. For example, PIGP attains a median output displacement of 16.61 for the thermal gripper, compared to 13.47 using COMSOL. In addition, PIGP yields substantially smaller standard deviations than COMSOL for both examples, indicating higher consistency of our approach across different initializations.

For the multi-material setting, we consider two combinations of three-material systems: (Al, Fe, Cu) and (Ti-Fe-Cu). Al has lower density but higher thermal conductivity and thermal expansion coefficient than Ti, making it a favorable candidate for thermo-mechanical applications. This expectation is supported by our results: For the (Al, Fe, Cu) combination, Al (green) occupies the majority of the optimized structures for both the thermal actuator and gripper design as shown in \Cref{fig thermo comp}. In contrast, for the (Ti-Fe-Cu) combination, Ti is not selected in the actuator design while only a small region of the gripper exhibits Ti. These trends further demonstrate that PIGP yields physically interpretable material distributions for thermo-mechanical TO. 
Similar to the studies in the previous sections we also notice some gray areas, as well as very thin or disconnected features (see the topologies in \Cref{fig thermo comp} under the Al–Cu–Fe setting), which can be removed or minimized by increasing the number of training epochs.

\Cref{tab thermo obj} also reports the predicted output displacements using ABAQUS for the multi-material designs. Notably, for the thermal gripper with the combination of (Al, Fe, Cu), the standard deviation reaches 3.203, which is substantially larger than the other cases. In particular, the maximum output displacement (18.21) is much higher than the median value (10.13), indicating that PIGP explores a wide region of the design space. 

\begin{table*}[!h]
    \centering
    \renewcommand{\arraystretch}{1.5}
    \small
    \setlength\tabcolsep{8pt}
    \caption{\textbf{Statistics of the objective and mass constraint for the thermo-mechanical design examples:}
    The median, mean, max, and standard deviation of the output displacement (objective) and mass (constraint) values are reported, computed using FEM in ABAQUS on the optimized topologies obtained from COMSOL and the proposed PIGP framework under both single- and multi-material settings for the thermal actuator and thermal gripper examples.}
    \begin{adjustbox}{width=\linewidth}
    \begin{tabular}{l l l | c c c c | c c c c}
        \hline
        \multirow{2}{*}{\textbf{Method}} 
        & \multirow{2}{*}{\textbf{Materials}} 
        & \multirow{2}{*}{\textbf{Metric}}
        & \multicolumn{4}{c|}{\textbf{Thermal Actuator}} 
        & \multicolumn{4}{c}{\textbf{Thermal Gripper}} \\
        \cline{4-11}
        & 
        & 
        & \textbf{Median} & \textbf{Mean} & \textbf{Max} & \textbf{Std}
        & \textbf{Median} & \textbf{Mean} & \textbf{Max} & \textbf{Std} \\
        \hline

        \multirow{2}{*}{\textbf{COMSOL}}
            & \multirow{2}{*}{\textbf{Ni}}
            & Objective
            & 18.23 & 17.75 & 19.51 & 1.954
            & 13.47 & 13.51 & 13.99 & 1.274 \\
            & 
            & Mass
            & 31132.8 & 31140.9 & 31177.9 & 30.53
            & 28778.0 & 28800.9 & 28976.0 & 91.47 \\
        \hline

        \multirow{6}{*}{\textbf{PIGP 2D}}
            & \multirow{2}{*}{\textbf{Ni}}
            & Objective
            & 21.88 & 21.35 & 22.48 & 0.955
            & 16.61 & 16.75 & 18.22 & 0.625\\
            & 
            & Mass
            & 31922.7 & 31892.2 & 32003.9 & 102.5
            & 29332.5 & 29356.0 & 29588.0 & 116.37\\
            
            & \multirow{2}{*}{\textbf{Al-Fe-Cu}}
            & Objective
            & 13.37 & 13.21 & 14.07 & 0.764 
            & 10.13 & 11.23 & 18.21 & 3.203 \\
            & 
            & Mass
            & 31075.1 & 30378.0 & 31326.4 & 1262.7
            & 28839.5 & 28785.4 & 29198.0 & 320.04\\
            
            & \multirow{2}{*}{\textbf{Ti-Fe-Cu}}
            & Objective
            & 14.32 & 14.21 & 14.91 & 0.727
            & 9.93 & 10.02 & 11.17 & 0.590 \\
            & 
            & Mass
            & 31782.1 & 31810.4 & 31949.8 & 95.46
            & 29285.8 & 29294.7 & 29514.7 & 102.01 \\
        \hline
    \end{tabular}
    \end{adjustbox}
    \label{tab thermo obj}
\end{table*}

\begin{figure*}[!t]
    \centering
    \begin{subfigure}[t]{\textwidth}
        \includegraphics[width=\linewidth]{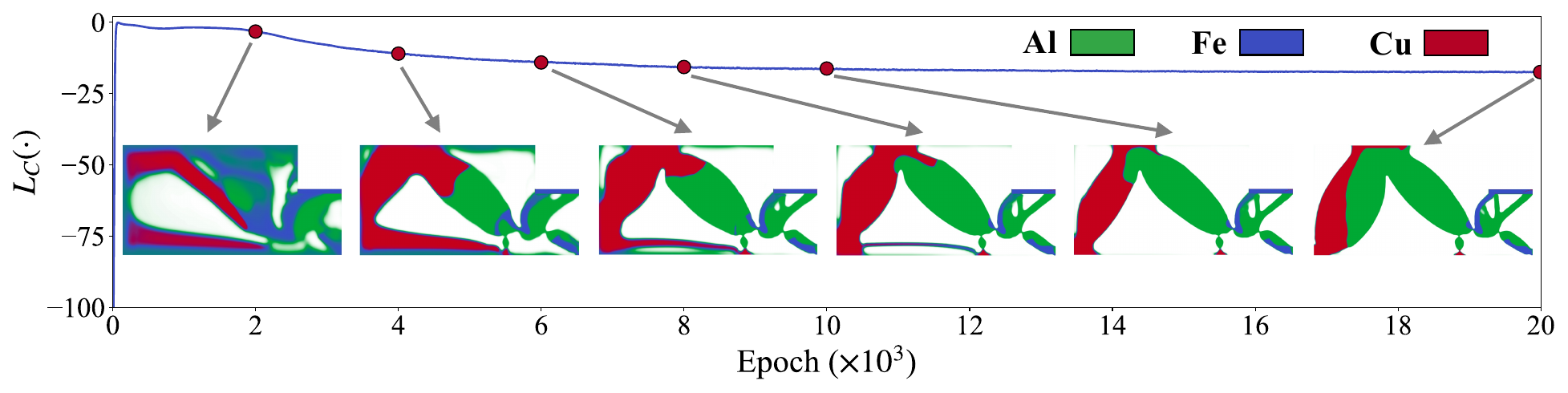}
        \vspace{-12.50em}
        \captionsetup{justification=raggedright, singlelinecheck=false, skip=-3.5pt, position=top}
        \caption[]{}
        \label{fig therm evo 1}
    \end{subfigure}
    \begin{subfigure}[t]{\textwidth}
        \includegraphics[width=\linewidth]{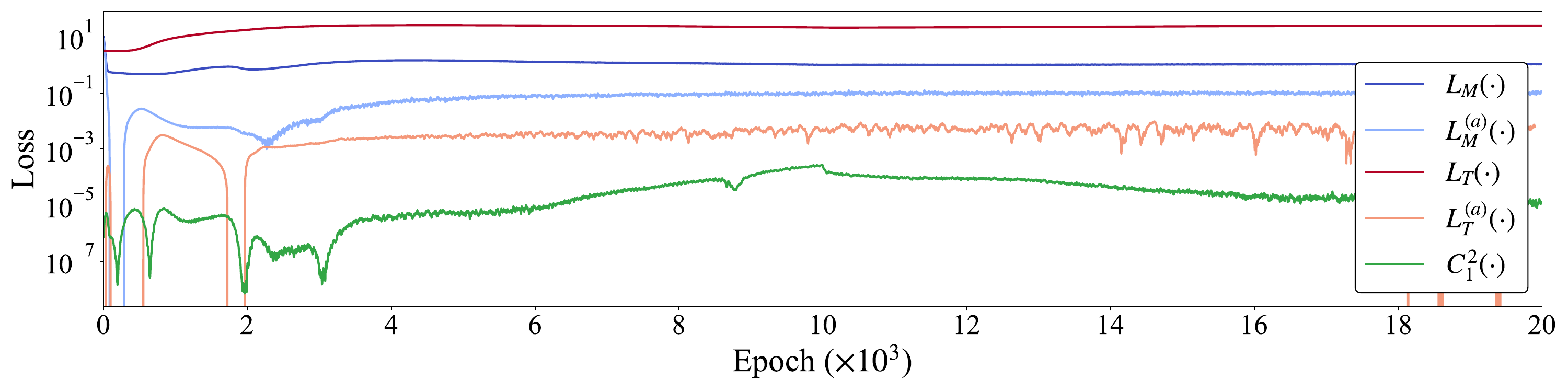}
        \vspace{-12em}
        \captionsetup{justification=raggedright, singlelinecheck=false, skip=-3.5pt, position=top}
        \caption[]{}
        \label{fig therm evo 2}
    \end{subfigure}
    \begin{subfigure}[t]{\textwidth}
        \includegraphics[width=\linewidth]{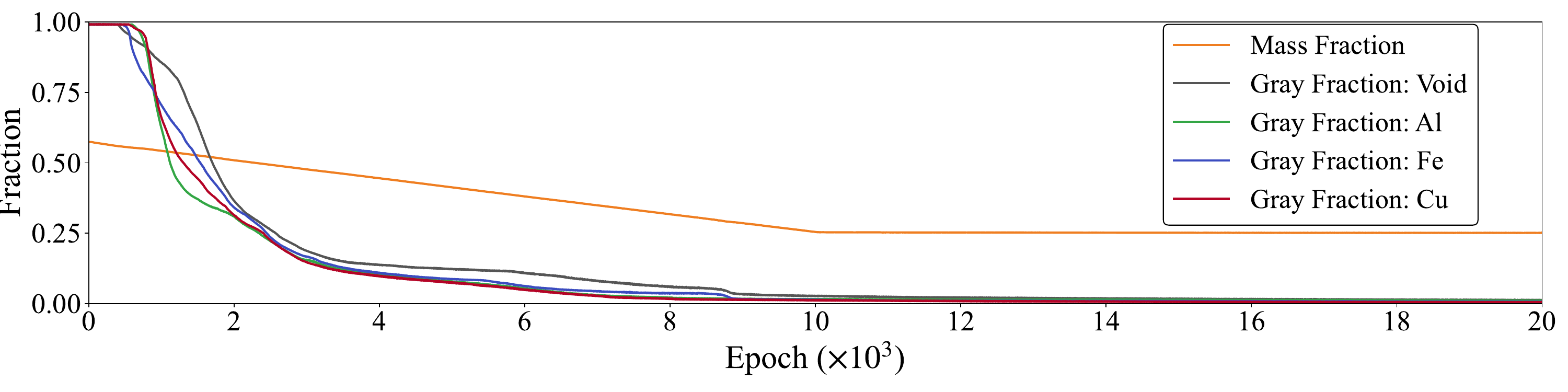}
        \vspace{-12em}
        \captionsetup{justification=raggedright, singlelinecheck=false, skip=-3.5pt, position=top}
        \caption[]{}
        \label{fig therm evo 3}
    \end{subfigure}
    \caption{\textbf{Topology evolution and training dynamics for the thermal gripper:}
    \textbf{(a)} Design objective history with the intermediate topologies at selected epochs. 
    \textbf{(b)} Evolution of the five loss components including the potential energy functionals for the displacement and temperature fields and their adjoint fields, along with the mass-fraction residual. 
    \textbf{(c)} Evolution of the gray area fraction and the scheduled mass fraction. Similar trends are observed for the other thermo-mechanical examples.}
    \label{fig therm evo}
\end{figure*}

To gain further insight into the performance and training dynamics of our PIGP approach, we visualize the objective history and topology evolution at selected epochs in \Cref{fig therm evo 1} for the thermal gripper with the (Al, Fe, Cu) material combination (the objective is multiplied by a negative sign during optimization to convert maximization to minimization). Overall, the objective decreases smoothly and stabilizes around $-20$ after approximately 10k epochs. At around 2k epochs, a meaningful structural layout has already emerged but the topology still contains large undefined regions and gray areas. As training proceeds, these regions are progressively eliminated, yielding a clean and well-defined design at convergence. We also observe a clear material-evolution trend: as the epoch number increases, Al (green) gradually expands while Cu (red) shrinks, indicating that PIGP preferentially selects Al over Cu due to its excellent thermo-mechanical properties. The stiffest material (Fe in blue) is consistently concentrated near the hinge region where stress localization is expected.

The training histories of the potential-energy functionals for the displacement and temperature fields, $L_M(\cdot)$ and $L_T(\cdot)$, their corresponding adjoint fields, $L_M^{(a)}(\cdot)$ and $L_T^{(a)}(\cdot)$, and the mass-fraction residual term $C_1^2(\cdot)$ are shown in \Cref{fig therm evo 2}. The trajectories of $L_M(\cdot)$ and $L_T(\cdot)$ evolve smoothly, indicating stable optimization of the primal displacement and temperature solutions. In contrast, the adjoint losses exhibit larger oscillations in the early epochs, reflecting the search for consistent adjoint equilibrium as the design field evolves. Notably, $C_1^2(\cdot)$ remains at the minimum among the five curves throughout training, suggesting that the mass fraction constraint is effectively enforced at all times.

As shown in \Cref{fig therm evo 3}, the mass fraction closely follows the curriculum training schedule described in \Cref{subsec method training}, which decreases linearly over the first 50\% of the epochs and remains fixed thereafter.
We define the gray area as the fraction of element centers whose volume fraction lies between 0.1 and 0.9 for each material phase including void \citep{sun_smo_2025}. The gray area fractions for the four material phases decrease rapidly during the first half of training and remain below 5\% subsequently. This effective reduction of intermediate phase fraction highlights the ability of PIGP to produce binary, well-defined topologies even when projection functions are not used.

\begin{figure*}[!t]
    \centering
    \begin{subfigure}[t]{0.48\textwidth}
        \includegraphics[width=\linewidth]{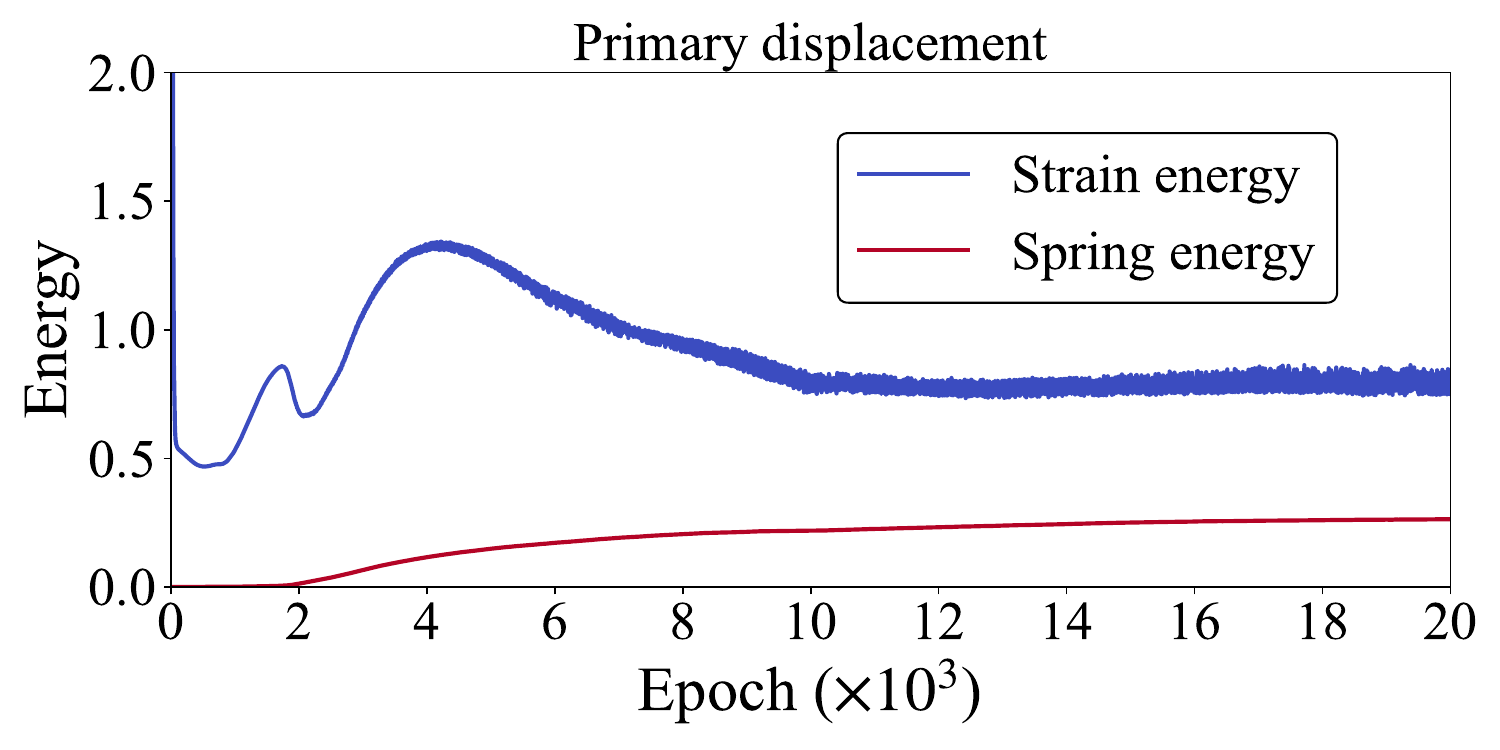}
        \vspace{-11.50em}
        \captionsetup{justification=raggedright, singlelinecheck=false, skip=-6.5pt, position=top}
        \caption[]{}
        \label{fig eng 1}
    \end{subfigure}
    \begin{subfigure}[t]{0.48\textwidth}
        \includegraphics[width=\linewidth]{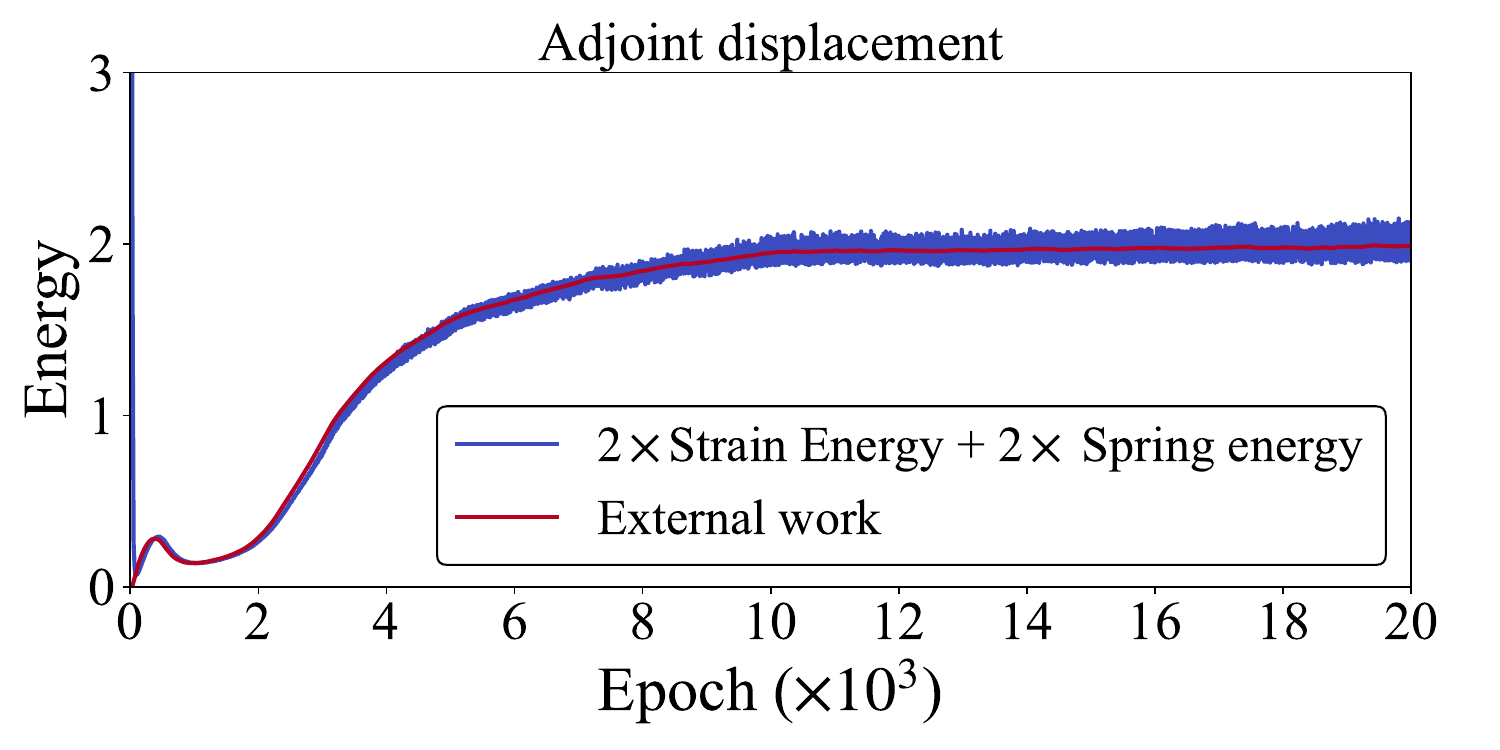}
        \vspace{-11.5em}
        \captionsetup{justification=raggedright, singlelinecheck=false, skip=-3.5pt, position=top}
        \caption[]{}
        \label{fig eng 2}
    \end{subfigure}
    \begin{subfigure}[t]{0.48\textwidth}
        \includegraphics[width=\linewidth]{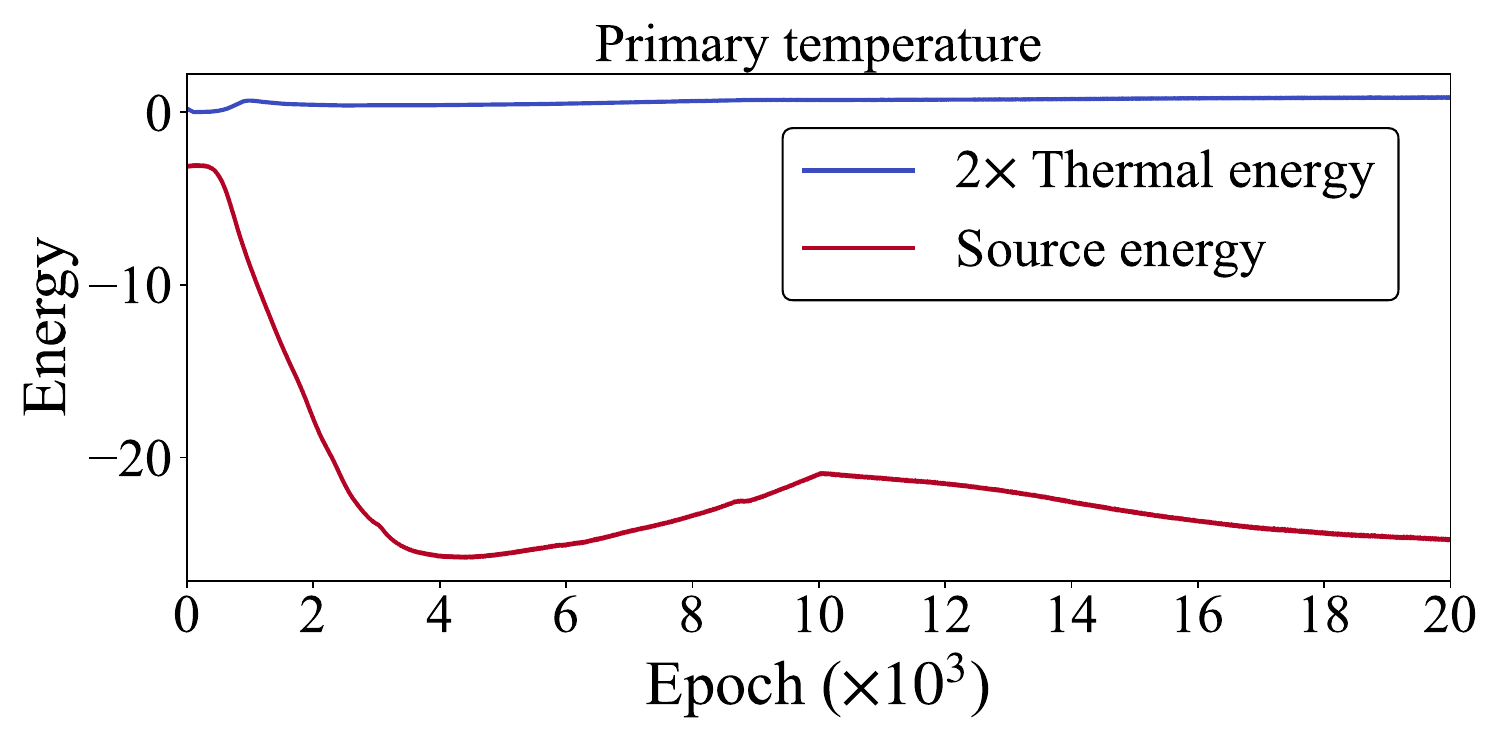}
        \vspace{-11.5em}
        \captionsetup{justification=raggedright, singlelinecheck=false, skip=-3.5pt, position=top}
        \caption[]{}
        \label{fig eng 3}
    \end{subfigure}
    \begin{subfigure}[t]{0.48\textwidth}
        \includegraphics[width=\linewidth]{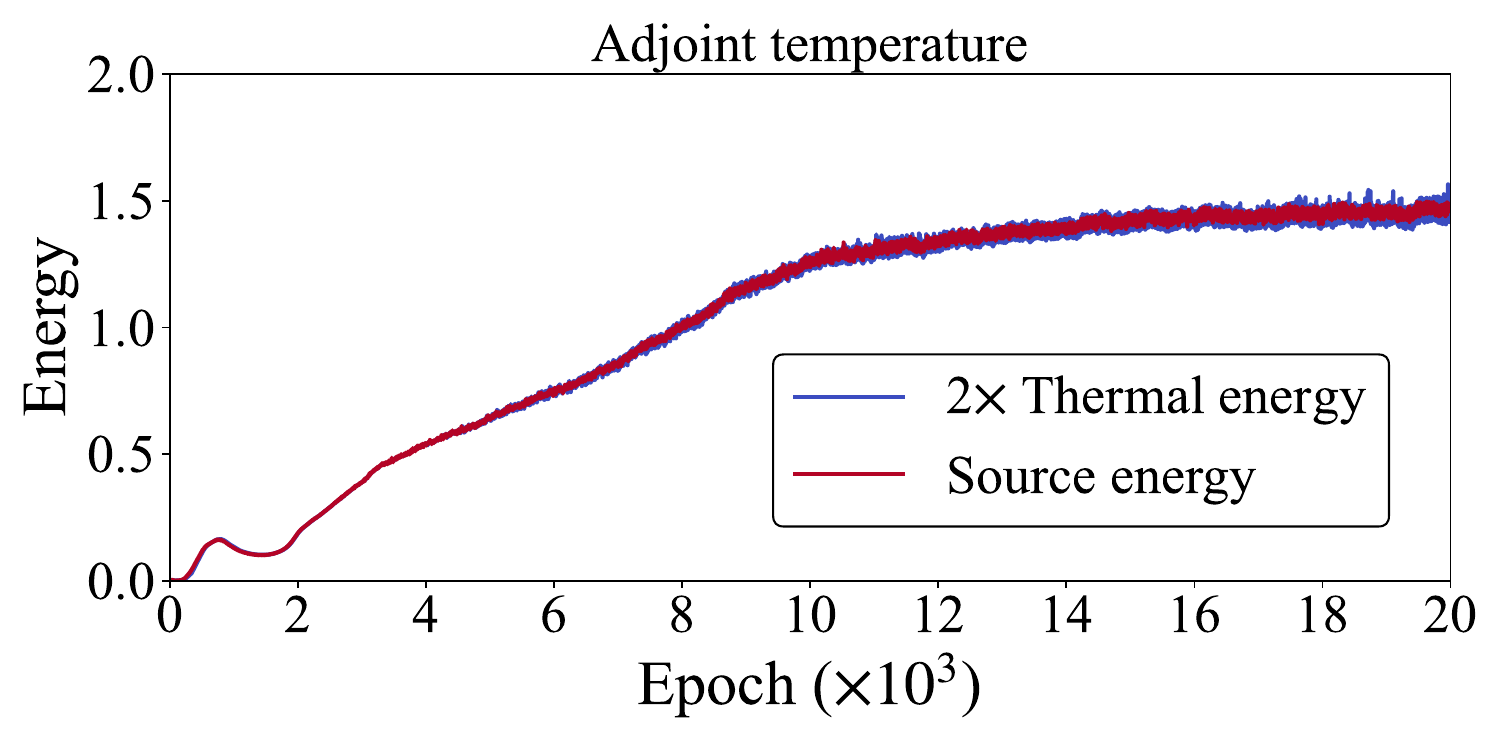}
        \vspace{-11.5em}
        \captionsetup{justification=raggedright, singlelinecheck=false, skip=-3.5pt, position=top}
        \caption[]{}
        \label{fig eng 4}
    \end{subfigure}
    \caption{\textbf{Energy histories for the displacement, temperature, and adjoint fields:}
    Evolution of \textbf{(a)} the stored strain energy and spring energy for the primal displacement field, 
    \textbf{(b)} the stored strain energy, spring energy, and external work for the adjoint displacement field, 
    \textbf{(c)} the thermal energy and source-loss energy for the temperature field, and 
    \textbf{(d)} the thermal energy and source energy for the adjoint temperature field.}
    \label{fig eng}
\end{figure*}

Since the potential-energy functionals for the displacement and temperature fields, $L_M(\cdot)$ and $L_T(\cdot)$, and their adjoint counterparts, $L_M^{(a)}(\cdot)$ and $L_T^{(a)}(\cdot)$ contain multiple energy contributions, we further decompose and visualize their energy evolutions in \Cref{fig eng}. The definitions of the constituent terms—e.g., strain energy, spring energy, and external work for the displacement fields, as well as thermal energy and source energy for the temperature field—are provided in \Cref{eq pi solid} and \Cref{eq pi heat}, respectively.
For the primal displacement field (\Cref{fig eng 1}), we plot the stored strain energy and the spring energy--observing a gradual increase in spring energy due to the growth of output displacement during optimization. For the adjoint displacement field (\Cref{fig eng 2}), $2\times$ the total stored energy (strain energy plus spring energy) matches the external work, indicating that the force equilibrium condition is well enforced throughout training \citep{sun_smo_2025}.
A similar trend is observed for the adjoint temperature field (\Cref{fig eng 4}), where the source energy equals $2\times$ the thermal energy, suggesting satisfactory enforcement of the adjoint thermal equilibrium. We emphasize that the equilibrium relation “$2\times$ stored energy equals external work” follows from the bilinear structure of the potential energy functional under homogeneous (zero) BCs. When the BCs are non-homogeneous, such as the prescribed high temperature along the left edge in our thermo-mechanical design, the same identity does not generally hold as demonstrated by the thermal energy and source energy curves for the primal temperature field in \Cref{fig eng 3}.

\section{Conclusions} \label{sec conclusion}

In this work, we developed a unified mesh-free framework for TO that is applicable to a broad range of design problems in both 2D and 3D conditions involving one or multiple material phases. To parameterize the state adjoint, and design variables in a consistent manner, the proposed approach adopts independent GPs as priors and employs separate NNs as mean functions. We selected the PGCAN as the NN architecture due to its strong capability in capturing stress localization and fine-scale design features. Using a Lagrangian penalty formulation, we recast the constrained TO into an unconstrained optimization with a single loss function that integrates the design objective and constraints as well as potential energy functionals for the solution and adjoint fields. Importantly, explicit penalty terms for BCs or design constraints are not required, as these conditions are naturally enforced through the GP priors.

To improve training efficiency and robustness, we incorporated an adaptive grid strategy together with a shape-function-based implementation for gradient evaluation and numerical integration. In addition, we employed a curriculum training scheme by scheduling the gradual decrease of mass or cost fraction to the target values, which stabilizes the optimization process. We found that the shape-function implementation combined with curriculum training substantially improves the consistency of PIGP compared to FD-based gradient evaluations.

We validated the proposed PIGP framework on benchmark TO problems including CM, heat conduction optimization, and compliant mechanism design. For these canonical benchmarks, we compared PIGP against conventional methods including ordered SIMP and PolyMat. The results demonstrate that our PIGP framework: (1) produces consistent and physically interpretable multi-material distributions; (2) achieves objective values comparable to established baselines; (3) accommodates complex dependencies between material properties and design variables; (4) maintains grid-size independence without requiring filtering; and (5) enables high-resolution and continuous topologies.

To assess the capability of our approach on more challenging settings, we also applied it to thermo-mechanical design problems that are not self-adjoint and involve multiple materials and physics. Under the single-material condition, we compared PIGP against a commercial solver (COMSOL) and the results show that it produces structures with improved objectives and gray areas. 


Overall, we believe this work demonstrates the potential of ML in TO. 
In terms of optimizing an objective (e.g., compliance), our approach performs similar to existing open-source codes but provides more flexibility and versatility. While this flexibility simplifies its adoption for solving different design problems (see the various examples in \Cref{sec result}), it comes at the expense of higher training costs compared to methods such as SIMP. It is also noted that, contrary to open-source codes and our method, we observed that general-purpose packages such as COMSOL are slower and require fine-tuning before they can provide competitive results. 

It must be highlighted that our approach may be considered as a hybrid strategy that integrates kernels with deep NNs and leverages numerical approximation to accelerate computations. Within this lens, we believe our work can be extended in a number of important directions that can further improve its performance and flexibility. For instance, the designs in \Cref{subsec result cmpt} include gray areas and very small hinges that pose manufacturability issues. We believe these undesirable features can be prevented by integrating well-known techniques (such as projection functions and dilation/erosion operators) with our approach. 


\section*{Acknowledgments} \label{sec ack}
We appreciate the support from the Office of the Naval Research (award number N000142312485) and National Science Foundation (award number 2238038).

\section*{Declarations} \label{sec dec}

\noindent \textbf{CRediT authorship contribution statement
} Xiangyu Sun: conceptualization, methodology, formal analysis and investigation, and writing—original draft; 
Shirin Hosseinmardi: conceptualization, methodology, and writing—review and editing; 
Amin Yousefpour: conceptualization and methodology; 
Ramin Bostanabad: supervision, funding acquisition, conceptualization, analysis of results, and writing—review and editing.

\noindent \textbf{Conflict of interest} The authors declare that they have no conflicts of interest.

\noindent \textbf{Replication of results} The codes of the model are accessible via \href{https://github.com/Bostanabad-Research-Group}{GitHub} upon publication. 



\bibliographystyle{sn-mathphys-num}
\bibliography{References}
\end{document}